\documentclass[10pt,twocolumn,letterpaper]{article}

\usepackage{cvpr}              

\newcommand{\bD}{\mathbf{D}}

\newcommand{\bF}{\mathbf{F}} 
\newcommand{\bG}{\mathbf{G}}

\newcommand{\bI}{\mathbf{I}}

\newcommand{\bK}{\mathbf{K}}

\newcommand{\bo}{\mathbf{o}}
\newcommand{\bp}{\mathbf{p}}\newcommand{\bP}{\mathbf{P}}
\newcommand{\bQ}{\mathbf{Q}}
\newcommand{\bR}{\mathbf{R}}

\newcommand{\bt}{\mathbf{t}}\newcommand{\bT}{\mathbf{T}}
\newcommand{\bu}{\mathbf{u}}
\newcommand{\bV}{\mathbf{V}}

\newcommand{\bZ}{\mathbf{Z}}

\newcommand{\bPsi}{\boldsymbol{\Psi}}

\newcommand{\cI}{\mathcal{I}}

\makeatletter
\DeclareRobustCommand\onedot{\futurelet\@let@token\@onedot}
\def\@onedot{\ifx\@let@token.\else.\null\fi\xspace}
\def\eg{e.g\onedot} 
\def\ie{i.e\onedot}

\def\etal{et~al\onedot}

\makeatother

\newcommand{\boldparagraph}[1]{\vspace{0.2cm}\noindent{\bf #1:}}

\definecolor{darkgreen}{rgb}{0,0.7,0}

\newcommand{\red}[1]{{\color{red}#1}}
\newcommand{\green}[1]{{\color{green}#1}}

\usepackage{graphicx}
\usepackage{amsmath}
\usepackage{amssymb}
\usepackage{booktabs}
\usepackage{pifont}

\usepackage[accsupp]{axessibility}  

\usepackage{algorithm}
\usepackage{algpseudocode}
\algnewcommand\algorithmicinput{\textbf{Input:}}
\algnewcommand\Input{\item[\algorithmicinput]}

\usepackage{times}
\usepackage{epsfig}
\usepackage{graphicx}
\usepackage{amssymb}
\usepackage{multirow}
\usepackage{float}
\usepackage{booktabs}
\usepackage{caption}
\usepackage{subcaption}
\usepackage[dvipsnames]{xcolor}
\usepackage{makecell}
\definecolor{lpp}{rgb}{0.69, 0.61, 0.85}
\definecolor{dpp}{rgb}{0.59, 0.44, 0.84}

\mathchardef\mhyphen="2D
\usepackage[utf8]{inputenc}
\usepackage{comment}
\usepackage{tabularx,colortbl}
\usepackage{xparse,mathtools}
\usepackage{diagbox}
\usepackage{adjustbox,lipsum}

\usepackage{breqn}
\usepackage[utf8]{inputenc}
\DeclareUnicodeCharacter{2713}{\checkmark} 
\usepackage{tabu}
\usepackage{tabulary,overpic}

\newlength\savewidth

\definecolor{cvprblue}{rgb}{0.21,0.49,0.74}
\usepackage[pagebackref,breaklinks,colorlinks,allcolors=cvprblue]{hyperref}

\usepackage[dvipsnames]{xcolor}
\usepackage[pagebackref,breaklinks,colorlinks,allcolors=cvprblue]{hyperref}

\definecolor{lightgray}{rgb}{0.89, 0.89, 0.89}

\usepackage[capitalize]{cleveref}
\crefname{section}{Sec.}{Secs.}
\Crefname{section}{Section}{Sections}
\Crefname{table}{Table}{Tables}
\crefname{table}{Tab.}{Tabs.}

\def\paperID{7666} 
\def\confName{CVPR}
\def\confYear{2025}

\title{ZeroVO: Visual Odometry with Minimal Assumptions }

\author{Lei Lai$^*$ \quad Zekai Yin$^*$ \quad Eshed Ohn-Bar\\
Boston University\\
{\tt\small \{leilai, zekaiyin, eohnbar\}@bu.edu}
}

\begin{document}
\maketitle

\newcommand\blfootnote[1]{%
\begingroup
\renewcommand\thefootnote{}\footnote{#1}%
\addtocounter{footnote}{-1}%
\endgroup
}
\blfootnote{$^*$ Equally contributed.}


\begin{abstract}
We introduce ZeroVO, a novel visual odometry (VO) algorithm that achieves zero-shot generalization across diverse cameras and environments, overcoming limitations in existing methods that depend on predefined or static camera calibration setups. Our approach incorporates three main innovations. 
First, we design a calibration-free, geometry-aware network structure capable of handling noise in estimated depth and camera parameters. Second, we introduce a language-based prior that infuses semantic information to enhance robust feature extraction and generalization to previously unseen domains. Third, we develop a flexible, semi-supervised training paradigm that iteratively adapts to new scenes using unlabeled data, further boosting the models' ability to generalize across diverse real-world scenarios. We analyze complex autonomous driving contexts, demonstrating over 30\% improvement against prior methods on three standard benchmarks—KITTI, nuScenes, and Argoverse 2—as well as a newly introduced, high-fidelity synthetic dataset derived from Grand Theft Auto (GTA).
By not requiring fine-tuning or camera calibration, our work broadens the applicability of VO, providing a versatile solution for real-world deployment at scale.
\end{abstract}
\section{Introduction}
\label{sec:intro}

For a robot or autonomous vehicle to function reliably in the real world, a \textit{generalized} Visual Odometry (VO) system is essential—one that can robustly estimate the relative camera pose in metric coordinates from a sequence of images under diverse and unforeseen conditions. However, generalization remains a significant challenge for current VO models, which often suffer from lost feature tracks, optimization instability, and drift, particularly when exposed to varying lighting, dynamic scenes, or adverse weather conditions~\cite{ceccarelli2022rgb,kaygusuz2021mdn,pretto2009visual,agarwal2014visual,kim2015robust,wang2020tartanair}.

Due to the inherent difficulty and ambiguity in modeling camera ego-motion, a dynamic 3D world, and real-world scale from 2D images, monocular VO algorithms have traditionally been built on strong assumptions and geometric constraints~\cite{chiuso2002structure,azarbayejani1995recursive,mur2017orb,engel2017direct,zhan2021df,cadena2016past,fraundorfer2011visual,mur2015orb,campos2021orb,rockwell20228,dickmanns2007dynamic}. While carefully designed camera calibration or evaluation on fixed data distributions can be effective in controlled settings, such approaches can limit adaptability and scalability to real-world scenarios with varying configurations that may not align with such assumptions.


VO techniques have increasingly adopted learning-based components to exploit statistical regularities in scene structure and motion dynamics.
However, most learning-based methods rely on privileged ground-truth data (\eg, accurate camera parameters, optical flow) for supervision and often train and evaluate on the same dataset~\cite{cadena2016past,fraundorfer2011visual,teed2021droid,wang2017deepvo,yang2020d3vo,teed2024deep,rockwell20228,kendall2015posenet,kendall2017geometric}.
Although recent studies explore generalization beyond single-dataset settings~\cite{lai2023xvogeneralizedvisualodometry,wang2020tartanvogeneralizablelearningbasedvo,teed2021droid,teed2024deep,lipson2024deeppatchvisualslam}, current models continue to exhibit significant errors in the presence of more complex everyday contexts~\cite{ceccarelli2022rgb,kaygusuz2021mdn,pretto2009visual,agarwal2014visual,kim2015robust}, including harsh conditions such as rainy or snowy nights (\eg, frequent glare, water streaks, reflections, and reduced visibility), lens degradation (\eg, condensation, scratches, dirt), or highly dynamic environments (\eg, dense intersections or aggressive motion). How can we design VO models that generalize across conditions instead of quickly suffering from instability and drift?

In this work, we aim to advance the capabilities of learning-based monocular VO. We introduce \textbf{ZeroVO}, a novel transformer-based approach for robustly predicting relative camera motion at real-world scale across variable scenes in a zero-shot manner. By leveraging cross-attention mechanisms~\cite{vaswani2017attention,dosovitskiy2020image} to efficiently integrate contextual and geometric priors directly into the network architecture, ZeroVO avoids common limiting assumptions—such as reliance on camera calibration or costly optimization steps. 
Specifically, we fuse versatile multimodal text~\cite{liu2024improved,liu2024visual,reimers-2019-sentence-bert} and depth-based priors~\cite{yin2023metric3d,hu2024metric3d,guizilini2023towards,piccinelli2024unidepth} to address inherent scale ambiguity in metric VO. We demonstrate that our proposed model is robust to noisy and uncalibrated setups. We further optimize the model using a novel multimodal semi-supervised training framework that filters noisy pseudo-labels in a geometry and language-guided process. 
Our flexible VO framework achieves state-of-the-art, off-the-shelf performance across diverse autonomous driving datasets. To comprehensively assess system generalizability, we also collect and analyze a novel Grand Theft Auto (GTA) dataset featuring challenging scenarios with harsh weather, high-speed motion, complex traffic scenes, and varied camera settings. Our dataset and code are available at \href{https://zvocvpr.github.io/}{https://zvocvpr.github.io/}.

\section{Related Work}
\label{sec:related_work}

Our framework builds on advances in foundational computer vision models, particularly in metric depth prediction and rich, generalized vision-and-language embeddings.

\boldparagraph{Learning-Based Monocular Visual Odometry}
Learning-based monocular visual odometry tasks can be roughly categorized into two main approaches: neural network models combined with multi-step geometric optimization (\eg, full SLAM~\cite{cadena2016past, teed2021droid, Zhu2023NICER, teed2024deep, lipson2024deeppatchvisualslam, messikommer2024reinforcementlearningmeetsvisual}) or direct, end-to-end relative pose estimation from two or few consecutive frames \cite{lai2023xvogeneralizedvisualodometry, wang2020tartanvogeneralizablelearningbasedvo, vijayanarasimhan2017sfmnetlearningstructuremotion, ye2023pvopanopticvisualodometry}. Hybrid methods such as Droid-SLAM~\cite{teed2021droid} have demonstrated strong performance in dense scene reconstruction and pose estimation. In contrast, two-frame pose regression tends to be more robust in short-distance tracking scenarios, while SLAM and other geometry-based approaches typically require continuous, long-frame sequences. These methods often rely on long-term feature matching and global optimization techniques, such as loop closure detection. Although certain methods~\cite{kanai2024self} can aid in initialization, SLAM remains sensitive to environmental features and accurate motion tracking, \ie, can fail to build and update a reliable map in feature-deficient environments (\eg, corridors or repetitive textures) or highly dynamic settings (\eg, crowds). In contrast, two-frame pose regression is less affected by such conditions as it does not rely on maintaining a global representation. However, two-frame pose regression can be prone to drift accumulation, as it lacks the temporal optimization over extended frame sequences needed to correct for drift. Our work improves over two-frame approaches due to inherent efficiency, versatility (\ie, as input to downstream optimization), and minimal assumptions. 


\boldparagraph{Metric Depth Estimation from Images}  
We leverage advances in metric depth estimation to address the inherent ambiguity in recovering camera translation at real-world scale. Traditional monocular depth models often rely on scale-invariant losses or sparse supervision, making them unsuitable for tasks such as visual odometry that require consistent metric scale.  
Recently, models for predicting \textit{metric} depth have demonstrated practical performance~\cite{depthanything, hu2024metric3d, zeng2024wordepthvariationallanguageprior, piccinelli2024unidepth}. Models such as Depth Anything~\cite{depthanything} and UniDepth~\cite{piccinelli2024unidepth} aim to generalize depth prediction across a wide range of scenes by leveraging large-scale vision foundation models. WordDepth~\cite{zeng2024wordepthvariationallanguageprior} proposes the use of language-guided priors to reduce ambiguity in unconstrained prediction of scale.
Metric3Dv2~\cite{hu2024metric3d} provides a zero-shot model that was trained across numerous datasets and is capable of predicting real-world scale depth (and surface normals) in diverse settings. By leveraging known camera intrinsics and extrinsics, the model learns to transform inputs into a canonical camera space. 
While existing models often struggle in challenging real-world scenarios, we adopt Metric3Dv2 to extract real-scale depth features that enable accurate and robust visual odometry. To further increase the flexibility and applicability of our approach, we do not rely on traditional camera calibration or predefined image information~\cite{888718,zhang2004camera,zhang2007camera,zhang2016flexible}. Instead, we consider settings where calibration may be unavailable or inaccurate, and incorporate single-image camera parameter estimation techniques such as WildCamera~\cite{zhu2024tame} to support inference under uncalibrated conditions.

\boldparagraph{Rich Vision-and-Language Embeddings}  
Language-guided models have shown strong generalization capabilities by effectively bridging multiple modalities. Through joint embedding spaces that capture generalized semantic relationships between images and language, Vision-Language Large Models (VLLMs) models have recently achieved state-of-the-art results in diverse tasks such as image captioning~\cite{dai2017contrastive, you2016image, yang2022unified}, visual question answering~\cite{antol2015vqa}, and cross-modal retrieval~\cite{hirose2019deep}. LLaVA~\cite{liu2024visual}, for instance, is now being broadly used across contexts and tasks~\cite{chen2023llavainteractiveallinonedemoimage,zhang2024feedback,liu2024improved, liu2023llavapluslearningusetools}. Preliminary studies in autonomous driving, \eg, Tian~\etal~\cite{tian2024tokenize}, have shown VLLMs to be useful for robustness under long-tail events. In our work, we propose to integrate VLLMs to extract high-level semantic descriptions of driving scenes that could serve as language-based priors that guide metric-scale odometry and complement adaptive inference under challenging visual conditions. 

\boldparagraph{Semi-Supervised Learning}  
Our work aims to develop flexible models that can effectively adapt to new environments, including through the use of unlabeled data. Semi-supervised learning (SSL) is being increasingly used in computer vision and machine learning tasks, particularly in domains where annotated data is scarce, costly, or requires expert supervision~\cite{berthelot2019mixmatch, berthelot2019remixmatch, tang2021humble, kim2024motion, zhu2023learning,gao2019note, caine2021pseudo, jang2020generalized,clark2018semi, shi2023rethinkingsemisupervisedlearninglanguage, gururangan2020don, gururangan2019variational,lai2024uncertainty}. In the context of visual odometry, SSL can potentially enable the use of large-scale, unlabeled video data, such as web videos~\cite{zhang2022selfd,lai2023xvogeneralizedvisualodometry}, to expand the diversity of training scenarios and further improve generalization. However, SSL also presents challenges, including noisy pseudo-labels and the risk of propagating errors through repetitive training cycles, which we address in our work through multimodal pseudo-label selection mechanisms.

\begin{figure*}[!t]
    \centering
      \includegraphics[trim={1cm 1cm 1.5cm 0.1cm},clip,width=6.7in]{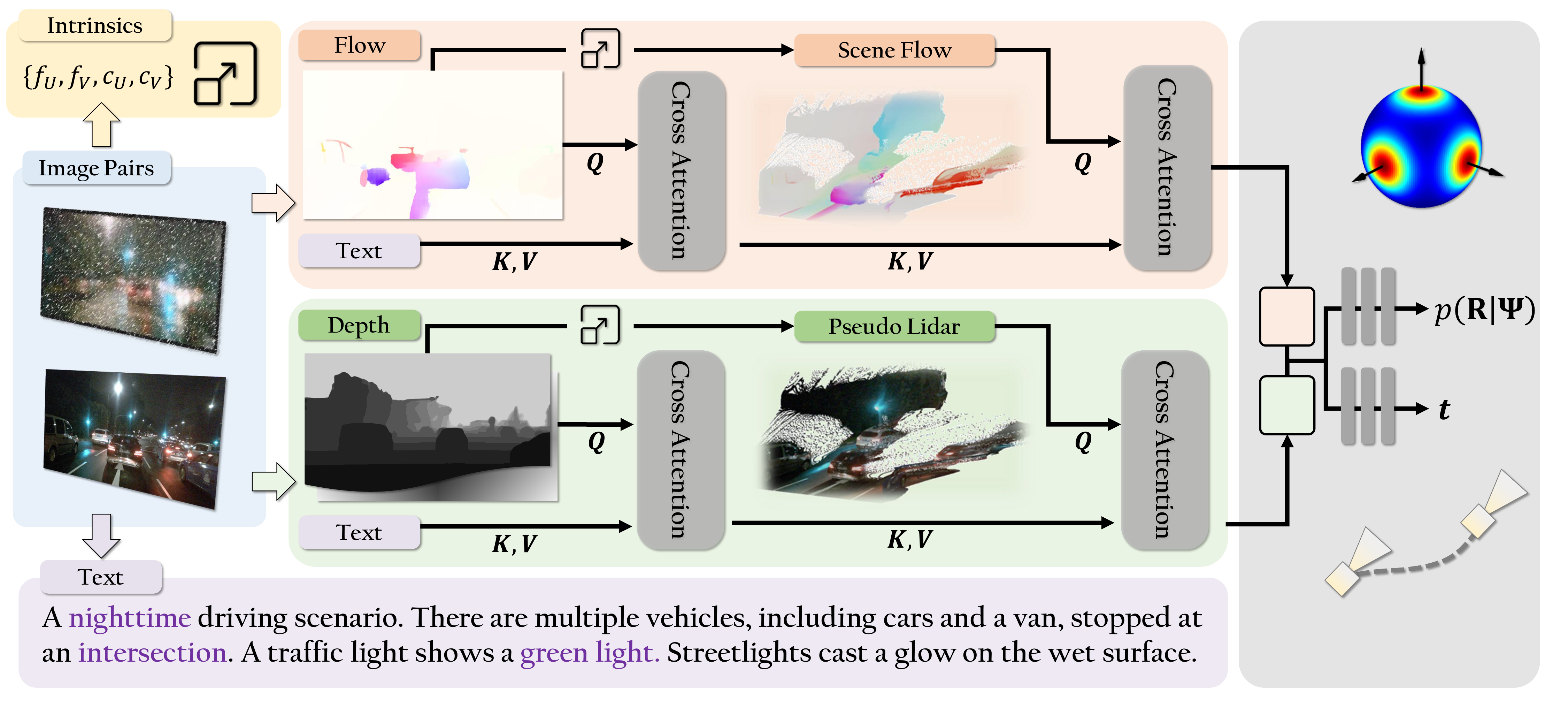}
    \vspace{-0.1cm}
    \caption{\textbf{Multimodal and Geometry-Guided Network Overview.}  
Given a pair of input images, our model computes a rich multimodal embedding through a transformer-based fusion module. The embedding is then passed to a two-branch decoder MLP that outputs real-world translation and rotation. Our architecture (Sec.~\ref{subsec:structure}) leverages cross-attention to fuse complementary cues, including flow, depth, camera intrinsics, and language-based features in a geometry-aware manner. 
The language prior is first used to refine both the depth map and 2D flow estimates. The refined depth is then unprojected into 3D (using estimated parameters) to compute scene flow, which is further enhanced and fused with additional features before decoding. By embedding geometric reasoning and multimodal priors directly into the network structure, our model achieves strong zero-shot generalization across diverse and challenging settings.}
    \label{fig:overview}
    \vspace{-0.2cm}
\end{figure*}

\section{Method}
\label{sec:method}

Our method (Fig.~\ref{fig:overview}) facilitates generalization via minimal and versatile image-based priors, integrated throughout our model structure. In this section, we first formalize our generalized, calibration-free monocular VO task. We then detail the proposed transformer-based geometry and prior-guided network structure in Sec.~\ref{subsec:structure} and the semi-supervised training process in Sec.~\ref{subsec:training}. 





\boldparagraph{Monocular VO with Minimal Assumptions}
In its most general form, monocular VO assumes two consecutive RGB frames $\cI = \{\bI_{i-1}, \bI_{i}\}$, $\bI \in \mathbb{R}^{W\times H\times 3}$ and learns to predict a real-world relative pose between the two camera views $\bT_i=[\bR_i|\bt_i]$, where $\bR_i\in \text{SO}(3), \bt_i\in \mathbb{R}^3$ are the relative rotation and translation, respectively. We focus on the efficient two-frame setup as it enables a fair comparison to other baselines methods (\eg,TartanVO~\cite{wang2020tartanvogeneralizablelearningbasedvo}) while quantifying \textit{real-time sequential drift}, \ie, prior to any additional global optimization steps, such as loop closure and
bundle adjustment~\cite{mur2017orb,tang2018ba,teed2021droid}. In Sec.~\ref{sec:analysis}, we find ZeroVO to outperform more complex methods that leverage computationally expensive, multi-frame refinement steps. We emphasize that monocular VO methods generally evaluate under up-to-scale settings~\cite{wang2020tartanvogeneralizablelearningbasedvo,teed2024deep,mur2017orb}, as estimating a metric-scaled transform from image pairs can be difficult, while reducing the solution space through known camera pose $\bT^{cam}_i$ and intrinsics, including the camera's focal length and center, $\{f_U, f_V, c_U, c_V\}$ (these are used in the camera intrinsic matrix, denoted as $\bK_i~\in~\mathbb{R}^{3\times3}$). However, in our formulation, we do not assume any prior knowledge of camera parameters, as it can be limiting and require re-calibration in cases of lens issues or different camera setups. Instead, to guide learning and inference, we rely on a set of versatile image-based priors built into the network structure. Specifically, we extract a rich set of modalities, including estimated flow $\hat{\bF}_i \in \mathbb{R}^{W\times H\times 2}$, depth map $\hat{\bD}_i \in \mathbb{R}^{W\times H}$, camera parameters $\hat{\bK}_i$, and rich language-based context features $\bZ_i^l \in \mathbb{R}^{W_l \times H_l}$ that provide complementary cues regarding scene semantics, layout characteristics, and scale. Our network structure fuses the estimated cues in a geometrically-guided process, discussed next. 




\subsection{Geometry and Prior-Guided Network}
\label{subsec:structure}

Our network structure comprises three key components: (1) an \textit{encoding module}, which estimates camera intrinsic parameters and extracts a rich, multimodal set of cues; (2) a text-conditional, \textit{geometry-guided transformer module} that leverages general structural priors to unproject data into 3D space and fuse the different modalities; and (3) a \textit{decoding module} for probabilistically predicting ego-motion.



\boldparagraph{Intrinsic Parameters Estimation}
VO methods generally rely on accurate knowledge of camera extrinsic and intrinsic parameters while training and testing on datasets with fixed camera settings. To enable more generalized VO, we do not rely on such restrictive assumptions. We instead propose to estimate the camera intrinsic parameters leveraging recent advances in in-the-wild, single-image intrinsic parameter estimation~\cite{zhu2024tame,hu2024metric3d} (primarily relying on 3D monocular priors). We leverage an off-the-shelf solution~\cite{zhu2024tame}, as we do not require the estimation to be completely accurate.
The intrinsic matrix will also be used to inform the geometry-aware transformer and semi-supervised network training (Sec.~\ref{subsec:training}). To align with image-level cues and enable the network to recover from noisy estimates, the intrinsic parameters are encoded into an image-sized array,
\begin{equation}
    \bI^{\hat{\bK}}(u,v) = \frac{\left| u-c_U \right|}{f_U} + \frac{\left| v-c_V \right|}{f_V}
\end{equation}
where the intrinsic information is explicitly preserved within each intrinsic map~\cite{wang2020tartanvogeneralizablelearningbasedvo}. Encoding parameter information into an image map provides an efficient approach for our transformer module to reason over noisy geometric information, as will be discussed below. We note that $\bI^{\hat{\bK}}$ uniquely represents a specific camera configuration.  


\boldparagraph{Extracting Multimodal Image Cues} 
To holistically represent general scene priors, scene dynamics, and camera motion and geometry, we employ a rich and complementary set of image-based features. As in standard VO methods, we extract optical flow~\cite{wang2020tartanvogeneralizablelearningbasedvo} from the image pair using a MaskFlownet~\cite{zhao2020maskflownet} encoder (
We extract the optical flow $\hat{\bF}$ as well as a correlation feature $\hat{\bF}^{c}$, which represents 2D correspondences between the images, from the intermediate layer of MaskFlownet). To estimate a metric-scale depth map $\hat{\bD}$, we utilize the estimated camera intrinsic parameters with Metric3Dv2~\cite{hu2024metric3d}. 
Finally, although camera information and metric depth can aid in understanding camera projection and motion, estimating these from a single image can be noisy and ill-posed. Thus, in addition to depth-based cues, we propose to leverage complementary text-based cues that can reduce ambiguity by capturing high-level scene semantics and layout characteristics. 
Specifically, we leverage LLaVA-NeXT~\cite{liu2024llavanext} to extract rich image descriptions which are encoded using Sentence Transformers~\cite{reimers-2019-sentence-bert}. In addition to providing useful context in arbitrary scenes during inference, we leverage the language-based cues to filter noisy pseudo-labels in Sec.~\ref{subsec:training}. 
We fuse modalities in a geometry-guided process, described next.  





\boldparagraph{Unprojection to Pseudo-3D}
The estimated depth map can be unprojected into a 3D point cloud $\bP \in \mathbb{R}^{W \times H \times 3}$ using the estimated camera matrix~\cite{wang2019pseudo}, \ie, by computing 3D world coordinates $\bp = d\hat{\bK}^{-1} \bu$, where $\bu = (u,v)$ is a pixel in homogeneous coordinate and $d = \hat{\bD}(\bu)$. We stack and normalize the resulting unprojection into a 3D array $\hat{\bD}^{\text{3D}}$.
We unproject the 2D optical flow into 3D to obtain a scene flow $\hat{\bF}^{\text{3D}}$ matrix (additional details regarding this step can be found in our supplementary). While these steps integrate physically-coherent camera and 3D information into a consistent representation, we expect the 3D maps to be noisy, particularly in our challenging generalization and adverse settings. Hence, instead of being explicit constraints, the 3D maps are integrated as minimal structures into a transformer-based module.  

\boldparagraph{Language and Geometry-Guided Transformer}
We employ transformer~\cite{dosovitskiy2020image,vaswani2017attention} to fuse the multimodal priors while reasoning over structure and noisy pseudo-3D information. We process the estimated flow and depth maps to compute two types of language-conditioned descriptors, a depth-based feature $\bZ^D$,
\begin{align}
    \bZ &= \text{CA}(\text{PE}([\hat{\bD},\bI^{\hat{\bK}}]),\bZ^l) \\
     \bZ^\bD &= \text{CA}(\text{PE}(\hat{\bD}^{3\text{D}}),\bZ) 
   \end{align}
and a flow-based feature $\bZ^F$ computed in a similar manner,
   \begin{align}
       \bZ &= \text{CA}(\text{PE}(\hat{\bF}^c),\bZ^l) \\
     \bZ^\bF &= \text{CA}(\text{PE}(\hat{\bF}^{3\text{D}}),\bZ) 
\end{align}
where $\text{CA}(\bQ,\bK\bV)$ denotes Cross-Attention, with query $Q$ and key-value pair $\bK\bV$, and PE denotes a patch and positional embedding~\cite{dosovitskiy2020image}. We note that we concatenate features with the intrinsic image to enable the model to learn coherence under noise, as accurate 3D reasoning is influenced by the focal length \cite{hu2024metric3d}.

\boldparagraph{Probabilistic Ego-Motion Decoder} 
The refined and aligned features, $\bZ^\bF$ and $\bZ^\bD$, are concatenated and decoded into ego-motion. Our decoder consists of two MLP output branches, one predicting translation and the other rotation. For translation, we leverage metric-scale regression~\cite{wang2020tartanvogeneralizablelearningbasedvo}. For rotation estimation, we fit a probabilistic distribution, specifically a matrix Fisher distribution (following~\cite{mardia2000directional,mohlin2020probabilistic,lai2023xvogeneralizedvisualodometry}) to model the rotation distribution in $\text{SO}(3)$. \begin{equation}
 p(\bR|\bPsi)=\frac{1}{c(\bPsi)}\exp(tr(\bPsi^\top \bR))
 \end{equation}
where $\bR \in \text{SO}(3)$ is the rotation matrix, $\bPsi\in \mathbb{R}^{3\times3}$ are the parameters of matrix Fisher distribution, and $c(\bPsi)$ is a normalization constant~\cite{mardia2000directional}. 

\subsection{Model Training via Semi-Supervision}
\label{subsec:training}

Due to the minimal assumptions employed by our calibration-free VO framework, the model can be effectively trained over in-the-wild, large-scale video collections. Hence, we consider both the standard supervised and a proposed semi-supervised training setup, detailed in this section. We employ the rich priors extracted from Sec.~\ref{subsec:structure} in the semi-supervised training to filter noisy pseudo-labeled samples.   

\boldparagraph{Supervised Training} 
Our model can be trained for a standard VO task, without requiring privileged information, \eg, ground-truth camera parameters, flow, or depth. We optimize the multi-head decoder MLP using Mean Squared Error (MSE) loss over predicted translation $\hat{\bt}$ and negative log-likelihood of rotation $\bR$ over the predicted distribution parameters $\hat{\bPsi}$,
\begin{equation}
    \mathcal{L} = \|\bt-\hat{\bt}\|_2^2 -\log(p(\bR|\hat{\bPsi}))
\end{equation}

While our supervised model already achieves strong performance, we further explore incorporating an additional training stage using pseudo-labeled samples generated by running the first-stage model on unlabeled data.


\boldparagraph{Generalization with Semi-Supervised Training} 
Our goal is to learn effective representations for generalized VO at scale. We thus investigate leveraging semi-supervised training to continue and update the model from unlabeled data. This training involves two stages, first with a supervised (\ie, teacher) model trained using the aforementioned objective function on an annotated dataset. Next, we sample pseudo-labels from the model~\cite{lee2013pseudo,caine2021pseudo,rizve2021defense} over a large unconstrained dataset collected from YouTube~\cite{yang2024generalized}, and re-train the model over the mixed annotated and pseudo-labeled dataset. Thus, the semi-supervised setup enables us to investigate the robustness and flexibility of our model in learning from diverse and challenging data with noisy supervision. While semi-supervised training has become a standard evaluation setup in computer vision~\cite{wang20213dioumatch,yalniz2019billion,li2019label,jeong2019consistency,souly2017semi,yang2021survey}, as in Sec.~\ref{subsec:structure} we explore the benefits of prior-informed mechanisms that can facilitate learning at scale from noisy examples.


\boldparagraph{Geometry-Guided Pseudo-Label Selection}   
To robustly learn from potentially noisy pseudo-labels, we employ a geometrical consistency error obtained based on estimated quantities. Specifically, motivated by prior work in unsupervised VO using known camera parameters~\cite{li2018undeepvo,mahjourian2018unsupervised,zhou2017unsupervised,zhan2018unsupervised,mahjourian2018unsupervised}, we warp a frame to the next frame with the estimated intrinsic matrix and ego-motion, 
$\bu_{i} = \hat{\bK}_{i}(d\hat{\bR_i}\hat{\bK}_{i-1}^{-1} \bu_{i-1} + \hat{\bt}_i$). We then employ a Structural Similarity Index Measure (SSIM) error~\cite{brunet2011mathematical} to quantify the similarity between an observed image $\bI_{i+1}$ and $\hat{\bI}_{i+1}$. To ensure that we capture diverse patterns of reconstruction challenges, we further normalize by the two-frame SSIM, \ie, 
\begin{equation}
  \text{normSSIM} = \frac{\text{SSIM}(\hat{\bI}_{i+1}, \bI_{i+1})}{\text{SSIM}(\bI_i, \bI_{i+1})}
\end{equation}
and exclude samples based on a fixed NormSSIM threshold. We note that SSIM assesses similarity by evaluating structural information, luminance, and contrast, thereby offering a perception-oriented measure of similarity in contrast to traditional measures based on pixel-wise errors. 



\boldparagraph{Language-Guided Pseudo-Label Selection}   
In addition to the geometry-based consistency pseudo-label check, we leverage our language-based module to filter redundant examples while maintaining an informative and diverse pseudo-labeled dataset. Although distinct text descriptions may not necessarily correspond to distinct pose transformations, we observe that two images characterized by nearly identical text descriptions are likely to be close in the visual space as well. To address sentence sequence variations within a paragraph, rather than serializing all text features into a single vector, we interpret the language feature as a subspace in a higher dimension. We leverage a subspace-based similarity over a short time window $H$, and compute the text feature similarity between the first image $\bI_i$ and the last image $\bI_{i+H}$ in the time window~\cite{knyazev2002principal}. Specifically,  
we compute similarity as:
\begin{equation}
      \text{subspace-sim} = \text{sin}\!\left( \text{arccos} ( \text{trace}(\Lambda) ) \right)^2
\end{equation}
where $\Lambda$ is the eigenvalues matrix obtained via Singular Value Decomposition over \( Q_i^\intercal Q_{i+H} \), the orthonormal matrices from the QR decompositions of text features \( \bZ^l_i \) and \( \bZ^l_{i+H} \). As in the geometric consistency selection, we remove sequences with low informativeness (\ie, high subspace-sim). The selection mechanism can thus help stabilize learning under the noisy and diverse pseudo-labels.

\subsection{Implementation Details}
In our implementation, we leverage the pre-trained WildCamera~\cite{zhu2024tame} model to estimate camera intrinsics. We utilize the MaskFlowNet encoder~\cite{zhao2020maskflownet} and Metric3Dv2~\cite{hu2024metric3d}, as flow and depth backbones, respectively. Sentence Transformers~\cite{reimers-2019-sentence-bert} is used to extract a $15\times768$ language-based feature matrix. For semi-supervised training, we follow prior work and collect a large-scale, unconstrained web video dataset for additional training~\cite{yang2024generalized,lai2023xvogeneralizedvisualodometry}. 
In our analysis, we present three model variants: ZeroVO, ZeroVO+, and LiteZeroVO+. ZeroVO serves as the default model in our experiments, while ZeroVO+ is further trained on the web video dataset with the proposed multimodal pseudo-label selection mechanism. LiteZeroVO+ shows a resource-constrained variant that omits the language-conditioned input modules by replacing the cross-attention module (for conditioning on the language cues and refining the estimated flow and depth maps) with self-attention. 
The training protocol remains consistent with that of the standard ZeroVO+. We train our network architecture using NVIDIA RTX 4090 GPU with a batch size of 16. ZeroVO+ achieves an inference speed of approximately 0.6 FPS, primarily constrained by the slower Lava-Next module (0.7 FPS), and LiteZeroVO+ obtains an inference speed of 5 FPS. Complete implementation and training details can be found in our supplementary.

\section{Experiments}
\label{sec:analysis}
\setlength{\tabcolsep}{3pt}

\begin{table*}[!t]
\centering
\caption{\textbf{Comparative Analysis Across Datasets.} We compare ZeroVO variants with existing baselines using standard metrics of translation, rotation, absolute trajectory, and scale errors. All methods are provided with estimated camera intrinsics and metric depth. 
ZeroVO+ is our model trained with further data using semi-supervision, and LiteZeroVO+ is a smaller model variant for resource-constrained settings. 
Our models demonstrate strong performance across metrics and datasets, particularly in metric translation estimation. As highlighted by the scale error, GTA and nuScenes contain challenging evaluation settings, including nighttime, weather variations, haze, and reflections. 
We note that TartanVO and DPVO baselines (in gray) only predict up-to-scale motion and use privileged information, \ie, ground-truth scale alignment in evaluation.
}
\label{tab:zero-shot}
{
\vspace{-0.3cm}
\resizebox{17.4cm}{!}{ 
\begin{tabular}{p{2.9cm}|cccc|cccc|cccc|cccc}
\toprule
\multirow{2}{*}{\textbf{Method}} & 
\multicolumn{4}{c|}{\textbf{KITTI 00-10}} &
\multicolumn{4}{c|}{\textbf{nuScenes}} &
\multicolumn{4}{c|}{\textbf{Argoverse}} &
\multicolumn{4}{c}{\textbf{GTA}} 
\\  \cline{2-17} 
        & $t_{err}$ &$r_{err}$ & $\text{ATE}$ & $s_{err}$
        & $t_{err}$ &$r_{err}$ & $\text{ATE}$ & $s_{err}$
        & $t_{err}$ &$r_{err}$ & $\text{ATE}$ & $s_{err}$
        & $t_{err}$ &$r_{err}$ & $\text{ATE}$ & $s_{err}$ \\ 
     \toprule     

    %
    
    {XVO~\cite{lai2023xvogeneralizedvisualodometry}} &16.82 &3.84 &168.43 &0.17 &12.75 &5.11 &8.30 &0.16 &9.13&4.86&5.70&0.12 & 25.56 &12.64 &28.02 &0.21 \\
    {M+DS~\cite{hu2024metric3d}} &14.22 &2.72 &154.77 &0.09  &17.08 &\textbf{1.46} &10.46 &0.18 &16.67&\textbf{1.79}&8.51&0.13  & 23.53 &10.38  &12.96  &0.26 \\
    \midrule
    {ZeroVO } &7.69 &2.72 &105.07 &0.07  &10.98 & 4.48& 6.79& 0.14 &6.83&3.13&4.10&0.11 &14.74 &10.63 &8.55 &0.17 \\
    {ZeroVO+ } &\textbf{6.81} &\textbf{2.69} &\textbf{104.69} &\textbf{0.06}   &\textbf{9.74} &4.37 &\textbf{6.03} &\textbf{0.12}  &\textbf{4.64}&2.83&\textbf{3.05}&\textbf{0.09} &\textbf{13.42} &\textbf{7.99} &\textbf{8.24} &\textbf{0.17} \\
    \midrule
    {LiteZeroVO+ } &8.85&2.90&118.54&0.08 &11.57&4.44&6.87&0.13 &7.65&3.82&5.28&0.11 &15.93&12.16&11.26&0.18 \\
   \midrule
   \textcolor{gray}{TartanVO~\cite{wang2020tartanvogeneralizablelearningbasedvo}} &\textcolor{gray}{13.85} &\textcolor{gray}{3.27} &\textcolor{gray}{103.07} & \textcolor{gray}{-} &\textcolor{gray}{10.27} &\textcolor{gray}{6.35} &\textcolor{gray}{6.26} & \textcolor{gray}{-} &\textcolor{gray}{11.17} &\textcolor{gray}{5.30} &\textcolor{gray}{7.03} & \textcolor{gray}{-} &\textcolor{gray}{10.56} &\textcolor{gray}{9.35} &\textcolor{gray}{3.82} & \textcolor{gray}{-}\\  
     \textcolor{gray}{DPVO~\cite{teed2024deep}} &\textcolor{gray}{8.31} &\textcolor{gray}{2.37} &\textcolor{gray}{78.53} & \textcolor{gray}{-} &\textcolor{gray}{4.34} &\textcolor{gray}{2.85} &\textcolor{gray}{2.66} & \textcolor{gray}{-} &\textcolor{gray}{2.66} &\textcolor{gray}{1.25} &\textcolor{gray}{1.59} & \textcolor{gray}{-} &\textcolor{gray}{12.65} &\textcolor{gray}{10.67} &\textcolor{gray}{4.33} & \textcolor{gray}{-}\\  
    \bottomrule
\end{tabular}}
}
\end{table*}

\setlength{\tabcolsep}{3.1pt}
\begin{table*}[!t]
\centering
\caption{\textbf{Ablation Analysis for Model and Training Components.} We analyze various model components: Flow module (\textbf{F}), Depth module (\textbf{D}), Language prior (\textbf{L}), Semi-supervised training (\textbf{S}), and Pseudo-label Selection (\textbf{P}). Flow, depth, and language correspond to the proposed supervised ZeroVO model. Results with additional semi-supervised training are shown as ZeroVO+ (showing state-of-the-art performance by integrating all of our proposed components). 
}
\label{tab:ablation}
\vspace{-0.3cm}
{
\begin{tabular}{ccccc|cccc|cccc|cccc|cccc}
\toprule
\multirow{2}{*}{\textbf{F}} & 
\multirow{2}{*}{\textbf{D}} & 
\multirow{2}{*}{\textbf{L}} & 
\multirow{2}{*}{\textbf{S}} & 
\multirow{2}{*}{\textbf{P}} & 
\multicolumn{4}{c|}{\textbf{KITTI 00-10}} &
\multicolumn{4}{c|}{\textbf{nuScenes}} &
\multicolumn{4}{c|}{\textbf{Argoverse}} &
\multicolumn{4}{c}{\textbf{GTA}}
\\  
\cline{6-21}
& & & & &
$t_{err}$ & $r_{err}$ & $\text{ATE}$ & $s_{err}$ &
$t_{err}$ & $r_{err}$ & $\text{ATE}$ & $s_{err}$ &
$t_{err}$ & $r_{err}$ & $\text{ATE}$ & $s_{err}$ &
$t_{err}$ & $r_{err}$ & $\text{ATE}$ & $s_{err}$ \\ 
\midrule
\checkmark&  &  &  &  &18.76 &5.49 &174.24 & 0.18 & 19.40 & 7.42 & 12.54 & 0.22 &12.23&6.34&9.42&0.20 &25.68 &15.52 &25.38 &0.25   \\
\checkmark&\checkmark  &  &  &  &8.99 &2.92 &123.42 &0.08  &12.26 &5.23 &8.40 &0.15 &8.62&4.11&5.71&0.11 &16.76 &12.75 &12.37 &0.19  \\
\checkmark&\checkmark  &\checkmark  &  &  &7.69 &2.72 &105.07 &0.07  &10.98 & 4.48& 6.79& 0.14 &6.83&3.13&4.10&0.11 &14.74 &10.63 &8.55 &0.17  \\
\checkmark& \checkmark &  \checkmark&  \checkmark&  &9.11 &2.88 &117.49 &0.08 &12.25 &5.39 &7.53 &0.14 &7.98&3.95&5.13&0.11 &16.49  &11.95  &10.27  &0.18 \\
\checkmark& \checkmark & \checkmark & \checkmark & \checkmark &\textbf{6.81} &\textbf{2.69} &\textbf{104.69} &\textbf{0.06}   &\textbf{9.74} &\textbf{4.37} &\textbf{6.03} &\textbf{0.12}  &\textbf{4.64}&\textbf{2.83}&\textbf{3.05}&\textbf{0.09} &\textbf{13.42} &\textbf{7.99} &\textbf{8.24} &\textbf{0.17}  \\
\bottomrule
\end{tabular}}
\vspace{-0.3cm}
\end{table*} 

\setlength{\tabcolsep}{3pt}

\subsection{Experimental Setup}

\boldparagraph{Real-World Datasets} To study the generalization ability of our model, we conduct experiments using five datasets including three widely adopted datasets for autonomous driving: nuScenes~\cite{nuscenes2019}, KITTI~\cite{Geiger2012CVPR}, and Argoverse 2~\cite{wilson2023argoverse2generationdatasets}, as well as an introduced Grand Theft Auto V (GTA) simulated dataset with challenging environmental and lens conditions. nuScenes covers four distinct regions across Boston and Singapore: Boston-Seaport, Singapore-OneNorth, Singapore-Queenstown, and Singapore-Holland Village. It encompasses various challenging conditions, such as heavy traffic, nighttime driving, and scenarios involving strong light reflections, making nuScenes particularly valuable for assessing the robustness of models under diverse and complex real-world conditions. In our evaluation, we train on a subset of nuScenes, and test on other benchmarks in a zero-shot manner. KITTI is the most widely evaluated dataset in the VO task. Specifically, the camera intrinsics in KITTI differ significantly from those of the other three benchmarks, making it an important dataset for evaluating a model's ability to adapt to varying camera configurations. Argoverse 2 collects data from six distinct U.S. cities and encompasses a wide range of weather conditions and driving scenarios. Notably, the dataset includes grayscale images captured by the stereo front camera, which provides another generalization stress-test for the model. We also follow Lai~\etal~\cite{lai2023xvogeneralizedvisualodometry} and leverage online driving videos from YouTube, encompassing footage across multiple cities, including urban areas, villages, national parks, mountainous regions, and coastal areas, under a wide range of weather conditions. This dataset enables us to study the benefits of diverse unlabeled data while providing an ideal environment for the model to self-learn numerous variations induced by camera motions.

\boldparagraph{GTA Dataset}
Besides the three public datasets, we introduce a newly generated simulated dataset derived from the high-fidelity, GTA simulation. 
Our GTA dataset consists of 922 driving sequences captured within a simulated city environment, encompassing a range of diverse weather conditions, driving speeds (particularly high-speed maneuvers not found in other public datasets), traffic scenarios, and times of day. Compared to other commonly used open-source simulation platforms such as CARLA~\cite{dosovitskiy2017carla}, GTA offers several key advantages: (1) enhanced image realism through the application of the reshade graphic settings that support higher quality rendering, and (2) a wider variety of road conditions across various weather scenarios. For on-road driving, these conditions include significant uphill and downhill gradients, tunnels, and underground parking facilities; for off-road driving, the environment features mountains, deserts, snow-covered terrains, and forests, thereby enabling more precise and complex rotational dynamics throughout the map.

\boldparagraph{Experimental Setting} Similar to XVO~\cite{lai2023xvogeneralizedvisualodometry}, our framework is trained on data from a single city in the nuScenes dataset. Unlike XVO, we observed that Boston-Seaport, Singapore-Queenstown, and Singapore-Holland Village contain the majority of challenging conditions, such as rain, nighttime driving, light reflections, and heavy traffic. Therefore, we use Singapore-OneNorth as our supervised training dataset and the remaining regions, KITTI, Argovere 2, and GTA, as test datasets. It is important to note the main evaluation is done on datasets that were unseen by our model during training and without assumed camera parameters. 

\boldparagraph{Baselines} We compared the four most related baselines that demonstrate generalization across datasets without requiring additional fine-tuning: TartanVO~\cite{wang2020tartanvogeneralizablelearningbasedvo}, XVO~\cite{lai2023xvogeneralizedvisualodometry}, DPVO~\cite{teed2024deep}, and Metric3D+Droid-SLAM (M+DS)~\cite{teed2021droid, hu2024metric3d}. TartanVO employs effective random cropping and resizing techniques to simulate diverse camera configurations, thereby enhancing the generalization of rotation estimation across unseen datasets. XVO leverages a multi-modality architecture to implicitly extract richer spatial features and integrates self-training to achieve robust generalization performance in both rotation estimation and real-world scale recovery. DPVO employs a recurrent update operator for patch-based correspondence, complemented by differentiable bundle adjustment, demonstrating strong zero-shot performance in rotation estimation. M+DS utilizes the generalization capabilities of Metric3D v2 and Droid-SLAM to accurately estimate metric depth and rotation, effectively recovering the motion trajectory at a real-world scale. Our main baseline is M+DS which achieves state-of-the-art generalization results across dataset.

\boldparagraph{Metrics} 
To provide a comprehensive analysis of the results, we utilize Translation Error ($t_{err}$), Rotation Error ($r_{err}$), Absolute Trajectory Error ($\text{ATE}$), and Scale Error ($s_{err}$)~\cite{Geiger2012CVPR, lai2023xvogeneralizedvisualodometry}. $t_{err}$ and $r_{err}$ compute the average translation error (\%) and rotation error ($^\circ$/100\, \text{m}) across all possible subsequences within a test sequence with lengths ranging from 100 to 800 meters. ATE measures the deviation between the estimated trajectory and the ground-truth trajectory by comparing the positions of corresponding poses, making it an effective metric for measuring drift over time. The scale error ($s_{err}$) measures the average discrepancy between the predicted translation and the ground truth translation. Combined with rotation error ($r_{err}$) and Absolute Trajectory Error ($\text{ATE}$), it allows us to effectively determine whether accumulated drift is attributed to scale inaccuracies or rotational deviations.

\subsection{Results}

\boldparagraph{Generalization Performance} To examine the generalization ability of our model, we evaluate it on entire sequences on KITTI, the unseen regions in nuScenes, and the simulated dataset GTA. Table~\ref{tab:zero-shot} compares ZeroVO+ with prior baselines in a zero-shot setting. For a fair comparison of the zero-shot performance, all models are provided with the same estimated camera intrinsics and metric depth (if required). TartanVO and DPVO can only estimate rotation and require scale alignment with ground-truth translation to reconstruct the trajectory at a real-world scale. From the results in Table~\ref{tab:zero-shot}, our model achieves superior performance across nearly all metrics on the four datasets. It is important to note that sequences on KITTI are significantly longer compared to those in other datasets, making them more prone to accumulating large drift (\ie, high ATE). Our method accurately predicts rotation and translation scale on KITTI, resulting in the lowest ATE among all baselines, even without incorporating multi-frame temporal optimization. The results on the GTA dataset further demonstrate the strong generalization capability of our model, achieving ATE results comparable to scale-aligned DPVO, which leverages privileged evaluation. 
\begin{figure*}[!t]
    \centering
   \adjustbox{center=\textwidth}{ \hspace{-0.5cm} \includegraphics[width=6.7in]{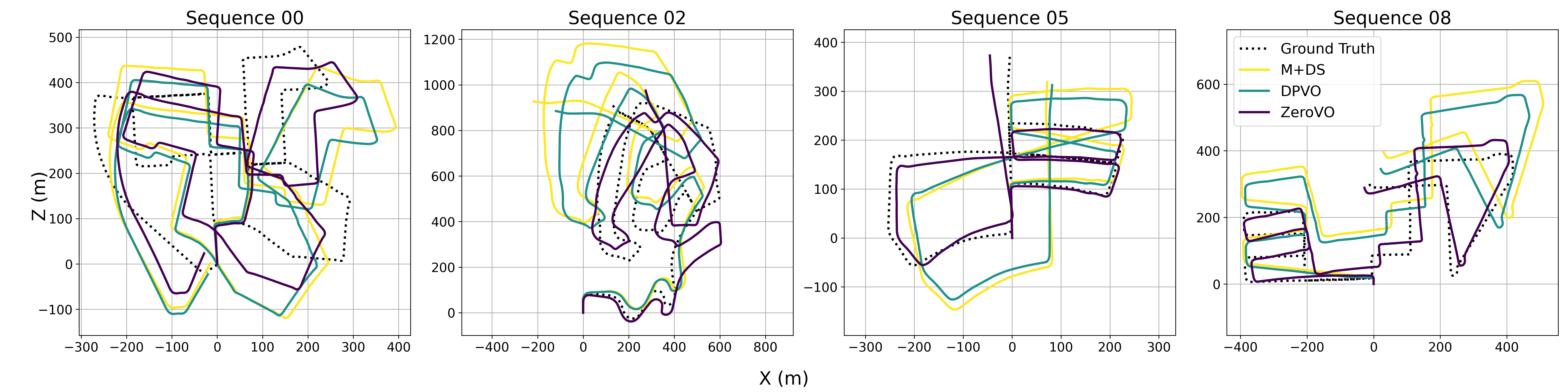}}
   \vspace{-0.6cm}
    \caption{\textbf{Qualitative Results on KITTI.} We show trajectory prediction results across the four most complex driving sequences (00, 02, 05, and 08) from the KITTI dataset. Each subplot illustrates the trajectories generated by our proposed model and the baseline models alongside the ground truth trajectory. The qualitative results demonstrate that our approach achieves the highest alignment with the ground truth, particularly in challenging turns and extended straight paths. These findings highlight the robustness of our method in handling complex and diverse driving scenarios.}
\label{fig:traj}
\vspace{-0.3cm}
\end{figure*}
In Table \ref{tab:weather}, we divide the remaining regions in nuScenes into different subsets based on various weather conditions: day, night, rain, and light. The strong light scenario is caused by severe light reflections. We find that night and strong light conditions present the most challenging scenarios, as it is difficult for the model to detect and extract valuable information. We demonstrate that our model achieves the best performance across all conditions, highlighting its robustness against external noise.

\boldparagraph{Ablation Study}
In Table~\ref{tab:ablation}, we study the roles of each module in our model structure. We begin by analyzing the impact of our depth module. When the model is equipped with only the flow module, the model struggles to generalize to unseen scenarios, particularly in terms of scale estimation. This outcome is expected, as predicting scale from a single image without any additional context is an ill-posed problem. By incorporating the depth module, the model demonstrates improvements across all metrics, particularly in scale estimation. This improvement indicates that by concatenating the estimated metric depth with the intrinsic image, the model can effectively learn coherent 3D spatial information, even in the presence of noise, and accurately estimate scale.  It is also noteworthy that the depth module improves rotation estimation performance. This demonstrates that leveraging both depth and optical flow to unproject 3D scene flow provides crucial 3D correspondence information that leads to improved rotation estimation. The experiment with textual information further demonstrates the model's robustness against noise. Under challenging driving conditions, such as numerous dynamic objects, darkness, strong light reflections, rain, and fog, the estimated camera intrinsics and metric depth are highly susceptible to becoming unreliable. The general text description is able to provide extra 3D information, such as object layouts and movements, which helps the model maintain robustness in highly noisy environments. At last, we demonstrate the effectiveness of our semi-supervision approach using pseudo-label selection. Without pseudo-label selection, we observe a drop in the model's performance compared to the supervised trained model. This decline is due to the introduction of excessive pseudo-labeled examples with redundancy and uncertain label quality, which hinders model training. Our pseudo-label selection process effectively filters out highly redundant and low-quality pseudo-labeled examples, achieving the best performance among all zero-shot metric-scale models. Further ablations and analysis can be found in our supplementary.

\begin{table}[!t]
\centering
\caption{\textbf{Condition Breakdown on nuScenes.} We show results breakdown (ATE) over scenes categorized by weather and lens settings. We sample from nuScenes the Day, Night, and Rainy scenes, along with particularly challenging frames that include severe light reflections. Our ZeroVO+ model performs best overall. We note that TartanVO and DPVO baselines only predict up-to-scale motion and use ground-truth scale alignment in inference.}
\label{tab:weather}
\vspace{-0.2cm}
{
\begin{tabular}{p{2.3cm}|c|c|c|c}
\toprule
\multirow{1}{*}{\textbf{Method}} & 
\multicolumn{1}{c|}{\makebox[1.2cm][c]{\textbf{Day}}} &
\multicolumn{1}{c|}{\makebox[1.2cm][c]{\textbf{Night}}} &
\multicolumn{1}{c|}{\makebox[1.2cm][c]{\textbf{Rainy}}} &
\multicolumn{1}{c}{\makebox[1.2cm][c]{\textbf{Light}}} 
\\   
    \toprule  
    {XVO~\cite{lai2023xvogeneralizedvisualodometry}} &  6.61 &  14.41 &  15.99 &  15.73\\  
    {M+DS~\cite{hu2024metric3d}} &  6.08 &  17.19 &  17.49 &  18.54\\  
    \midrule
    {{ZeroVO}} &  3.90 &  10.33 &  12.63 &  13.33\\  
    {ZeroVO+} &  \textbf{3.60} &  \textbf{10.26} &  \textbf{10.10} &  \textbf{11.15}\\  

    
    \bottomrule
\end{tabular}}
\vspace{-0.5cm}
\end{table}

\boldparagraph{Qualitative Analysis} Fig.~\ref{fig:traj} depicts the most complex and longest trajectories on KITTI, compared with the two best-performing baselines. The trajectory of DPVO is aligned with the ground-truth translation after scale adjustment. Therefore, it is straightforward to see how inaccurate rotation estimation results in drift accumulation. A comparison between the results of DPVO and M+DS reveals how inaccuracies in translation estimation further exacerbate drift accumulation. By leveraging general textual information and unprojecting 2D data into 3D space, our model effectively extracts more accurate and inherent correspondence features, which enhance robustness even when the estimated depth or camera intrinsics are noisy.


\section{Conclusion}

We introduced ZeroVO, a novel transformer-based framework designed to tackle the challenge of visual odometry generalization under adverse and unseen conditions. ZeroVO integrates rich multimodal cues—spanning geometry, language, and vision—within a unified architecture to enhance robustness and adaptability in complex environments. Its camera-agnostic design, combined with a semi-supervised training paradigm, enables effective handling of noisy data and seamless adaptation to novel scenarios. Extensive evaluation across diverse and challenging benchmarks demonstrates that ZeroVO establishes a new standard for zero-shot VO performance, underscoring its promise for real-world deployment without the need for camera recalibration or domain-specific tuning.

\section{Acknowledgments}
We thank the Red Hat Collaboratory (awards \#2024-01-RH02, \#2024-01-RH07) and National Science Foundation (IIS-2152077) for supporting this research.


{
    \small
    \bibliographystyle{ieeenat_fullname}
    \bibliography{main}

\begin{thebibliography}{104}
\providecommand{\natexlab}[1]{#1}
\providecommand{\url}[1]{\texttt{#1}}
\expandafter\ifx\csname urlstyle\endcsname\relax
  \providecommand{\doi}[1]{doi: #1}\else
  \providecommand{\doi}{doi: \begingroup \urlstyle{rm}\Url}\fi

\bibitem[Agarwal et~al.(2014)Agarwal, Maturana, and Scherer]{agarwal2014visual}
Aditya Agarwal, Daniel Maturana, and Sebastian Scherer.
\newblock Visual odometry in smoke occluded environments.
\newblock \emph{Robotics Institute, Carnegie Mellon University, Pittsburgh, PA, Tech. Rep. CMU-RI-TR-15-07}, 2014.

\bibitem[Antol et~al.(2015)Antol, Agrawal, Lu, Mitchell, Batra, Zitnick, and Parikh]{antol2015vqa}
Stanislaw Antol, Aishwarya Agrawal, Jiasen Lu, Margaret Mitchell, Dhruv Batra, C~Lawrence Zitnick, and Devi Parikh.
\newblock Vqa: Visual question answering.
\newblock In \emph{CVPR}, 2015.

\bibitem[Azarbayejani and Pentland(1995)]{azarbayejani1995recursive}
Ali Azarbayejani and Alex~P Pentland.
\newblock Recursive estimation of motion, structure, and focal length.
\newblock \emph{PAMI}, 1995.

\bibitem[Berthelot et~al.(2019{\natexlab{a}})Berthelot, Carlini, Cubuk, Kurakin, Sohn, Zhang, and Raffel]{berthelot2019remixmatch}
David Berthelot, Nicholas Carlini, Ekin~D Cubuk, Alex Kurakin, Kihyuk Sohn, Han Zhang, and Colin Raffel.
\newblock Remixmatch: Semi-supervised learning with distribution alignment and augmentation anchoring.
\newblock \emph{arXiv preprint arXiv:1911.09785}, 2019{\natexlab{a}}.

\bibitem[Berthelot et~al.(2019{\natexlab{b}})Berthelot, Carlini, Goodfellow, Papernot, Oliver, and Raffel]{berthelot2019mixmatch}
David Berthelot, Nicholas Carlini, Ian Goodfellow, Nicolas Papernot, Avital Oliver, and Colin~A Raffel.
\newblock Mixmatch: A holistic approach to semi-supervised learning.
\newblock \emph{NeurIPS}, 2019{\natexlab{b}}.

\bibitem[Bian et~al.(2019)Bian, Li, Wang, Zhan, Shen, Cheng, and Reid]{bian2019unsupervised}
Jiawang Bian, Zhichao Li, Naiyan Wang, Huangying Zhan, Chunhua Shen, Ming-Ming Cheng, and Ian Reid.
\newblock Unsupervised scale-consistent depth and ego-motion learning from monocular video.
\newblock In \emph{NeurIPS}, 2019.

\bibitem[Brunet et~al.(2011)Brunet, Vrscay, and Wang]{brunet2011mathematical}
Dominique Brunet, Edward~R Vrscay, and Zhou Wang.
\newblock On the mathematical properties of the structural similarity index.
\newblock \emph{T-IP}, 2011.

\bibitem[Burri et~al.(2016)Burri, Nikolic, Gohl, Schneider, Rehder, Omari, Achtelik, and Siegwart]{burri2016euroc}
Michael Burri, Janosch Nikolic, Pascal Gohl, Thomas Schneider, Joern Rehder, Sammy Omari, Markus~W Achtelik, and Roland Siegwart.
\newblock The euroc micro aerial vehicle datasets.
\newblock \emph{IJRR}, 2016.

\bibitem[Cadena et~al.(2016)Cadena, Carlone, Carrillo, Latif, Scaramuzza, Neira, Reid, and Leonard]{cadena2016past}
Cesar Cadena, Luca Carlone, Henry Carrillo, Yasir Latif, Davide Scaramuzza, Jos{\'e} Neira, Ian Reid, and John~J Leonard.
\newblock Past, present, and future of simultaneous localization and mapping: Toward the robust-perception age.
\newblock \emph{T-RO}, 2016.

\bibitem[Caesar et~al.(2020)Caesar, Bankiti, Lang, Vora, Liong, Xu, Krishnan, Pan, Baldan, and Beijbom]{nuscenes2019}
Holger Caesar, Varun Bankiti, Alex~H Lang, Sourabh Vora, Venice~Erin Liong, Qiang Xu, Anush Krishnan, Yu Pan, Giancarlo Baldan, and Oscar Beijbom.
\newblock nuscenes: A multimodal dataset for autonomous driving.
\newblock In \emph{CVPR}, 2020.

\bibitem[Caine et~al.(2021)Caine, Roelofs, Vasudevan, Ngiam, Chai, Chen, and Shlens]{caine2021pseudo}
Benjamin Caine, Rebecca Roelofs, Vijay Vasudevan, Jiquan Ngiam, Yuning Chai, Zhifeng Chen, and Jonathon Shlens.
\newblock Pseudo-labeling for scalable 3d object detection.
\newblock In \emph{arXiv preprint arXiv:2103.02093}, 2021.

\bibitem[Campos et~al.(2021)Campos, Elvira, Rodr{\'\i}guez, Montiel, and Tard{\'o}s]{campos2021orb}
Carlos Campos, Richard Elvira, Juan J~G{\'o}mez Rodr{\'\i}guez, Jos{\'e}~MM Montiel, and Juan~D Tard{\'o}s.
\newblock Orb-slam3: An accurate open-source library for visual, visual--inertial, and multimap slam.
\newblock \emph{T-RO}, 2021.

\bibitem[Ceccarelli and Secci(2022)]{ceccarelli2022rgb}
Andrea Ceccarelli and Francesco Secci.
\newblock Rgb cameras failures and their effects in autonomous driving applications.
\newblock \emph{T-DSC}, 2022.

\bibitem[Chen et~al.(2023)Chen, Spiridonova, Yang, Gao, and Li]{chen2023llavainteractiveallinonedemoimage}
Wei-Ge Chen, Irina Spiridonova, Jianwei Yang, Jianfeng Gao, and Chunyuan Li.
\newblock Llava-interactive: An all-in-one demo for image chat, segmentation, generation and editing.
\newblock \emph{arXiv preprint arXiv:2311.00571}, 2023.

\bibitem[Chiuso et~al.(2002)Chiuso, Favaro, Jin, and Soatto]{chiuso2002structure}
Alessandro Chiuso, Paolo Favaro, Hailin Jin, and Stefano Soatto.
\newblock Structure from motion causally integrated over time.
\newblock \emph{PAMI}, 2002.

\bibitem[Clark et~al.(2018)Clark, Luong, Manning, and Le]{clark2018semi}
Kevin Clark, Minh-Thang Luong, Christopher~D Manning, and Quoc~V Le.
\newblock Semi-supervised sequence modeling with cross-view training.
\newblock \emph{arXiv preprint arXiv:1809.08370}, 2018.

\bibitem[Dai and Lin(2017)]{dai2017contrastive}
Bo Dai and Dahua Lin.
\newblock Contrastive learning for image captioning.
\newblock \emph{NeurIPS}, 30, 2017.

\bibitem[Dickmanns(2007)]{dickmanns2007dynamic}
Ernst~Dieter Dickmanns.
\newblock \emph{Dynamic vision for perception and control of motion}.
\newblock Springer, 2007.

\bibitem[Dosovitskiy et~al.(2017)Dosovitskiy, Ros, Codevilla, Lopez, and Koltun]{dosovitskiy2017carla}
Alexey Dosovitskiy, German Ros, Felipe Codevilla, Antonio Lopez, and Vladlen Koltun.
\newblock Carla: An open urban driving simulator.
\newblock In \emph{CoRL}, 2017.

\bibitem[Dosovitskiy et~al.(2021)Dosovitskiy, Beyer, Kolesnikov, Weissenborn, Zhai, Unterthiner, Dehghani, Minderer, Heigold, Gelly, Uszkoreit, and Houlsby]{dosovitskiy2020image}
Alexey Dosovitskiy, Lucas Beyer, Alexander Kolesnikov, Dirk Weissenborn, Xiaohua Zhai, Thomas Unterthiner, Mostafa Dehghani, Matthias Minderer, Georg Heigold, Sylvain Gelly, Jakob Uszkoreit, and Neil Houlsby.
\newblock An image is worth 16x16 words: Transformers for image recognition at scale.
\newblock \emph{ICLR}, 2021.

\bibitem[Engel et~al.(2017)Engel, Koltun, and Cremers]{engel2017direct}
Jakob Engel, Vladlen Koltun, and Daniel Cremers.
\newblock Direct sparse odometry.
\newblock In \emph{PAMI}, 2017.

\bibitem[Fraundorfer and Scaramuzza(2011)]{fraundorfer2011visual}
Friedrich Fraundorfer and Davide Scaramuzza.
\newblock Visual odometry: Part i: The first 30 years and fundamentals.
\newblock \emph{RAM}, 2011.

\bibitem[Gao et~al.(2019)Gao, Wang, Dai, Li, and Nevatia]{gao2019note}
Jiyang Gao, Jiang Wang, Shengyang Dai, Li-Jia Li, and Ram Nevatia.
\newblock Note-rcnn: Noise tolerant ensemble rcnn for semi-supervised object detection.
\newblock In \emph{CVPR}, 2019.

\bibitem[Geiger et~al.(2012)Geiger, Lenz, and Urtasun]{Geiger2012CVPR}
Andreas Geiger, Philip Lenz, and Raquel Urtasun.
\newblock Are we ready for autonomous driving? the kitti vision benchmark suite.
\newblock In \emph{CVPR}, 2012.

\bibitem[Guizilini et~al.(2022)Guizilini, Lee, Ambru{\c{s}}, and Gaidon]{guizilini2022learning}
Vitor Guizilini, Kuan-Hui Lee, Rare{\c{s}} Ambru{\c{s}}, and Adrien Gaidon.
\newblock Learning optical flow, depth, and scene flow without real-world labels.
\newblock \emph{RA-L}, 2022.

\bibitem[Guizilini et~al.(2023)Guizilini, Vasiljevic, Chen, Ambruș, and Gaidon]{guizilini2023towards}
Vitor Guizilini, Igor Vasiljevic, Dian Chen, Rareș Ambruș, and Adrien Gaidon.
\newblock Towards zero-shot scale-aware monocular depth estimation.
\newblock In \emph{CVPR}, 2023.

\bibitem[Gururangan et~al.(2019)Gururangan, Dang, Card, and Smith]{gururangan2019variational}
Suchin Gururangan, Tam Dang, Dallas Card, and Noah~A Smith.
\newblock Variational pretraining for semi-supervised text classification.
\newblock \emph{arXiv preprint arXiv:1906.02242}, 2019.

\bibitem[Gururangan et~al.(2020)Gururangan, Marasovi{\'c}, Swayamdipta, Lo, Beltagy, Downey, and Smith]{gururangan2020don}
Suchin Gururangan, Ana Marasovi{\'c}, Swabha Swayamdipta, Kyle Lo, Iz Beltagy, Doug Downey, and Noah~A Smith.
\newblock Don't stop pretraining: Adapt language models to domains and tasks.
\newblock \emph{arXiv preprint arXiv:2004.10964}, 2020.

\bibitem[Hirose et~al.(2019)Hirose, Xia, Mart{\'\i}n-Mart{\'\i}n, Sadeghian, and Savarese]{hirose2019deep}
Noriaki Hirose, Fei Xia, Roberto Mart{\'\i}n-Mart{\'\i}n, Amir Sadeghian, and Silvio Savarese.
\newblock Deep visual mpc-policy learning for navigation.
\newblock \emph{RA-L}, 4\penalty0 (4), 2019.

\bibitem[Hu et~al.(2024)Hu, Yin, Zhang, Cai, Long, Chen, Wang, Yu, Shen, and Shen]{hu2024metric3d}
Mu Hu, Wei Yin, Chi Zhang, Zhipeng Cai, Xiaoxiao Long, Hao Chen, Kaixuan Wang, Gang Yu, Chunhua Shen, and Shaojie Shen.
\newblock Metric3d v2: A versatile monocular geometric foundation model for zero-shot metric depth and surface normal estimation.
\newblock \emph{PAMI}, 2024.

\bibitem[Jang and Cho(2020)]{jang2020generalized}
Young~Kyun Jang and Nam~Ik Cho.
\newblock Generalized product quantization network for semi-supervised image retrieval.
\newblock In \emph{CVPR}, 2020.

\bibitem[Jeong et~al.(2019)Jeong, Lee, Kim, and Kwak]{jeong2019consistency}
Jisoo Jeong, Seungeui Lee, Jeesoo Kim, and Nojun Kwak.
\newblock Consistency-based semi-supervised learning for object detection.
\newblock In \emph{NeurIPS}, 2019.

\bibitem[Kanai et~al.(2024)Kanai, Vasiljevic, Guizilini, and Shintani]{kanai2024self}
Takayuki Kanai, Igor Vasiljevic, Vitor Guizilini, and Kazuhiro Shintani.
\newblock Self-supervised geometry-guided initialization for robust monocular visual odometry.
\newblock \emph{arXiv preprint arXiv:2406.00929}, 2024.

\bibitem[Kaygusuz et~al.(2021)Kaygusuz, Mendez, and Bowden]{kaygusuz2021mdn}
Nimet Kaygusuz, Oscar Mendez, and Richard Bowden.
\newblock Mdn-vo: Estimating visual odometry with confidence.
\newblock In \emph{IROS}, 2021.

\bibitem[Kendall and Cipolla(2017)]{kendall2017geometric}
Alex Kendall and Roberto Cipolla.
\newblock Geometric loss functions for camera pose regression with deep learning.
\newblock In \emph{CVPR}, 2017.

\bibitem[Kendall et~al.(2015)Kendall, Grimes, and Cipolla]{kendall2015posenet}
Alex Kendall, Matthew Grimes, and Roberto Cipolla.
\newblock Posenet: A convolutional network for real-time 6-dof camera relocalization.
\newblock In \emph{ICCV}, 2015.

\bibitem[Kim and Ohn-Bar(2024)]{kim2024motion}
Hee~Jae Kim and Eshed Ohn-Bar.
\newblock Motion diversification networks.
\newblock In \emph{CVPR}, 2024.

\bibitem[Kim et~al.(2015)Kim, Lim, and Kim]{kim2015robust}
Pyojin Kim, Hyon Lim, and H~Jin Kim.
\newblock Robust visual odometry to irregular illumination changes with rgb-d camera.
\newblock In \emph{IROS}, 2015.

\bibitem[Knyazev and Argentati(2002)]{knyazev2002principal}
Andrew~V Knyazev and Merico~E Argentati.
\newblock Principal angles between subspaces in an a-based scalar product: algorithms and perturbation estimates.
\newblock \emph{SIAM Journal on Scientific Computing}, 2002.

\bibitem[Lai et~al.(2023)Lai, Shangguan, Zhang, and Ohn-Bar]{lai2023xvogeneralizedvisualodometry}
Lei Lai, Zhongkai Shangguan, Jimuyang Zhang, and Eshed Ohn-Bar.
\newblock {XVO}: Generalized visual odometry via cross-modal self-training.
\newblock In \emph{ICCV}, 2023.

\bibitem[Lai et~al.(2024)Lai, Ohn-Bar, Arora, and Yi]{lai2024uncertainty}
Lei Lai, Eshed Ohn-Bar, Sanjay Arora, and John Seon~Keun Yi.
\newblock Uncertainty-guided never-ending learning to drive.
\newblock In \emph{CVPR}, 2024.

\bibitem[Lee et~al.(2013)]{lee2013pseudo}
Dong-Hyun Lee et~al.
\newblock Pseudo-label: The simple and efficient semi-supervised learning method for deep neural networks.
\newblock In \emph{ICMLW}, 2013.

\bibitem[Li et~al.(2019)Li, Wu, Liu, Zhang, and Guan]{li2019label}
Qimai Li, Xiao-Ming Wu, Han Liu, Xiaotong Zhang, and Zhichao Guan.
\newblock Label efficient semi-supervised learning via graph filtering.
\newblock In \emph{CVPR}, 2019.

\bibitem[Li et~al.(2018)Li, Wang, Long, and Gu]{li2018undeepvo}
Ruihao Li, Sen Wang, Zhiqiang Long, and Dongbing Gu.
\newblock Undeepvo: Monocular visual odometry through unsupervised deep learning.
\newblock In \emph{ICRA}, 2018.

\bibitem[Li et~al.(2020)Li, Wang, Cao, Xue, Yan, and Zha]{li2020self}
Shunkai Li, Xin Wang, Yingdian Cao, Fei Xue, Zike Yan, and Hongbin Zha.
\newblock Self-supervised deep visual odometry with online adaptation.
\newblock In \emph{CVPR}, 2020.

\bibitem[Lipson et~al.(2024)Lipson, Teed, and Deng]{lipson2024deeppatchvisualslam}
Lahav Lipson, Zachary Teed, and Jia Deng.
\newblock Deep patch visual slam.
\newblock In \emph{ECCV}, 2024.

\bibitem[Liu et~al.(2024{\natexlab{a}})Liu, Li, Li, and Lee]{liu2024improved}
Haotian Liu, Chunyuan Li, Yuheng Li, and Yong~Jae Lee.
\newblock Improved baselines with visual instruction tuning.
\newblock In \emph{CVPR}, 2024{\natexlab{a}}.

\bibitem[Liu et~al.(2024{\natexlab{b}})Liu, Li, Li, Li, Zhang, Shen, and Lee]{liu2024llavanext}
Haotian Liu, Chunyuan Li, Yuheng Li, Bo Li, Yuanhan Zhang, Sheng Shen, and Yong~Jae Lee.
\newblock Llava-next: Improved reasoning, ocr, and world knowledge, 2024{\natexlab{b}}.

\bibitem[Liu et~al.(2024{\natexlab{c}})Liu, Li, Wu, and Lee]{liu2024visual}
Haotian Liu, Chunyuan Li, Qingyang Wu, and Yong~Jae Lee.
\newblock Visual instruction tuning.
\newblock \emph{NeurIPS}, 2024{\natexlab{c}}.

\bibitem[Liu et~al.(2023)Liu, Cheng, Liu, Zhang, Li, Ren, Zou, Yang, Su, Zhu, et~al.]{liu2023llavapluslearningusetools}
Shilong Liu, Hao Cheng, Haotian Liu, Hao Zhang, Feng Li, Tianhe Ren, Xueyan Zou, Jianwei Yang, Hang Su, Jun Zhu, et~al.
\newblock Llava-plus: Learning to use tools for creating multimodal agents.
\newblock \emph{arXiv preprint arXiv:2311.05437}, 2023.

\bibitem[Mahjourian et~al.(2018)Mahjourian, Wicke, and Angelova]{mahjourian2018unsupervised}
Reza Mahjourian, Martin Wicke, and Anelia Angelova.
\newblock Unsupervised learning of depth and ego-motion from monocular video using 3d geometric constraints.
\newblock In \emph{CVPR}, 2018.

\bibitem[Mardia et~al.(2000)Mardia, Jupp, and Mardia]{mardia2000directional}
Kanti~V Mardia, Peter~E Jupp, and KV Mardia.
\newblock \emph{Directional statistics}.
\newblock 2000.

\bibitem[Messikommer et~al.(2024)Messikommer, Cioffi, Gehrig, and Scaramuzza]{messikommer2024reinforcementlearningmeetsvisual}
Nico Messikommer, Giovanni Cioffi, Mathias Gehrig, and Davide Scaramuzza.
\newblock Reinforcement learning meets visual odometry.
\newblock \emph{ECCV}, 2024.

\bibitem[Michaelis et~al.(2019)Michaelis, Mitzkus, Geirhos, Rusak, Bringmann, Ecker, Bethge, and Brendel]{michaelis2019dragon}
Claudio Michaelis, Benjamin Mitzkus, Robert Geirhos, Evgenia Rusak, Oliver Bringmann, Alexander~S. Ecker, Matthias Bethge, and Wieland Brendel.
\newblock Benchmarking robustness in object detection: Autonomous driving when winter is coming.
\newblock \emph{arXiv preprint arXiv:1907.07484}, 2019.

\bibitem[Mohlin et~al.(2020)Mohlin, Sullivan, and Bianchi]{mohlin2020probabilistic}
David Mohlin, Josephine Sullivan, and G{\'e}rald Bianchi.
\newblock Probabilistic orientation estimation with matrix fisher distributions.
\newblock In \emph{NeurIPS}, 2020.

\bibitem[Mur-Artal and Tard{\'o}s(2017)]{mur2017orb}
Raul Mur-Artal and Juan~D Tard{\'o}s.
\newblock Orb-slam2: An open-source slam system for monocular, stereo, and rgb-d cameras.
\newblock \emph{T-RO}, 2017.

\bibitem[Mur-Artal et~al.(2015)Mur-Artal, Montiel, and Tardos]{mur2015orb}
Raul Mur-Artal, Jose Maria~Martinez Montiel, and Juan~D Tardos.
\newblock Orb-slam: a versatile and accurate monocular slam system.
\newblock \emph{T-RO}, 2015.

\bibitem[Piccinelli et~al.(2024)Piccinelli, Yang, Sakaridis, Segu, Li, Van~Gool, and Yu]{piccinelli2024unidepth}
Luigi Piccinelli, Yung-Hsu Yang, Christos Sakaridis, Mattia Segu, Siyuan Li, Luc Van~Gool, and Fisher Yu.
\newblock Unidepth: Universal monocular metric depth estimation.
\newblock In \emph{CVPR}, 2024.

\bibitem[Pretto et~al.(2009)Pretto, Menegatti, Bennewitz, Burgard, and Pagello]{pretto2009visual}
Alberto Pretto, Emanuele Menegatti, Maren Bennewitz, Wolfram Burgard, and Enrico Pagello.
\newblock A visual odometry framework robust to motion blur.
\newblock In \emph{ICRA}, 2009.

\bibitem[Reimers and Gurevych()]{reimers-2019-sentence-bert}
Nils Reimers and Iryna Gurevych.
\newblock Sentence-bert: Sentence embeddings using siamese bert-networks.
\newblock In \emph{EMNLP}.

\bibitem[Rizve et~al.(2021)Rizve, Duarte, Rawat, and Shah]{rizve2021defense}
Mamshad~Nayeem Rizve, Kevin Duarte, Yogesh~S Rawat, and Mubarak Shah.
\newblock In defense of pseudo-labeling: An uncertainty-aware pseudo-label selection framework for semi-supervised learning.
\newblock In \emph{ICLR}, 2021.

\bibitem[Rockwell et~al.(2022)Rockwell, Johnson, and Fouhey]{rockwell20228}
Chris Rockwell, Justin Johnson, and David~F Fouhey.
\newblock The 8-point algorithm as an inductive bias for relative pose prediction by vits.
\newblock In \emph{3DV}. IEEE, 2022.

\bibitem[Schubert et~al.(2018)Schubert, Goll, Demmel, Usenko, St{\"u}ckler, and Cremers]{schubert2018tum}
David Schubert, Thore Goll, Nikolaus Demmel, Vladyslav Usenko, J{\"o}rg St{\"u}ckler, and Daniel Cremers.
\newblock The tum vi benchmark for evaluating visual-inertial odometry.
\newblock In \emph{IROS}, 2018.

\bibitem[Shi et~al.(2023)Shi, Tonolini, Aletras, Yilmaz, Kazai, and Jiao]{shi2023rethinkingsemisupervisedlearninglanguage}
Zhengxiang Shi, Francesco Tonolini, Nikolaos Aletras, Emine Yilmaz, Gabriella Kazai, and Yunlong Jiao.
\newblock Rethinking semi-supervised learning with language models.
\newblock \emph{arXiv preprint arXiv:2305.13002}, 2023.

\bibitem[Souly et~al.(2017)Souly, Spampinato, and Shah]{souly2017semi}
Nasim Souly, Concetto Spampinato, and Mubarak Shah.
\newblock Semi supervised semantic segmentation using generative adversarial network.
\newblock In \emph{ICCV}, 2017.

\bibitem[Tang and Tan(2018)]{tang2018ba}
Chengzhou Tang and Ping Tan.
\newblock {BA}-net: Dense bundle adjustment network.
\newblock \emph{arXiv preprint arXiv:1806.04807}, 2018.

\bibitem[Tang et~al.(2021)Tang, Chen, Luo, and Zhang]{tang2021humble}
Yihe Tang, Weifeng Chen, Yijun Luo, and Yuting Zhang.
\newblock Humble teachers teach better students for semi-supervised object detection.
\newblock In \emph{CVPR}, 2021.

\bibitem[Teed and Deng(2021)]{teed2021droid}
Zachary Teed and Jia Deng.
\newblock {DROID-SLAM}: Deep visual slam for monocular, stereo, and rgb-d cameras.
\newblock \emph{NeurIPS}, 2021.

\bibitem[Teed et~al.(2023)Teed, Lipson, and Deng]{teed2024deep}
Zachary Teed, Lahav Lipson, and Jia Deng.
\newblock Deep patch visual odometry.
\newblock \emph{NeurIPS}, 2023.

\bibitem[Tian et~al.(2024)Tian, Li, Weng, Chen, Schmerling, Wang, Ivanovic, and Pavone]{tian2024tokenize}
Ran Tian, Boyi Li, Xinshuo Weng, Yuxiao Chen, Edward Schmerling, Yue Wang, Boris Ivanovic, and Marco Pavone.
\newblock Tokenize the world into object-level knowledge to address long-tail events in autonomous driving.
\newblock \emph{arXiv preprint arXiv:2407.00959}, 2024.

\bibitem[Vaswani et~al.(2017)Vaswani, Shazeer, Parmar, Uszkoreit, Jones, Gomez, Kaiser, and Polosukhin]{vaswani2017attention}
Ashish Vaswani, Noam Shazeer, Niki Parmar, Jakob Uszkoreit, Llion Jones, Aidan~N Gomez, {\L}ukasz Kaiser, and Illia Polosukhin.
\newblock Attention is all you need.
\newblock In \emph{NeurIPS}, 2017.

\bibitem[Vijayanarasimhan et~al.(2017)Vijayanarasimhan, Ricco, Schmid, Sukthankar, and Fragkiadaki]{vijayanarasimhan2017sfmnetlearningstructuremotion}
Sudheendra Vijayanarasimhan, Susanna Ricco, Cordelia Schmid, Rahul Sukthankar, and Katerina Fragkiadaki.
\newblock Learning of structure and motion from video.
\newblock In \emph{CVPR}, 2017.

\bibitem[Wang et~al.(2021{\natexlab{a}})Wang, Cong, Litany, Gao, and Guibas]{wang20213dioumatch}
He Wang, Yezhen Cong, Or Litany, Yue Gao, and Leonidas~J Guibas.
\newblock {3DIoUMatch}: Leveraging {IoU} prediction for semi-supervised {3D} object detection.
\newblock In \emph{CVPR}, 2021{\natexlab{a}}.

\bibitem[Wang et~al.(2017)Wang, Clark, Wen, and Trigoni]{wang2017deepvo}
Sen Wang, Ronald Clark, Hongkai Wen, and Niki Trigoni.
\newblock Deepvo: Towards end-to-end visual odometry with deep recurrent convolutional neural networks.
\newblock In \emph{ICRA}, 2017.

\bibitem[Wang et~al.(2020)Wang, Zhu, Wang, Hu, Qiu, Wang, Hu, Kapoor, and Scherer]{wang2020tartanair}
Wenshan Wang, Delong Zhu, Xiangwei Wang, Yaoyu Hu, Yuheng Qiu, Chen Wang, Yafei Hu, Ashish Kapoor, and Sebastian Scherer.
\newblock Tartanair: A dataset to push the limits of visual slam.
\newblock In \emph{IROS}, 2020.

\bibitem[Wang et~al.(2021{\natexlab{b}})Wang, Hu, and Scherer]{wang2020tartanvogeneralizablelearningbasedvo}
Wenshan Wang, Yaoyu Hu, and Sebastian Scherer.
\newblock Tartanvo: A generalizable learning-based vo.
\newblock In \emph{CoRL}, 2021{\natexlab{b}}.

\bibitem[Wang et~al.(2019)Wang, Chao, Garg, Hariharan, Campbell, and Weinberger]{wang2019pseudo}
Yan Wang, Wei-Lun Chao, Divyansh Garg, Bharath Hariharan, Mark Campbell, and Kilian~Q Weinberger.
\newblock Pseudo-lidar from visual depth estimation: Bridging the gap in 3d object detection for autonomous driving.
\newblock In \emph{CVPR}, 2019.

\bibitem[Wang et~al.(2024)Wang, Lipson, and Deng]{wang2025sea}
Yihan Wang, Lahav Lipson, and Jia Deng.
\newblock {SEA-RAFT}: Simple, efficient, accurate raft for optical flow.
\newblock In \emph{ECCV}, 2024.

\bibitem[Wilson et~al.(2023)Wilson, Qi, Agarwal, Lambert, Singh, Khandelwal, Pan, Kumar, Hartnett, Pontes, Ramanan, Carr, and Hays]{wilson2023argoverse2generationdatasets}
Benjamin Wilson, William Qi, Tanmay Agarwal, John Lambert, Jagjeet Singh, Siddhesh Khandelwal, Bowen Pan, Ratnesh Kumar, Andrew Hartnett, Jhony~Kaesemodel Pontes, Deva Ramanan, Peter Carr, and James Hays.
\newblock Argoverse 2: Next generation datasets for self-driving perception and forecasting, 2023.

\bibitem[Xue et~al.(2019)Xue, Wang, Li, Wang, Wang, and Zha]{xue2019beyond}
Fei Xue, Xin Wang, Shunkai Li, Qiuyuan Wang, Junqiu Wang, and Hongbin Zha.
\newblock Beyond tracking: Selecting memory and refining poses for deep visual odometry.
\newblock In \emph{CVPR}, 2019.

\bibitem[Yalniz et~al.(2019)Yalniz, J{\'e}gou, Chen, Paluri, and Mahajan]{yalniz2019billion}
I~Zeki Yalniz, Herv{\'e} J{\'e}gou, Kan Chen, Manohar Paluri, and Dhruv Mahajan.
\newblock Billion-scale semi-supervised learning for image classification.
\newblock \emph{arXiv preprint arXiv:1905.00546}, 2019.

\bibitem[Yang et~al.(2022)Yang, Li, Zhang, Xiao, Liu, Yuan, and Gao]{yang2022unified}
Jianwei Yang, Chunyuan Li, Pengchuan Zhang, Bin Xiao, Ce Liu, Lu Yuan, and Jianfeng Gao.
\newblock Unified contrastive learning in image-text-label space.
\newblock In \emph{CVPR}, 2022.

\bibitem[Yang et~al.(2024{\natexlab{a}})Yang, Gao, Qiu, Chen, Li, Dai, Chitta, Wu, Zeng, Luo, et~al.]{yang2024generalized}
Jiazhi Yang, Shenyuan Gao, Yihang Qiu, Li Chen, Tianyu Li, Bo Dai, Kashyap Chitta, Penghao Wu, Jia Zeng, Ping Luo, et~al.
\newblock Generalized predictive model for autonomous driving.
\newblock In \emph{CVPR}, 2024{\natexlab{a}}.

\bibitem[Yang et~al.(2024{\natexlab{b}})Yang, Kang, Huang, Xu, Feng, and Zhao]{depthanything}
Lihe Yang, Bingyi Kang, Zilong Huang, Xiaogang Xu, Jiashi Feng, and Hengshuang Zhao.
\newblock Depth anything: Unleashing the power of large-scale unlabeled data.
\newblock In \emph{CVPR}, 2024{\natexlab{b}}.

\bibitem[Yang et~al.(2020)Yang, Stumberg, Wang, and Cremers]{yang2020d3vo}
Nan Yang, Lukas~von Stumberg, Rui Wang, and Daniel Cremers.
\newblock D3vo: Deep depth, deep pose and deep uncertainty for monocular visual odometry.
\newblock In \emph{CVPR}, 2020.

\bibitem[Yang et~al.(2021)Yang, Song, King, and Xu]{yang2021survey}
Xiangli Yang, Zixing Song, Irwin King, and Zenglin Xu.
\newblock A survey on deep semi-supervised learning.
\newblock \emph{arXiv preprint arXiv:2103.00550}, 2021.

\bibitem[Ye et~al.(2023)Ye, Lan, Chen, Ming, Yu, Bao, Cui, and Zhang]{ye2023pvopanopticvisualodometry}
Weicai Ye, Xinyue Lan, Shuo Chen, Yuhang Ming, Xingyuan Yu, Hujun Bao, Zhaopeng Cui, and Guofeng Zhang.
\newblock Pvo: Panoptic visual odometry.
\newblock In \emph{CVPR}, 2023.

\bibitem[Yin et~al.(2023)Yin, Zhang, Chen, Cai, Yu, Wang, Chen, and Shen]{yin2023metric3d}
Wei Yin, Chi Zhang, Hao Chen, Zhipeng Cai, Gang Yu, Kaixuan Wang, Xiaozhi Chen, and Chunhua Shen.
\newblock Metric3d: Towards zero-shot metric 3d prediction from a single image.
\newblock In \emph{CVPR}, 2023.

\bibitem[Yin and Shi(2018)]{yin2018geonet}
Zhichao Yin and Jianping Shi.
\newblock Geonet: Unsupervised learning of dense depth, optical flow and camera pose.
\newblock In \emph{CVPR}, 2018.

\bibitem[You et~al.(2016)You, Jin, Wang, Fang, and Luo]{you2016image}
Quanzeng You, Hailin Jin, Zhaowen Wang, Chen Fang, and Jiebo Luo.
\newblock Image captioning with semantic attention.
\newblock In \emph{CVPR}, 2016.

\bibitem[Zeng et~al.(2024)Zeng, Wang, Yang, Park, Soatto, Lao, and Wong]{zeng2024wordepthvariationallanguageprior}
Ziyao Zeng, Daniel Wang, Fengyu Yang, Hyoungseob Park, Stefano Soatto, Dong Lao, and Alex Wong.
\newblock Wordepth: Variational language prior for monocular depth estimation.
\newblock In \emph{CVPR}, 2024.

\bibitem[Zhan et~al.(2018)Zhan, Garg, Weerasekera, Li, Agarwal, and Reid]{zhan2018unsupervised}
Huangying Zhan, Ravi Garg, Chamara~Saroj Weerasekera, Kejie Li, Harsh Agarwal, and Ian Reid.
\newblock Unsupervised learning of monocular depth estimation and visual odometry with deep feature reconstruction.
\newblock In \emph{CVPR}, 2018.

\bibitem[Zhan et~al.(2021)Zhan, Weerasekera, Bian, Garg, and Reid]{zhan2021df}
Huangying Zhan, Chamara~Saroj Weerasekera, Jia-Wang Bian, Ravi Garg, and Ian Reid.
\newblock Df-vo: What should be learnt for visual odometry?
\newblock \emph{arXiv preprint arXiv:2103.00933}, 2021.

\bibitem[Zhang et~al.(2007)Zhang, Kwan-yee, and Zhang]{zhang2007camera}
Hui Zhang, K~Wong Kwan-yee, and Guoqiang Zhang.
\newblock Camera calibration from images of spheres.
\newblock \emph{PAMI}, 2007.

\bibitem[Zhang et~al.(2022)Zhang, Zhu, and Ohn-Bar]{zhang2022selfd}
Jimuyang Zhang, Ruizhao Zhu, and Eshed Ohn-Bar.
\newblock Selfd: Self-learning large-scale driving policies from the web.
\newblock In \emph{CVPR}, 2022.

\bibitem[Zhang et~al.(2024)Zhang, Huang, Ray, and Ohn-Bar]{zhang2024feedback}
Jimuyang Zhang, Zanming Huang, Arijit Ray, and Eshed Ohn-Bar.
\newblock Feedback-guided autonomous driving.
\newblock In \emph{CVPR}, 2024.

\bibitem[Zhang et~al.()Zhang, Zhou, Liu, and Shang]{zhang2016flexible}
Yueqiang Zhang, Langming Zhou, Haibo Liu, and Yang Shang.
\newblock A flexible online camera calibration using line segments.
\newblock \emph{Journal of Sensors}, 2016.

\bibitem[Zhang(2000)]{888718}
Z. Zhang.
\newblock A flexible new technique for camera calibration.
\newblock \emph{PAMI}, 2000.

\bibitem[Zhang(2004)]{zhang2004camera}
Zhengyou Zhang.
\newblock Camera calibration with one-dimensional objects.
\newblock \emph{PAMI}, 2004.

\bibitem[Zhao et~al.(2020)Zhao, Sheng, Dong, Chang, Xu, et~al.]{zhao2020maskflownet}
Shengyu Zhao, Yilun Sheng, Yue Dong, Eric~I Chang, Yan Xu, et~al.
\newblock Maskflownet: Asymmetric feature matching with learnable occlusion mask.
\newblock In \emph{CVPR}, 2020.

\bibitem[Zhou et~al.(2017)Zhou, Brown, Snavely, and Lowe]{zhou2017unsupervised}
Tinghui Zhou, Matthew Brown, Noah Snavely, and David~G Lowe.
\newblock Unsupervised learning of depth and ego-motion from video.
\newblock In \emph{CVPR}, 2017.

\bibitem[Zhu et~al.(2023)Zhu, Huang, Ohn-Bar, and Saligrama]{zhu2023learning}
Ruizhao Zhu, Peng Huang, Eshed Ohn-Bar, and Venkatesh Saligrama.
\newblock Learning to drive anywhere.
\newblock In \emph{CoRL}, 2023.

\bibitem[Zhu et~al.(2024{\natexlab{a}})Zhu, Kumar, Hu, and Liu]{zhu2024tame}
Shengjie Zhu, Abhinav Kumar, Masa Hu, and Xiaoming Liu.
\newblock Tame a wild camera: in-the-wild monocular camera calibration.
\newblock \emph{NeurIPS}, 2024{\natexlab{a}}.

\bibitem[Zhu et~al.(2024{\natexlab{b}})Zhu, Peng, Larsson, Cui, Oswald, Geiger, and Pollefeys]{Zhu2023NICER}
Zihan Zhu, Songyou Peng, Viktor Larsson, Zhaopeng Cui, Martin~R Oswald, Andreas Geiger, and Marc Pollefeys.
\newblock Nicer-slam: Neural implicit scene encoding for rgb slam.
\newblock In \emph{3DV}, 2024{\natexlab{b}}.

\end{thebibliography}
}
\clearpage
\onecolumn
\begin{center}
  {\LARGE\bf Supplementary Material for ZeroVO: \\
  Visual Odometry with Minimal Assumptions \par}
  \vspace{1em}
  {\large Lei Lai$^*$ \quad Zekai Yin$^*$ \quad Eshed Ohn-Bar \par}
  Boston University \\
  {\tt\small \{leilai, zekaiyin, eohnbar\}@bu.edu}
\end{center}
\vspace{2em}
\setcounter{section}{0}

\begin{abstract}
We provide additional details regarding the methodology and implementation of ZeroVO. Due to the limited prior research on scalable metric-scale VO under challenging autonomous driving settings (\eg, with unknown camera intrinsics), we include multiple ablation studies and failure case analysis. Additional qualitative comparisons are available in the attached \textbf{supplementary video}.
\end{abstract}

\section{Implementation Details}
\label{sec:supp_implementation}



\subsection{Network Architecture}
\label{sec:supp_struct}


\boldparagraph{Image Encoders}
To ensure \textbf{broad generalization} of our findings, in our work we focus on the \textbf{pure two-frame VO task}. Thus, our findings can be potentially applicable to diverse autonomous driving scenarios, \eg, to scale over scenarios with abrupt camera motion or lens settings. Our network takes as input a $640 \times 384 \times 6$ array comprising two sequential frames. The images are processed using a MaskFlownet~\cite{zhao2020maskflownet} encoder to extract the optical flow $\hat{\bF}$, which has dimensions $640 \times 384 \times 2$, and a correlation feature $\hat{\bF}^c$ of dimension $102 \times 160 \times 96$. As discussed in the main paper, we have experimented with various encoders~\cite{wang2025sea} and found MaskFlownet to perform best under our experimental conditions. To extract language-based features $\bZ^l$ with dimensions $15 \times 768$, we leverage LLaVA-NeXT~\cite{liu2024llavanext}, a state-of-the-art multi-modal large language model (LLM), alongside Sentence Transformers~\cite{reimers-2019-sentence-bert}. 
We further extract a metric-scale depth map using Metric3Dv2~\cite{hu2024metric3d}, which estimates intrinsic parameters from the image following Zhu~\etal~\cite{zhu2024tame}. Although single-image depth prediction is a well-studied task, we observe \textbf{persistent failure} cases under our experimental settings (detailed in Sec.~\ref{subsec:suppanaly}).

\boldparagraph{3D Depth and Scene Flow}
ZeroVO leverages cross-attention throughout our network to facilitate integrated learning over intrinsic parameters estimation noise, context, and geometry. The cross-attention structure first fuses language-based context to refine predicted depth and flow arrays, and subsequently incorporate geometry-aware cues. We process depth and flow information through dedicated transformer blocks. Specifically, we first compute 3D point cloud based on the predicted depth~\cite{wang2019pseudo}. 
Second, we compute the 3D scene flow by employing the predicted depth for the current and previous frames and unprojecting the matching 2D pixels (given by the optical flow), \ie,   
\begin{align*}
 \hat{\bF}^{\text{3D}} = \hat{\bD}^{\text{3D}}(\bu+\bo) - \hat{\bD}^{\text{3D}}(\bu)
 \end{align*}
 where $\bu = (u,v)$ is a pixel in the current 2D frame with flow displacement vector $\bo = (m,n)$. While this involves a naive and potentially noisy 3D estimation process~\cite{guizilini2022learning}, our key insight is to leverage it to guide the network via a cross-attention module. 
 
 
 


\boldparagraph{Decoder MLP} We directly decode the concatenated 3D-informed features into rotation and metric-scale translation. Our decoder leverages an MLP with two output branches. Each branch has three layers with a hidden dimension of 256 and Tanh
activations. Next, we detail the optimization process of our generalized VO network.

%









\subsection{Training Protocol}
\label{sec:supptraining}


We propose a semi-supervised training framework that leverages the multi-modal and generalized model structure described in Sec.\ref{sec:supp_struct}. The model is trained using the SGD optimizer for $100$ epochs with a batch size of 16. The initial learning rate is set to $0.001$. To analyze realistic use cases with a constrained initial dataset, the initial (\ie, teacher) ZeroVO model is first trained on the nuScenes-OneNorth dataset~\cite{nuscenes2019}. It is subsequently refined using VO pseudo-labels~\cite{lee2013pseudo}, as detailed in this section. Importantly, we do not utilize depth or flow supervision during training. Furthermore, we highlight that most self-supervised VO approaches incorporate consistency objectives based on \textbf{known intrinsic parameters} (\eg, from the known calibration parameters on the KITTI dataset~\cite{li2018undeepvo,zhou2017unsupervised,kendall2017geometric,bian2019unsupervised,yin2018geonet,li2020self,kanai2024self,guizilini2022learning}).



\boldparagraph{Semi-Supervised Training} We explore the impact of semi-supervised learning on the model's generalization capabilities by training our model on an unconstrained YouTube dataset~\cite{yang2024generalized}. This dataset is pseudo-labeled with VO predictions, which are subsequently filtered to remove redundant and noisy labels and used as training targets. In the \textbf{geometry-guided pseudo-label selection process}, a sample is deemed noisy and filtered out if $\text{normSSIM} < 0.5$ (\ie, inconsistent with warping). In the \textbf{language-guided pseudo-label selection process}, a temporal window of $H=10$ is applied to ensure consistency while selecting diverse samples. Samples are filtered out if $\text{subspace-sim} < 5$. 



\boldparagraph{Data Augmentations} During training, we employ two data augmentation strategies: Random Crop and Resize (\textbf{RCR})~\cite{wang2020tartanvogeneralizablelearningbasedvo} and Image Horizontal Flip (\textbf{IHF}).
RCR is used to simulate a diverse range of camera intrinsic settings, including variations in focal lengths, principal points, and skew factors. This augmentation introduces a broader spectrum of imaging conditions, thereby enhancing the model’s robustness and generalization capabilities. Additionally, IHF is applied to expand the dataset further. Given the harsh evaluation conditions (particularly on GTA), we experimented with additional data augmentation techniques, such as simulating visual artifacts like snow and fog~\cite{michaelis2019dragon} (\textbf{COR}, visual corruptions). However, as shown in Sec.~\ref{sec:supp_anaylysis}, these simplistic stylizations did not yield improvements in generalization to higher-fidelity simulation renderings.

\subsection{Validation With a Diverse GTA V Dataset}
\label{sec:dataset}

\begin{figure}[t!]
    \centering
\includegraphics[width=0.8\textwidth]{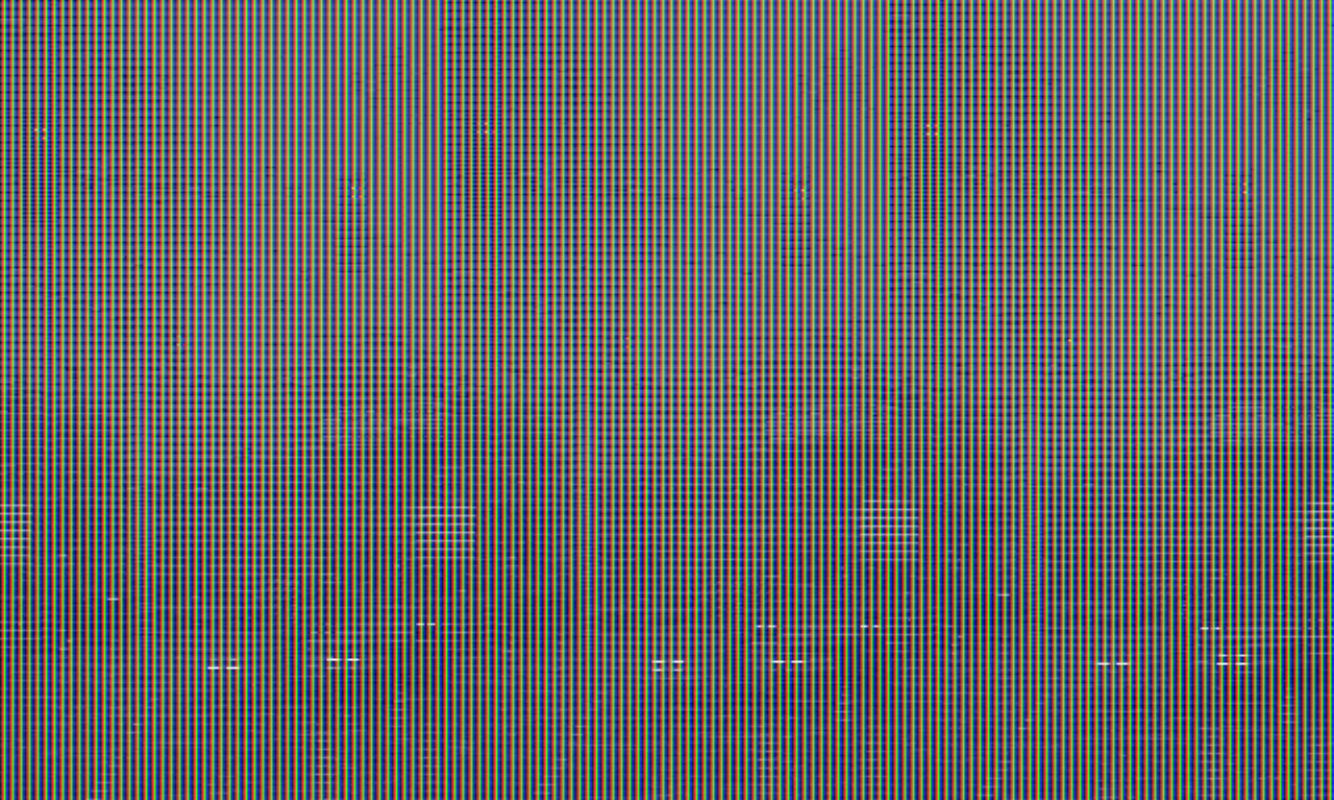}
\caption{\textbf{Example from Our GTA Dataset.} We leverage GTA to construct a diverse and challenging generalization dataset for validation.} 
    \label{fig:corruption}
\end{figure}

We evaluate ZeroVO in cross-dataset autonomous driving settings, which demand robust generalization across various camera setups. However, autonomous driving datasets such as nuScenes~\cite{nuscenes2019} and Argoverse~\cite{wilson2023argoverse2generationdatasets} \textbf{rarely feature diverse and adversarial conditions, including traffic, terrain, weather, time of day, and lens effects} (Table~\ref{tab:dataset_comparison}). Similarly, existing VO datasets are typically collected in controlled indoor environments without variations in weather or reflections~\cite{burri2016euroc,schubert2018tum}. TartanAir~\cite{wang2020tartanair}, a related simulation benchmark, includes diverse scenes and settings (such as fog and rain effects) from a drone’s perspective. However, TartanAir contains limited large dynamic objects and traffic scenarios, which can result in frequent trajectory drift. Additionally, it lacks lens corruption effects, such as rain droplets on the camera lens. When compared to the TarTanAir dataset, GTA provides significantly higher resolution, a greater focus on traffic-related environments, and a wider variety of dynamic objects, making it a robust resource for advancing VO research and development.
We collect a large and diverse dataset of on-road driving and off-road driving (each clip is 25 seconds long, captured at 10 FPS). 
Our dataset features high-resolution scenes with varied terrains, driving scenarios, visual artifacts, and lens corruption (\eg rain drops on the lens, see Fig.~\ref{fig:corruption}). Qualitative examples are provided in Sec.~\ref{sec:supp_anaylysis} and the supplementary video.

\begin{table}[!h]
\centering
\caption{\textbf{Dataset Comparison.} GTA provides diverse scenes that allow extensive validation of VO for autonomous driving. GTA includes off-road scenes, capturing challenging conditions such as uphill and downhill driving with abrupt movements, as well as lens corruption (with raindrops on the camera, see Fig.~\ref{fig:corruption}) and severe illumination effects. 
}
\vspace{-0.2cm}
\label{tab:dataset_comparison}
\resizebox{\columnwidth}{!}{
\begin{tabular}{lcccccccc}
\toprule
    Dataset & 
    Weathers & 
    Lens Corruption & 
    Resolution &
    Off-Road &
    Size & 
    Type \\ \midrule
KITTI~\cite{Geiger2012CVPR} & Day & \red{x} & 1226 $\times$ 370 & \red{x} & 23{,}201 Images  &  Real World \\ 

nuScenes~\cite{nuscenes2019} & Rain, Day, Night & \green{✓} (rare) & 1600 $\times$ 900 & \red{x} & 193{,}815 Images  &   Real World \\ 

Argoverse~\cite{wilson2023argoverse2generationdatasets} & Rain, Day, Night & \red{x} & 1550 $\times$ 2048 & \red{x} & 320{,}159 Images  &   Real World \\ 

 \rowcolor{lightgray}
\textbf{GTA (Ours)} &  Snow, Rain, Day, Night & \green{✓} & 1920 $\times$ 1080 & \green{✓} & 230{,}500 Images  &   Synthetic \\ 
\bottomrule
\end{tabular}
}
\end{table}

\subsection{Evaluation Metrics}
\label{sec: metrics}
We evaluate standard VO metrics, including translation, rotation, and Absolute Trajectory Error (ATE, the most commonly used metric~\cite{wang2020tartanvogeneralizablelearningbasedvo,wang2020tartanair,teed2021droid}). We emphasize that most prior VO methods evaluate translation predictions up-to-scale (in a process that leverages the ground-truth scale at each frame). Thus, to isolate the role of scale, we also directly compute a scale error metric (following~\cite{lai2023xvogeneralizedvisualodometry}). As the various error measures are complementary in principle, we report all for completeness (and formally define them below). Nonetheless, in practice, we find ATE to provide a good summarized metric for performance in our settings.





Given a scene with $N$ frames, after estimating all the relative poses $\{\bT_i\}_{i=2}^{N}$, where $\bT_i$ represents the relative pose between frames $\{\bI_{i-1}, \bI_i\}$, the global trajectory is reconstructed by calculating the pose $\bG_{i\rightarrow 1}$ of each frame $i$ with respect to the initial frame, \ie, expressed in the camera coordinate system of the first frame: $\bG_{i\rightarrow 1} = \prod_{j=2}^i \bT_j $ for $\forall i>1$, with $\bG_{1\rightarrow 1}$ defined as the $4\times4$ identity matrix. Once the trajectory is recovered, evaluation metrics can be computed as follows: 

\boldparagraph{Translation and Rotation Errors} We follow standard definitions for the translation and rotation errors. For each subsequence of length $l\in \{100,...,800\}$ meters, derived from the recovered trajectory based on ground-truth poses, with the initial and final frames $\bI_i$ and $\bI_{i+n}$, we calculate the relative pose between frames $\{\bI_i, \bI_{i+n}\}$ as follows:
\begin{equation}
\begin{gathered}
    \bG_{i+n\rightarrow i} = \bG_{i\rightarrow 1}^{-1} \bG_{i+n\rightarrow 1} \\
    \hat{\bG}_{i+n\rightarrow i} = \hat{\bG}_{i\rightarrow 1}^{-1} \hat{\bG}_{i+n\rightarrow 1} \\
    \bar{\bG} = \hat{\bG}_{i+n\rightarrow i}^{-1} \bG_{i+n\rightarrow i}
\end{gathered}
\end{equation}
where $\bG_{i+n\rightarrow i}$, $\hat{\bG}_{i+n\rightarrow i}$ are the ground-truth and estimated relative pose of frame $\bI_{i+n}$ with respect to $\bI_{i}$, and $\bar{\bG}=[\bar{\bR}|\bar{\bt}]$ is the pose difference between $\bG_{i+n\rightarrow i}$ and $\hat{\bG}_{i+n\rightarrow i}$. Translation and Rotation Error $t_{err}$ (\%) and $r_{err}$ ($^\circ$/100\, \text{m}) on each subsequence are then calculated as:
\begin{equation}
    t_{err} = \frac{\| \bar{\bt} \|_2}{l} \cdot 100,\ \  r_{err} = \frac{\arccos(0.5\cdot (\text{trace}(\bar{\bR})-1))}{l} \cdot 100
\end{equation}

\boldparagraph{Absolute Trajectory Error} The metrics $t_{err}$ and $r_{err}$ quantify the average translation drift distance and rotation drift angle per 100 meters, providing insight into the model's average performance in pose estimation. However, $t_{err}$ and $r_{err}$ do not adequately illustrate how the drift accumulates across the global trajectory. ATE on the hand provides a measure of the overall accuracy of the estimated trajectory in comparison to the ground-truth trajectory:
\begin{equation}
    \text{ATE} = \sqrt{\frac{1}{N} \sum_{i=1}^{N} \| \bt_{i\rightarrow 1} - \hat{\bt}_{i\rightarrow 1} \|^2}
\end{equation}
where $\bt_{i\rightarrow 1}$ and $\hat{\bt}_{i\rightarrow 1}$ represent the ground-truth and estimated translation in the poses $\bG_{i\rightarrow 1}$ and $\hat{\bG}_{i\rightarrow 1}$, respectively.

\boldparagraph{Scale Error} While VO methods are typically evaluated using the aforementioned metrics~\cite{wang2017deepvo, li2018undeepvo, lai2023xvogeneralizedvisualodometry, zhou2017unsupervised, yin2018geonet, xue2019beyond, wang2020tartanvogeneralizablelearningbasedvo}, many studies employ \textit{up-to-scale} settings where the ground-truth scale is used to rescale predictions at every time step. In contrast, we focus on estimating metric-scale translation, as both $t_{err}$ and ATE overlook instantaneous scale errors. To address this, we compute an isolated scale error. For example, consider a scenario where the ground-truth pose involves moving forward by $1.0$ meter in two consecutive steps, between frames $\{\bI_1, \bI_2\}$ and $\{\bI_2, \bI_3\}$. If the estimated pose instead involves moving forward by $0.5$ meters followed by $1.5$ meters, $t_{err}$ will not capture this instantaneous scale discrepancy. Furthermore, while \textbf{ATE does not account for scale error} in this case (since the global trajectory aligns with the ground truth by the second step, resulting in an ATE of zero), the scale estimation in both steps is inaccurate. To address this limitation, we report a scale error (following~\cite{lai2023xvogeneralizedvisualodometry}) that isolates real-world scale recovery performance at each time step:
\begin{equation} 
s_{err}^i = 1-\min\left(\frac{\|\hat{\bt}_i\|_2}{\max(\|\bt_i\|_2,\epsilon)}, \frac{\|\bt_i\|_2}{\max(\|\hat{\bt}_i\|_2,\epsilon)}\right),\ \  s_{err} = \frac{1}{N-1} \sum_{i=2}^N s_{err}^i
\end{equation}
where $\bt_{i}, \hat{\bt}_{i}$ represent the ground-truth and estimated translation in the relative poses $\bT_{i}$, $\hat{\bT}_{i}$, respectively, and $\epsilon$ is a small constant to prevent division by zero. While we report all four metrics for a holistic evaluation, we find that the overall trends remain consistent across them. (\ie, the scale error differs significantly only in contrived examples, making ATE sufficient for most evaluations.)

\subsection{State-of-the-Art Baselines}
\label{sec: baselines}
As few existing models predict metric-scale translation, we adapt the state-of-the-art DROID-SLAM~\cite{teed2021droid} for this purpose. Specifically, we replace the depth module in DROID-SLAM with Metric3Dv2~\cite{hu2024metric3d}, enabling the prediction of real-world scaled poses and facilitating a fair comparison with ZeroVO. However, as shown in Sec.~\ref{sec:supp_anaylysis}, ZeroVO significantly outperforms DROID-SLAM in translation prediction, which is a critical capability for real-world autonomous systems. We also compare our approach to XVO~\cite{lai2023xvogeneralizedvisualodometry} using its publicly available model which was trained on nuScenes and YouTube data.  




\section{Detailed Analysis}
\label{sec:supp_anaylysis}


\subsection{Additional Ablations}
\label{subsec:suppanaly}

\boldparagraph{Pseud-Label Selection} Table $\ref{tab:ablation_filters}$ shows an ablation over the proposed semi-supervised training framework. We find both the language and geometry consistency-based mechanisms to complement, resulting in best overall performance. This is intuitive, as the language-based selection facilitates selecting diverse unlabeled samples while the geometry-based objective excludes noisy samples.



\boldparagraph{Data Augmentations Strategies} Table~\ref{tab:ablation_aug} illustrates the impact of different data augmentation strategies. We find RCR to particularly impact the model's rotation estimation capability. RCR is crucial for learning to handle diverse camera intrinsic parameters (consistent with~\cite{wang2020tartanvogeneralizablelearningbasedvo}). Interestingly, we do not find augmentation via synthetic weathers and stylization to improve generalization, potentially due to the lack of realism. 


\boldparagraph{Impact of Noisy Intrinsic Parameters} We also investigate the robustness of the models against noisy camera intrinsics. To do this, we randomly add noise to the estimated camera intrinsics at various noise levels. Table \ref{tab:noisy_level} demonstrates that camera intrinsics play a crucial role for all models, as increased noise levels lead to significant drops in performance. We show ZeroVO to demonstrate greater robustness to noise in camera intrinsics, maintaining strong performance despite the highly noisy settings.

\boldparagraph{DROID-SLAM Ablation} Given that our work focuses on two-frame pose regression, we evaluate the performance of DROID-SLAM~\cite{teed2021droid} when provided with shorter snippets and not long sequences of frames. To conduct this experiment, we segment the long sequences into shorter snippets and task M+DS (integrated Metric3D~\cite{hu2024metric3d} and DROID-SLAM) with estimating poses for each snippet, subsequently concatenating them to recover the global trajectory. Table \ref{tab:ablation_frames} indicates that as the snippet length decreases, the performance of M+DS also declines, primarily due to the multi-frame optimization method's dependence on consistent frame tracking. When the snippet length is reduced to two frames, M+DS is unable to perform accurate estimation.

\setlength{\tabcolsep}{3pt}
\begin{table*}[!t]
\centering
\caption{\textbf{Pseudo-Label Selection.} Language and geometry-based selection are complementary for semi-supervised training.   }
\label{tab:ablation_filters}
{
\begin{tabular}{p{1.5cm}p{1.5cm}|cccc|cccc|cccc|cccc}
\toprule
\multirow{2}{*}{\textbf{Language}} & 
\multirow{2}{*}{\textbf{Geometry}} & 
\multicolumn{4}{c|}{\textbf{KITTI 00-10}} &
\multicolumn{4}{c|}{\textbf{nuScenes}} &
\multicolumn{4}{c|}{\textbf{Argoverse}} &
\multicolumn{4}{c}{\textbf{GTA}}
\\  
\cline{3-18}
& & 
$t_{err}$ & $r_{err}$ & $\text{ATE}$ & $s_{err}$ &
$t_{err}$ & $r_{err}$ & $\text{ATE}$ & $s_{err}$ &
$t_{err}$ & $r_{err}$ & $\text{ATE}$ & $s_{err}$ &
$t_{err}$ & $r_{err}$ & $\text{ATE}$ & $s_{err}$ \\ 
\midrule

\checkmark& &8.86&2.85&119.15&0.08 &13.02&5.45&7.98&0.15 &8.58&4.35&4.88&0.11 &16.02&11.93&13.27&0.19    \\
\checkmark&\checkmark &\textbf{6.81} &\textbf{2.69} &\textbf{104.69} &\textbf{0.06}   &\textbf{9.74} &\textbf{4.37} &\textbf{6.03} &\textbf{0.12}  &\textbf{4.64}&\textbf{2.83}&\textbf{3.05}&\textbf{0.09} &\textbf{13.42} &\textbf{7.99} &\textbf{8.24} &\textbf{0.17}    \\

\bottomrule
\end{tabular}}
\end{table*} 

\setlength{\tabcolsep}{3pt}
\begin{table*}[!t]
\centering
\caption{\textbf{Impact of Data Augmentation.} We find both \textbf{RCR} (Random Crop and Resize) and \textbf{IHF} (Image Horizontal Flip) to benefit model generalization. However, leveraging image corruption and stylized augmentations~\cite{michaelis2019dragon} (denoted as \textbf{COR}) provides marginal benefits. 
}
\label{tab:ablation_aug}
{
\begin{tabular}{p{1cm}p{1cm}p{0.8cm}|cccc|cccc|cccc|cccc}
\toprule
\multirow{2}{*}{\textbf{RCR}} & 
\multirow{2}{*}{\textbf{IHF}} & 
\multirow{2}{*}{\textbf{COR}} & 
\multicolumn{4}{c|}{\textbf{KITTI 00-10}} &
\multicolumn{4}{c|}{\textbf{nuScenes}} &
\multicolumn{4}{c|}{\textbf{Argoverse}} &
\multicolumn{4}{c}{\textbf{GTA}}
\\  
\cline{4-19}
& & &
$t_{err}$ & $r_{err}$ & $\text{ATE}$ & $s_{err}$ &
$t_{err}$ & $r_{err}$ & $\text{ATE}$ & $s_{err}$ &
$t_{err}$ & $r_{err}$ & $\text{ATE}$ & $s_{err}$ &
$t_{err}$ & $r_{err}$ & $\text{ATE}$ & $s_{err}$ \\ 
\midrule
\checkmark& & &7.99&2.74&109.55&0.06 &12.92&5.17&8.13&0.15 &6.97&4.34&4.02&0.10 &14.73&10.35&11.73&0.19    \\
\checkmark&\checkmark & &\textbf{6.81} &\textbf{2.69} &\textbf{104.69} &\textbf{0.06}   &\textbf{9.74} &\textbf{4.37} &\textbf{6.03} &\textbf{0.12}  &\textbf{4.64}&\textbf{2.83}&\textbf{3.05}&\textbf{0.09} &\textbf{13.42} &\textbf{7.99} &\textbf{8.24} &\textbf{0.17}    \\
\checkmark&\checkmark &\checkmark &7.81&2.67&113.62&0.07 &12.69&4.82&7.88&0.14 &7.05&4.34&4.16&0.10 &15.28&9.36&12.36&0.18    \\
\bottomrule
\end{tabular}}
\end{table*} 





\setlength{\tabcolsep}{2.69pt}
\begin{table*}[!t]
\centering
\caption{\textbf{Impact of Noisy Intrinsics Parameters.} We inject Gaussian noise into the estimated camera intrinsic parameters to analyze the impact on DROID-SLAM and ZeroVO. We note that scale error cannot be computed for up-to-scale methods.   }
\label{tab:noisy_level}
{
\begin{tabular}{l|cccc|cccc|cccc|cccc}
\toprule
\multirow{2}{*}{\diagbox{\textbf{Method}}{\textbf{Noise}}} & 
\multicolumn{4}{c|}{0} &
\multicolumn{4}{c|}{$10\%$} &
\multicolumn{4}{c|}{$20\%$} &
\multicolumn{4}{c}{$30\%$}
\\  
\cline{2-17}
& 
$t_{err}$ & $r_{err}$ & $\text{ATE}$ & $s_{err}$ &
$t_{err}$ & $r_{err}$ & $\text{ATE}$ & $s_{err}$ &
$t_{err}$ & $r_{err}$ & $\text{ATE}$ & $s_{err}$ &
$t_{err}$ & $r_{err}$ & $\text{ATE}$ & $s_{err}$ \\ 
\bottomrule
    \multicolumn{9}{l}{\textit{Baselines Requiring Ground-Truth Scale Alignment:}} \\
    \toprule    
TartanVO~\cite{wang2020tartanvogeneralizablelearningbasedvo} &13.85 &3.27 &103.07 & - &13.87&3.39&109.78&- &16.90&3.83&130.09&- &18.55&4.11&133.24&-    \\
DPVO~\cite{teed2024deep} &8.31 &2.37 &78.53 & - &15.32&2.79&152.56&- &23.54&3.15&196.46&- &36.92&4.90&280.49&-    \\
\bottomrule
    \multicolumn{9}{l}{\textit{Metric-Scale Zero-Shot Setting:}} \\
    \toprule   
M+DS~\cite{teed2021droid} &14.22 &2.72 &154.77 &0.09 &15.65&2.90&161.58&0.09 &17.55&3.74&179.92&0.10 &23.33&4.45&259.74&0.11    \\
ZeroVO+ &\textbf{6.81} &\textbf{2.69} &\textbf{104.69} &\textbf{0.06} &\textbf{7.44}&\textbf{2.71}&\textbf{117.10}&\textbf{0.06} &\textbf{11.33}&\textbf{3.27}&\textbf{153.95}&\textbf{0.08} &\textbf{13.75}&\textbf{3.85}&\textbf{163.87}&\textbf{0.10}    \\
\bottomrule
\end{tabular}}
\end{table*} 

\setlength{\tabcolsep}{3.7pt}
\begin{table*}[!t]
\centering
\caption{\textbf{DROID-SLAM Ablation.} DROID-SLAM~\cite{teed2021droid} leverages multi-frame input and optimization (in contrast, ZeroVO only relies on a two frame input). We find significant degradation in performance when reducing the number of input frames into DROID-SLAM (with Metric3DV2~\cite{hu2024metric3d}). 
}
\label{tab:ablation_frames}
{
\begin{tabular}{c|cccc|cccc|cccc|cccc}
\toprule
\multirow{3}{*}{\makecell{\textbf{Sequence} \\ \textbf{Length}}} & 

\multicolumn{4}{c|}{Entire Sequence} &
\multicolumn{4}{c|}{50 Frames} &
\multicolumn{4}{c|}{10 Frames} &
\multicolumn{4}{c}{2 Frames}
\\  
\cline{2-17}
& 
$t_{err}$ & $r_{err}$ & $\text{ATE}$ & $s_{err}$ &
$t_{err}$ & $r_{err}$ & $\text{ATE}$ & $s_{err}$ &
$t_{err}$ & $r_{err}$ & $\text{ATE}$ & $s_{err}$ &
$t_{err}$ & $r_{err}$ & $\text{ATE}$ & $s_{err}$ \\ 
\cline{2-17}
&14.22&2.72&154.77&0.09 &14.40&2.85&158.23&0.09 &16.31&3.51&180.67&0.10 &75.24&40.78&409.95&0.94    \\
\bottomrule
\end{tabular}}
\end{table*}

\subsection{Qualitative Examples}
\label{subsec:vis}
We provide several qualitative examples on the introduced GTA benchmark (Table~\ref{tab:gta_off} and Table~\ref{tab:gta_on}). Moreover, we analyze the impact of inaccurate depth estimation in Table \ref{tab:nusc_depth}. Finally, we show failure cases across datasets below (Table \ref{tab:gta_fail}, Table \ref{tab:argo_fail1}, Table \ref{tab:argo_fail2}, and Table \ref{tab:argo_fail3}). Based on our ablations, we observe several existing limitations. While single-image depth estimation modules has been extensively studied in literature, we still observe persistent failures within our harsh generalization settings. Scenarios related to camera lens and reflection are particularly difficult. Errors then propagate to the VO decoder both in baselines and ZeroVO. This highlights a future direction. 

\begin{table}[h]
\centering
\caption{\textbf{Off-Road Examples from GTA Dataset With Captions.} Off-road GTA visualizations across desert, forest, and mountain terrains under varying weather conditions. Captions include terrain descriptions, weather conditions, visibility, and other relevant factors.}
\label{tab:gta_off}
\begin{tabular}{|c|c|c|}
\hline
\rowcolor{lightgray}
\textbf{Desert} & \textbf{Forest} & \textbf{Mountain} \\
\hline
\begin{subfigure}[t]{0.3\textwidth}
   \includegraphics[width=\textwidth]{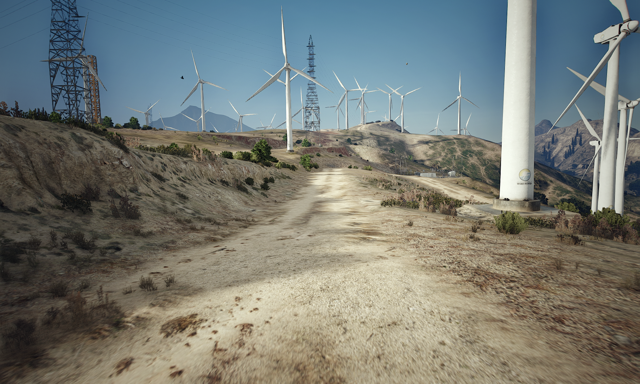}
   \caption{The image depicts a driving scenario where a vehicle is traveling
   on a \textcolor{dpp}{dirt road} that cuts through a landscape dominated by wind turbines. The road appears to be unpaved and dusty, suggesting it might be a \textcolor{dpp}{rural or remote area}. The wind turbines are tall and white, with multiple blades that are likely designed to capture wind energy. The sky is clear and blue, indicating good weather conditions. The \textcolor{dpp}{terrain is hilly} with sparse vegetation, and the overall setting suggests a location that is likely used for renewable energy production. }
\end{subfigure} &
\begin{subfigure}[t]{0.3\textwidth}
   \includegraphics[width=\textwidth]{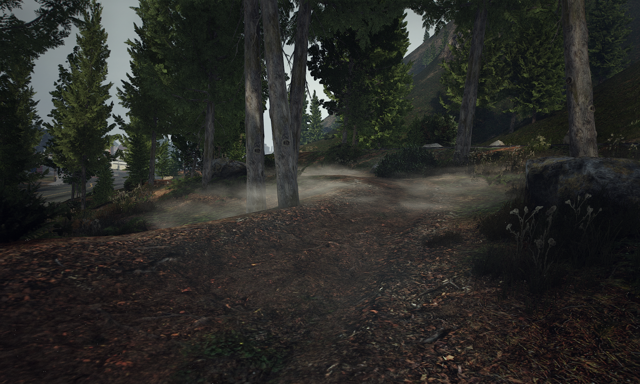}
   \caption{
The image depicts a driving scenario in a \textcolor{dpp}{forested area}. The road appears to be unpaved and is surrounded by tall trees, suggesting a \textcolor{dpp}{rural or remote location}. There is a layer of \textcolor{dpp}{fog or mist} that is rising from the ground, which could indicate recent rain or damp conditions. The lighting suggests it might be either early morning or late afternoon, as the sky is not brightly lit, and the shadows are long. The overall atmosphere is serene and somewhat mysterious due to the fog and the quiet, natural setting.}
\end{subfigure} &
\begin{subfigure}[t]{0.3\textwidth}
   \includegraphics[width=\textwidth]{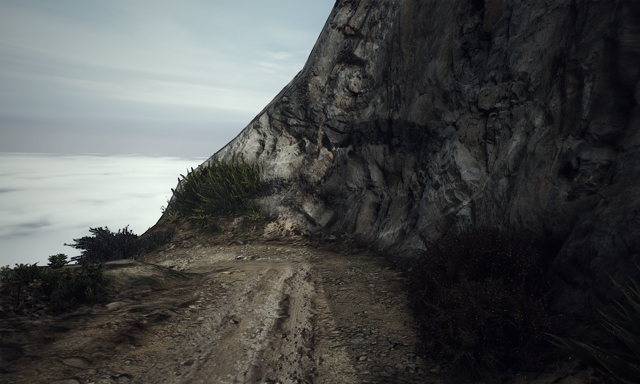}
   \caption{The image depicts a driving scenario on a narrow, unpaved road that appears to be winding along a cliff edge. \textcolor{dpp}{The road is not well-maintained}, with visible tire tracks and loose gravel, suggesting that it is frequently used by vehicles. The cliff on the left side of the road is \textcolor{dpp}{steep and rocky}, with sparse vegetation growing along the edge. The sky is overcast, and the lighting suggests it could be either early morning or late afternoon. The overall atmosphere is one of a \textcolor{dpp}{rugged, possibly remote} location, and the road seems to be a challenge for drivers.}
\end{subfigure} \\
\hline

\hline
\rowcolor{lightgray}
\textbf{Night Desert} & \textbf{Snowy Forest} & \textbf{Snowy Mountain} \\
\hline
\begin{subfigure}[t]{0.3\textwidth}
   \includegraphics[width=\textwidth]{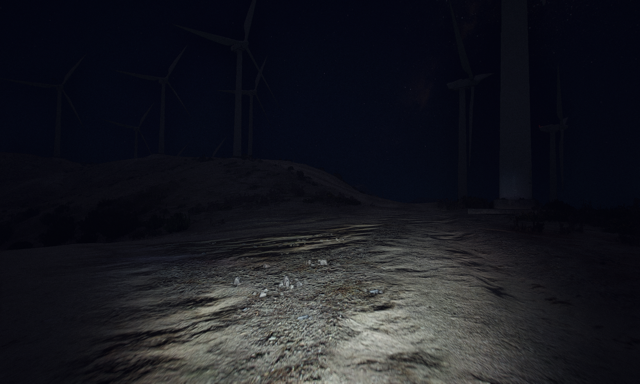}
   \caption{
The image depicts a nighttime driving scenario with a focus on the road ahead. The road appears to be a winding, possibly unpaved path \textcolor{dpp}{with a rough texture}, suggesting it might be a rural or remote area. The lighting is dim, with the headlights of the vehicle illuminating the path directly in front, creating a tunnel-like effect. The surrounding environment is dark, with the sky appearing to be deep blue, indicating it might be late evening or night. There are silhouettes of wind turbines in the background, indicating this might be a rural area. The overall atmosphere is quiet and serene, with \textcolor{dpp}{no visible traffic} or other vehicles on the road.}
\end{subfigure} &
\begin{subfigure}[t]{0.3\textwidth}
   \includegraphics[width=\textwidth]{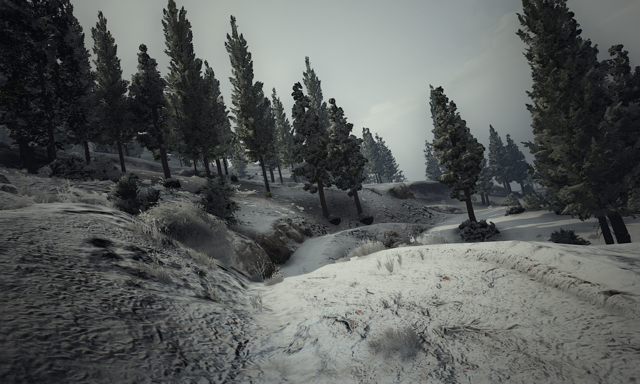}
   \caption{
The image depicts a snowy landscape with a \textcolor{dpp}{winding} road that appears to be a mountain pass or a \textcolor{dpp}{hilly road}. The road is covered in snow, and the surrounding environment is heavily forested with coniferous trees, suggesting a cold, possibly winter season. The sky is overcast, with a uniform gray color, which contributes to the chilly atmosphere of the scene. The lighting is soft and diffused, indicating either early morning or late afternoon, as the sun is not directly overhead. The overall driving scenario would likely involve navigating the winding road with caution due to the \textcolor{dpp}{slippery} conditions and the potential for \textcolor{dpp}{reduced visibility.}}
\end{subfigure} &
\begin{subfigure}[t]{0.3\textwidth}
   \includegraphics[width=\textwidth]{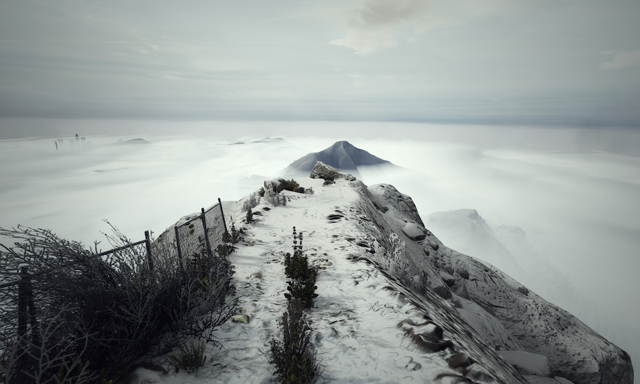}
   \caption{
The image depicts a dramatic and challenging driving scenario. The viewer's perspective is from the driver's seat, looking out over a \textcolor{dpp}{steep, snow-covered mountain pass.} The road appears to be narrow and winding, with a sharp drop-off on one side, suggesting a high level of risk and requiring a high degree of skill and caution from the driver. The presence of snow on the road and the surrounding landscape indicates that the conditions are likely to be \textcolor{dpp}{slippery and potentially hazardous.} The driver would need to navigate carefully, \textcolor{dpp}{maintaining a safe speed} and distance from the edge of the road, while also being vigilant for any sudden changes.}
\end{subfigure} \\
\hline
\end{tabular}
\end{table}
\begin{table}[h]
\centering
\caption{\textbf{On-Road Examples from GTA Dataset With Captions.} On-road GTA visualization across downtown, tunnel, and highway settings under varying weather conditions. Captions include traffic descriptions, visibility, and other relevant information about on-road driving conditions.}
\label{tab:gta_on}
\begin{tabular}{|c|c|c|}
\hline
\rowcolor{lightgray}

\textbf{Sunny City} & \textbf{Raining City at Night} & \textbf{Snowy City} \\
\hline
\begin{subfigure}[t]{0.3\textwidth}
   \includegraphics[width=\textwidth]{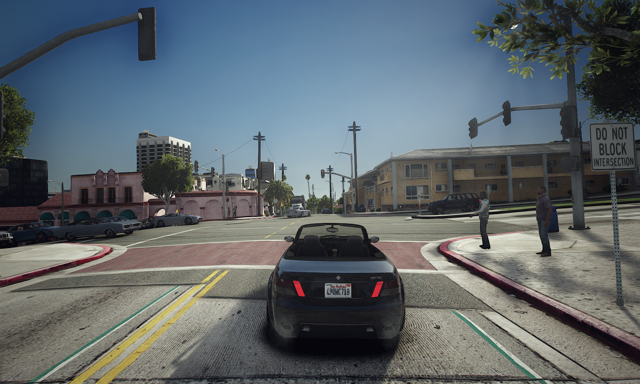}
   \caption{In the image, we see a driving scenario where a black convertible car is \textcolor{dpp}{stopped} at a traffic light. The car is positioned on the right side of the road, as indicated by the traffic light and the road markings. The traffic light is displaying a \textcolor{dpp}{red signal}, indicating that the car must stop. The car's \textcolor{dpp}{brake lights are on}, and the driver appears to be waiting for the light to change before proceeding. There are buildings on both sides of the street, and the architecture suggests a mix of commercial and residential structures. The road is marked with a crosswalk, and there are traffic signs visible, including one that reads "DO NOT BLOCK INTERSECTION," which is likely there to prevent vehicles from stopping in a way that could obstruct traffic flow. The overall scene is typical of an urban environment where traffic rules are enforced, and drivers are expected to follow the signals and signs to ensure the smooth flow of traffic and pedestrian safety.}
\end{subfigure} &
\begin{subfigure}[t]{0.3\textwidth}
   \includegraphics[width=\textwidth]{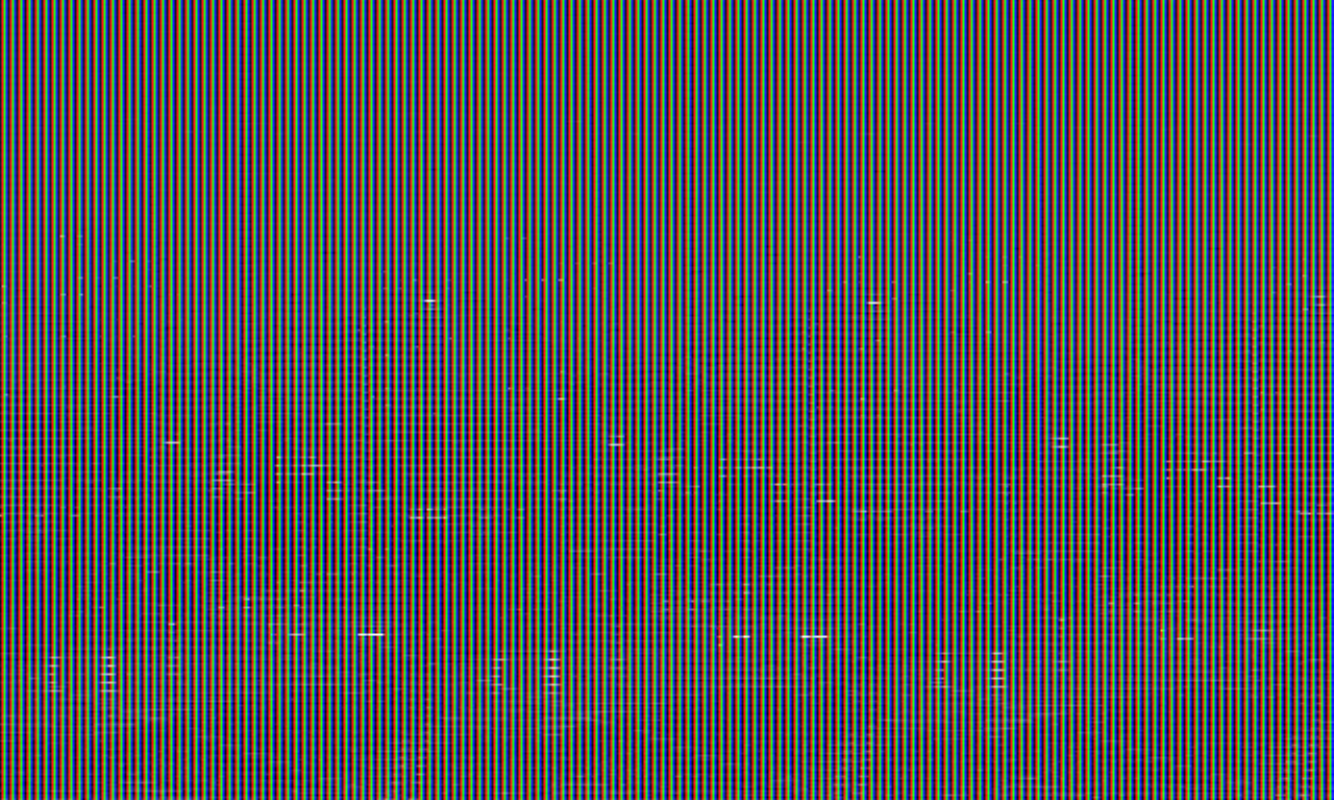}
   \caption{
The image depicts a nighttime driving scenario with a car on a wet road. The car's headlights are on, illuminating the immediate surroundings. The road appears to be slick from recent rain, as evidenced by the reflections of the headlights and the wet sheen on the surface. The weather is rainy, as indicated by the \textcolor{dpp}{raindrops} visible on the windshield and the wet conditions on the road. The street is lined with buildings, and there are other vehicles \textcolor{dpp}{parked or moving} in the distance. The overall atmosphere is moody and atmospheric, with the darkness of the night and the glow of the headlights creating a contrast.}
\end{subfigure} &
\begin{subfigure}[t]{0.3\textwidth}
   \includegraphics[width=\textwidth]{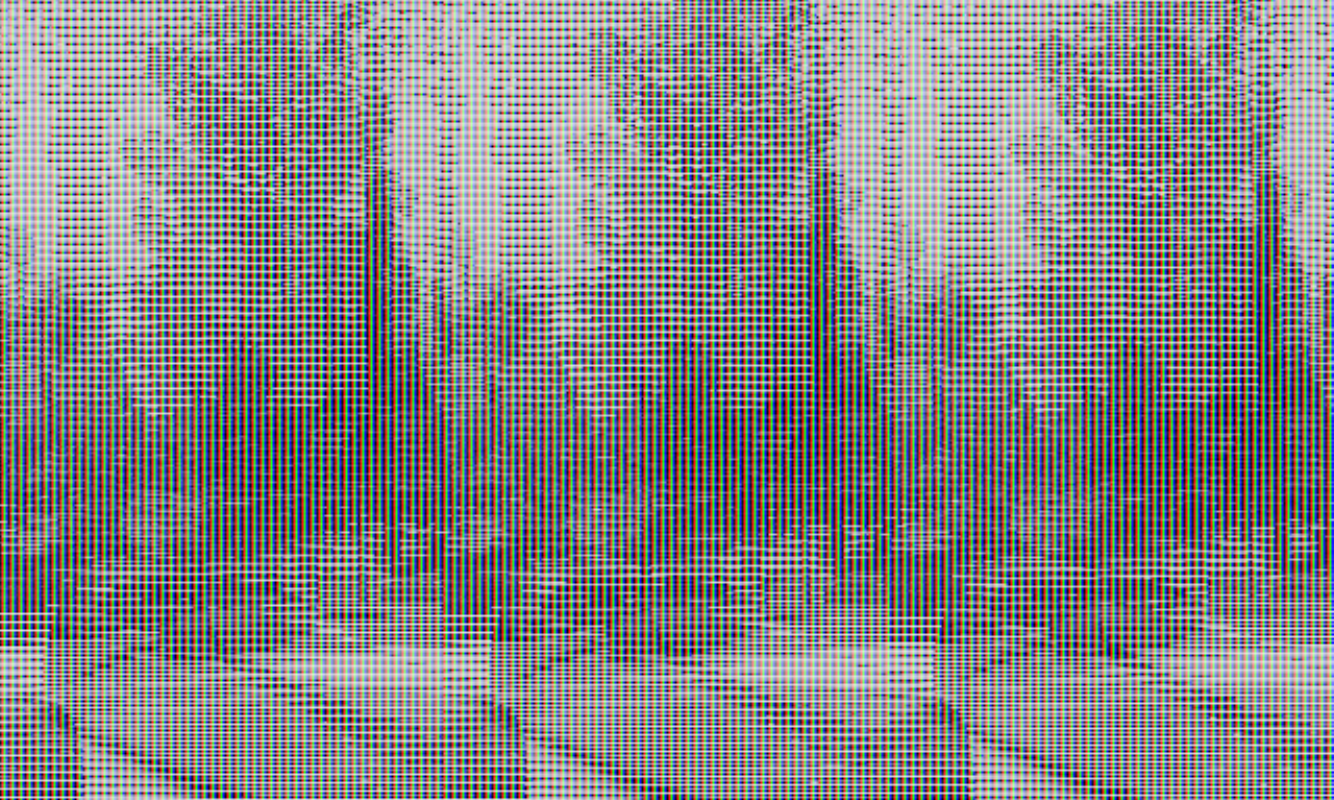}
   \caption{
The image depicts a snowy driving scenario. There are three cars visible: one in the foreground on the left, another in the middle ground, and a third one in the background on the right. The car in the foreground appears to be a sports car, possibly a convertible, given its sleek design and the absence of a roof. The middle ground shows a car that seems to be a sports car as well. The car in the background is partially obscured by the snow and appears to be a different model. The road is covered in a layer of snow, indicating recent snowfall or cold weather. The environment suggests \textcolor{dpp}{a residential area} with houses and trees, some of which are dusted with snow. The lighting in the image is soft and diffused, which could be due to the overcast sky or the time of day, possibly early morning or late afternoon. The overall scene is quiet and serene, with no visible movement or activity, which is typical for a snowy day when \textcolor{dpp}{roads are not busy.}}
\end{subfigure} \\
\hline

\hline
\rowcolor{lightgray}

\textbf{Sunny Highway} & \textbf{Raining Highway} & \textbf{Raining Highway at night} \\
\hline
\begin{subfigure}[t]{0.3\textwidth}
   \includegraphics[width=\textwidth]{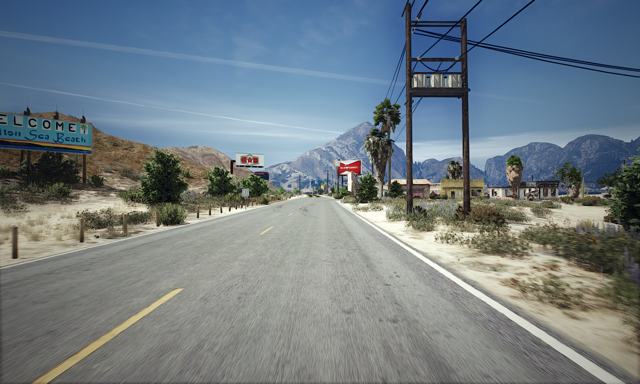}
   \caption{In the image, you see a driving scenario that appears to be from a video game or a simulation. The road is a two-lane highway with a clear, blue sky above and a desert-like environment with sparse vegetation on both sides. The road is marked with a yellow centerline, and there are no visible vehicles or pedestrians. On the left side of the road, there are billboards and a welcome sign that reads "Welcome to Sea Beach," suggesting the name of the location or the destination. On the right side, there is a utility pole with power lines and a ladder leading up to a platform, which is likely for maintenance purposes. The landscape is \textcolor{dpp}{flat} with distant mountains, and the overall setting is serene and \textcolor{dpp}{devoid of any immediate traffic} or activity. The image has a slightly blurred effect, which is common in video games to give a sense of motion or speed.}
\end{subfigure} &
\begin{subfigure}[t]{0.3\textwidth}
   \includegraphics[width=\textwidth]{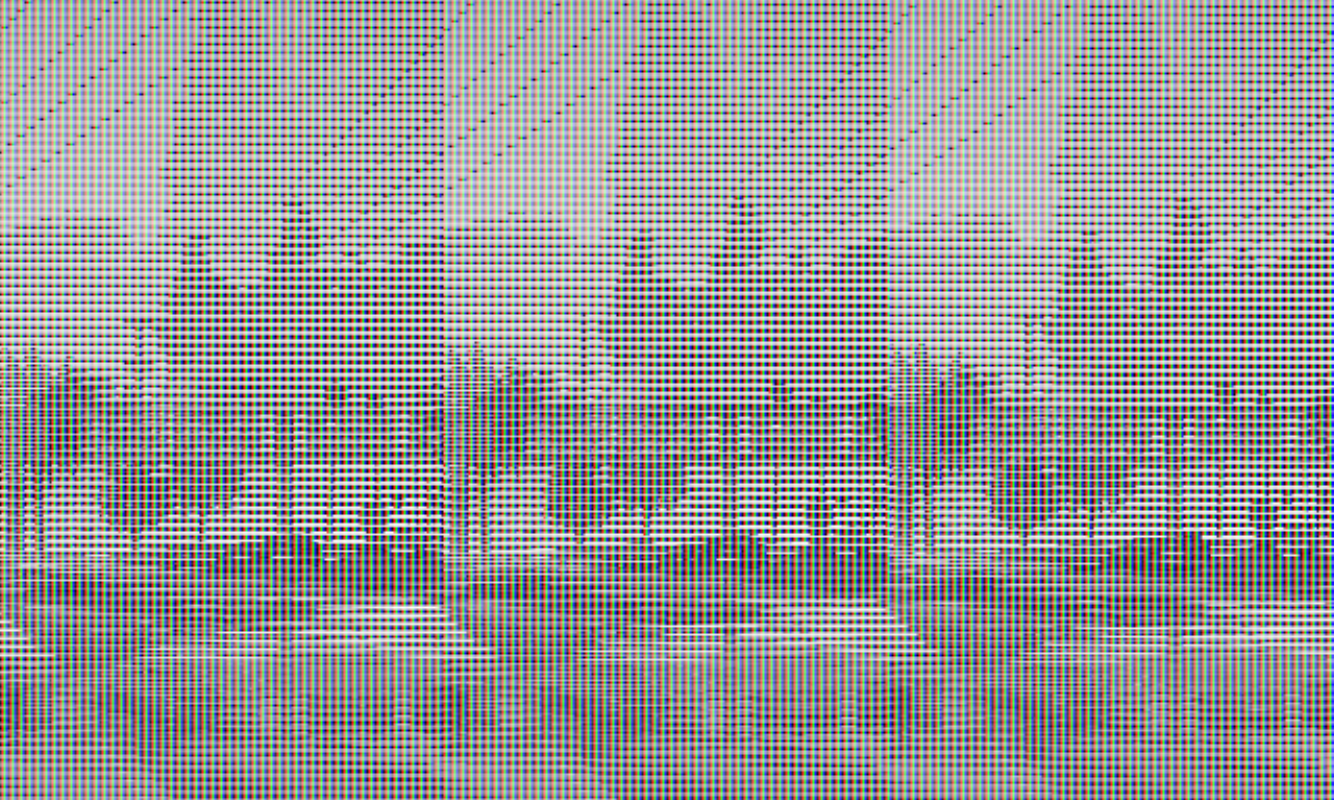}
   \caption{
The image depicts a driving scenario on a highway during what appears to be a rainy day. The road is wet, reflecting the light, and there are multiple lanes with vehicles traveling in both directions. The vehicles include a black car in the foreground, which is the main focus of the image, and other cars in the background. The highway is bordered by a barrier on the right side, and there are trees and a hillside on the left side, suggesting a coastal or hilly region. Above \textcolor{dpp}{the highway}, there is a bridge or overpass with a green sign that is not clearly legible. The sky is overcast, and the lighting suggests it could be either dawn or dusk. The overall atmosphere of the image is somewhat moody and atmospheric, with a focus on the motion of the vehicles and the \textcolor{dpp}{wet road conditions.}}
\end{subfigure} &
\begin{subfigure}[t]{0.3\textwidth}
   \includegraphics[width=\textwidth]{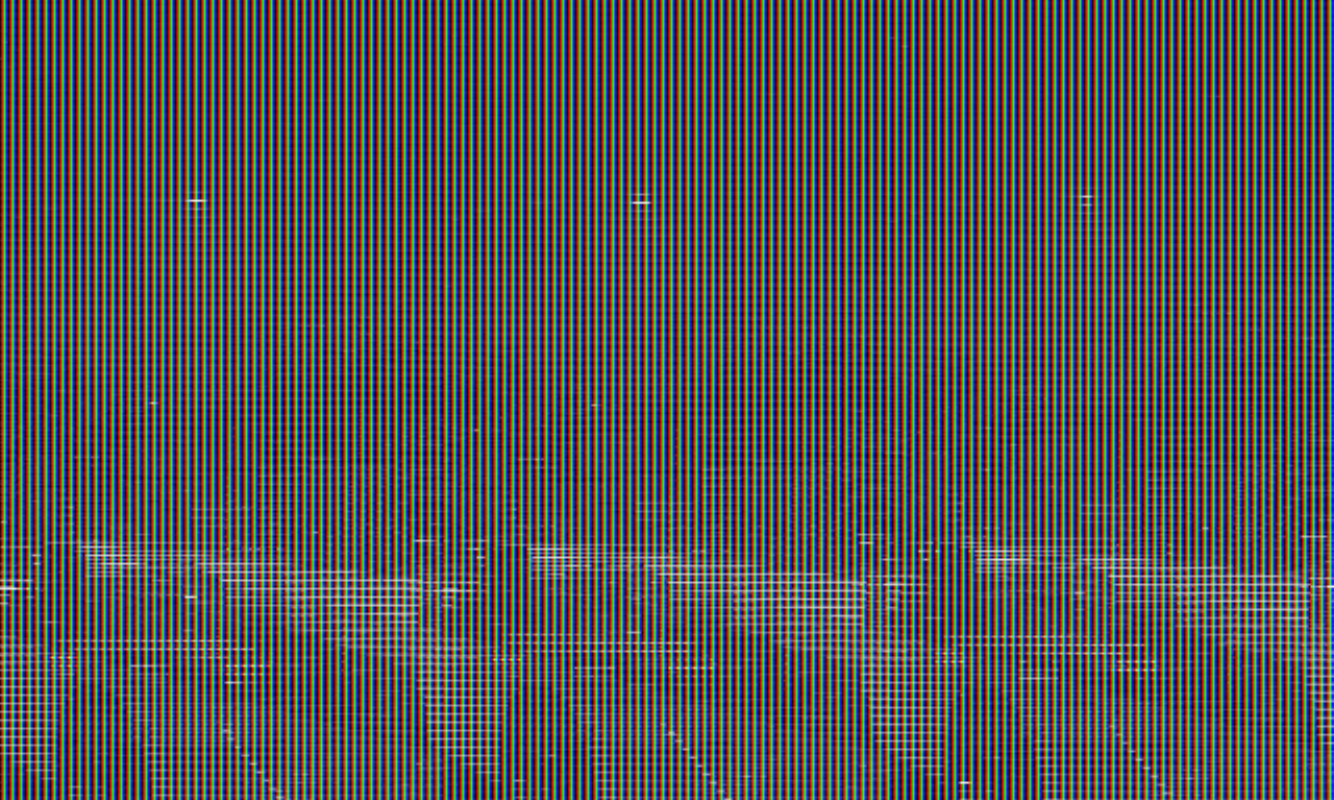}
   \caption{
The image depicts a nighttime driving scenario on a \textcolor{dpp}{road that appears to be wet}, possibly due to recent rain. The visibility is reduced, and the atmosphere is dark, with the streetlights casting a glow on the wet surface. There are several vehicles on the road, including a blue and yellow taxi in the foreground, a white bus in the background, and other cars in between. The vehicles \textcolor{dpp}{are moving} along the road, and the traffic seems to be \textcolor{dpp}{flowing}, albeit at a slower pace due to the wet conditions. The road is bordered by a barrier on the left side, and there are palm trees lining the right side, suggesting a location that could be in a region with a warm climate. The overall scene is typical of a city street during a rainy evening, with the \textcolor{dpp}{challenge} of navigating through the reduced visibility and wet road conditions.}
\end{subfigure} \\
\hline

\end{tabular}
\end{table}




\begin{table*}[!t]
\centering
\caption{\textbf{Qualitative Failure Cases on Depth.} We illustrate failure cases on nuScenes with Metric3Dv2 combined with Droid-SLAM. The left column displays the RGB input images, the middle column shows the corresponding depth predictions (darker pixels indicate closer distances), and the right column compares trajectory estimations. The green boxes in the depth images highlight regions where Metric3Dv2 struggles due to environmental factors such as sky reflections, clouds, lens artifacts, and glass surfaces. These challenges lead to significant trajectory drifts.}
\label{tab:nusc_depth}
\begin{tabular}{|c|c|c|}
\hline
\rowcolor{lightgray}
\textbf{RGB Image} & \textbf{Wrong Depth Image} & \textbf{Plotted Trajectory} \\

\hline
\begin{subfigure}[b]{0.3\textwidth}
   \includegraphics[width=\textwidth]{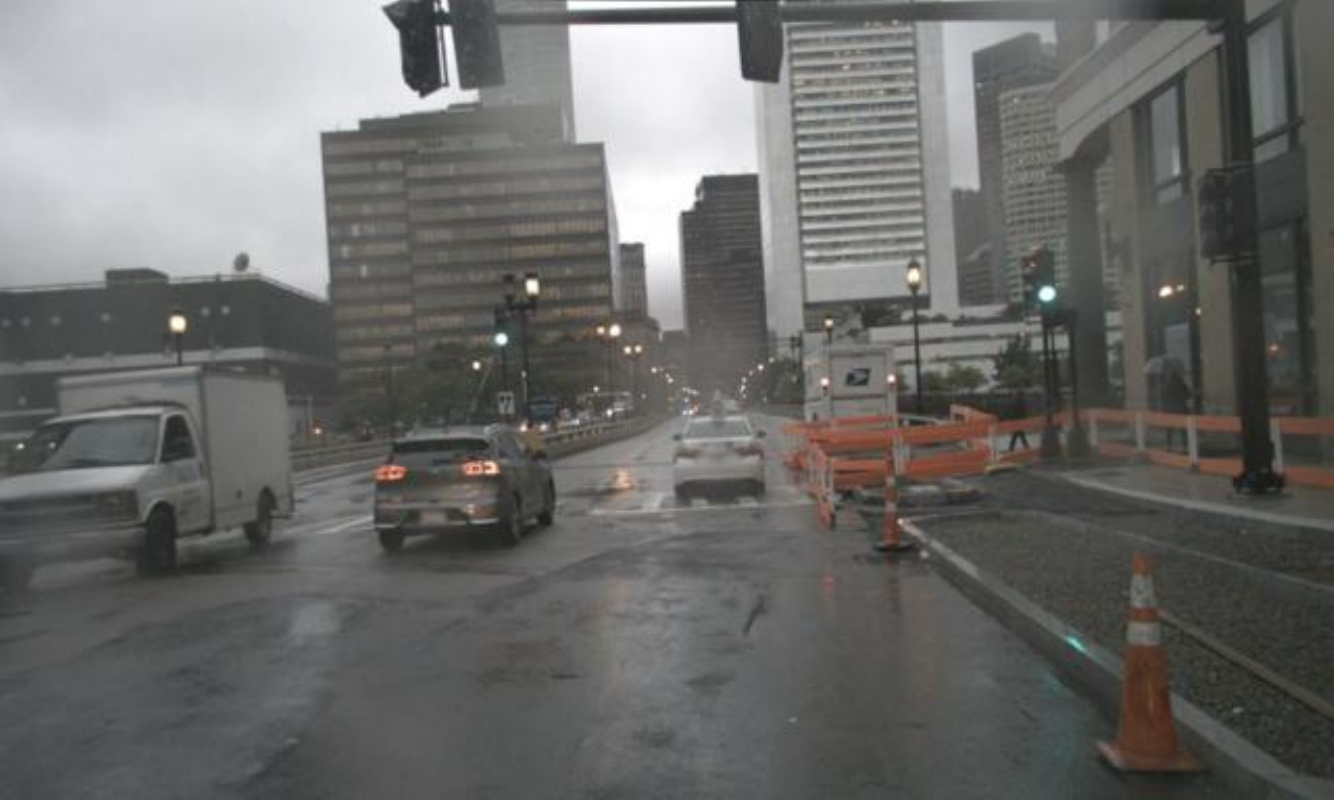}
\end{subfigure} &
\begin{subfigure}[b]{0.3\textwidth}
   \includegraphics[width=\textwidth]{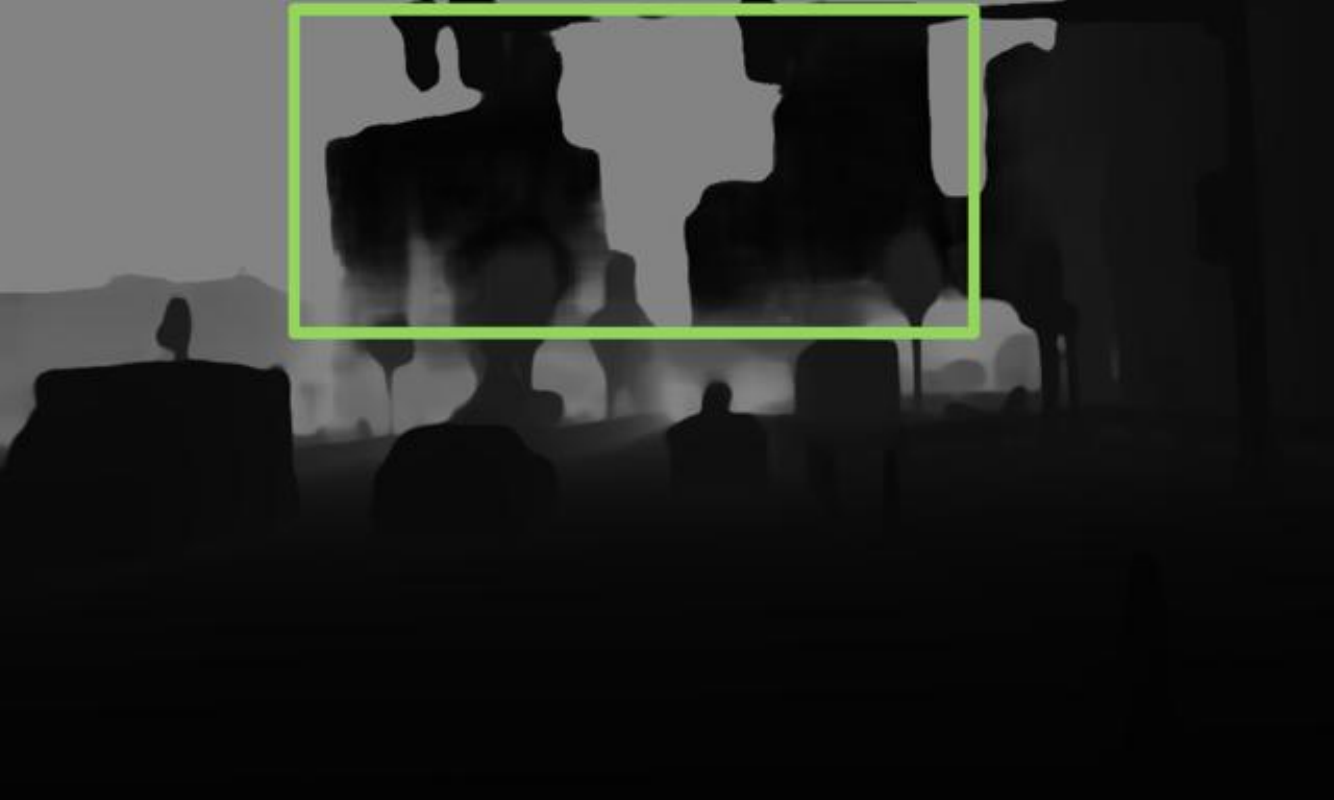}
\end{subfigure} &
\begin{subfigure}[b]{0.3\textwidth}
   \includegraphics[width=\textwidth]{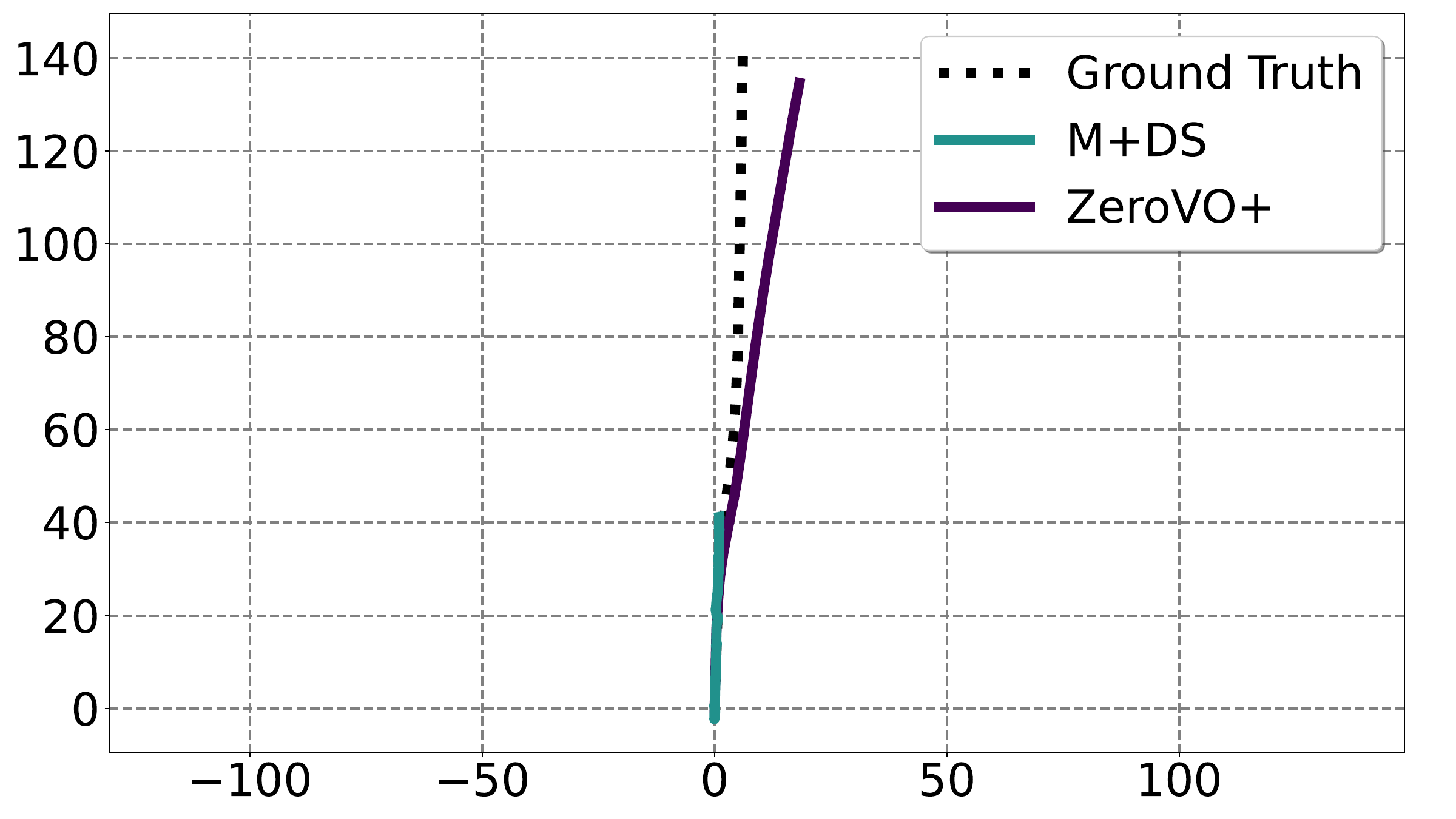}
\end{subfigure} \\
\hline

\hline
\begin{subfigure}[b]{0.3\textwidth}
   \includegraphics[width=\textwidth]{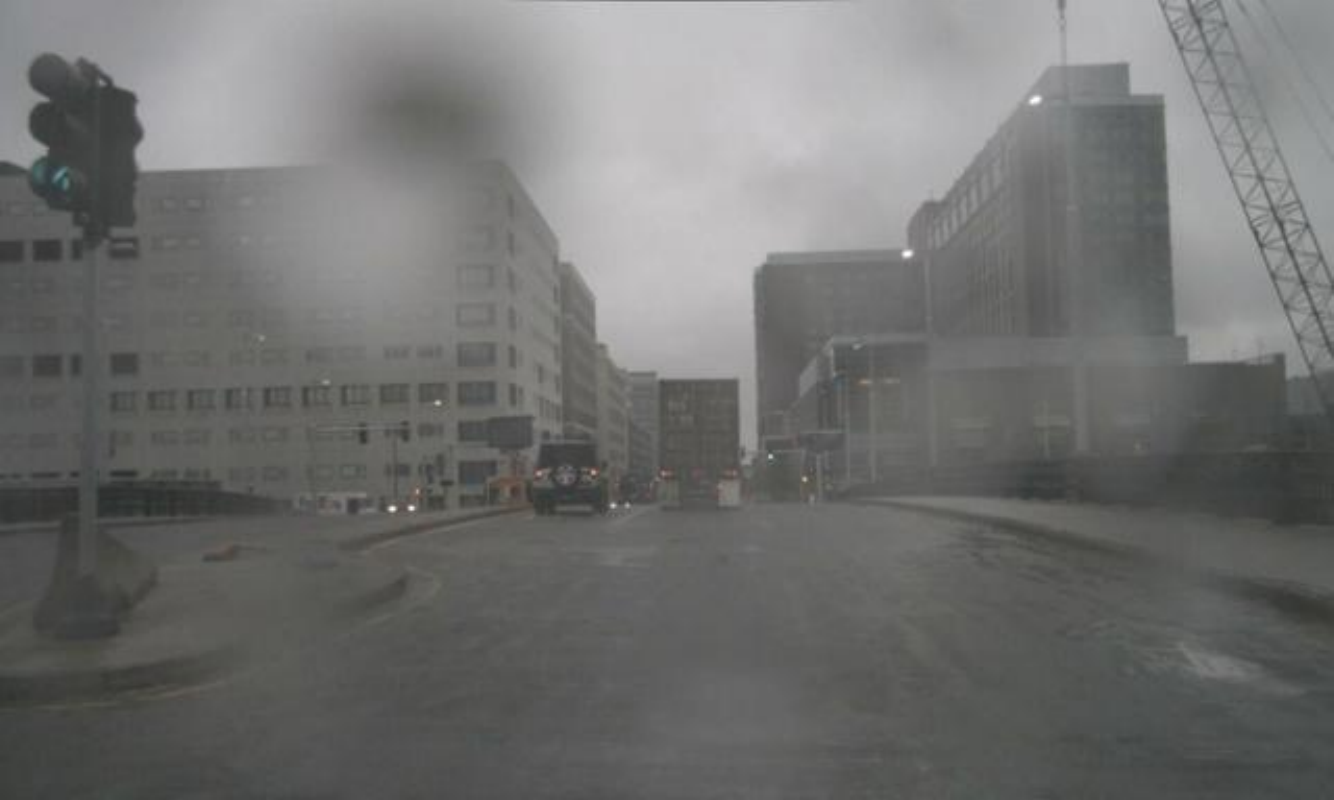}
\end{subfigure} &
\begin{subfigure}[b]{0.3\textwidth}
   \includegraphics[width=\textwidth]{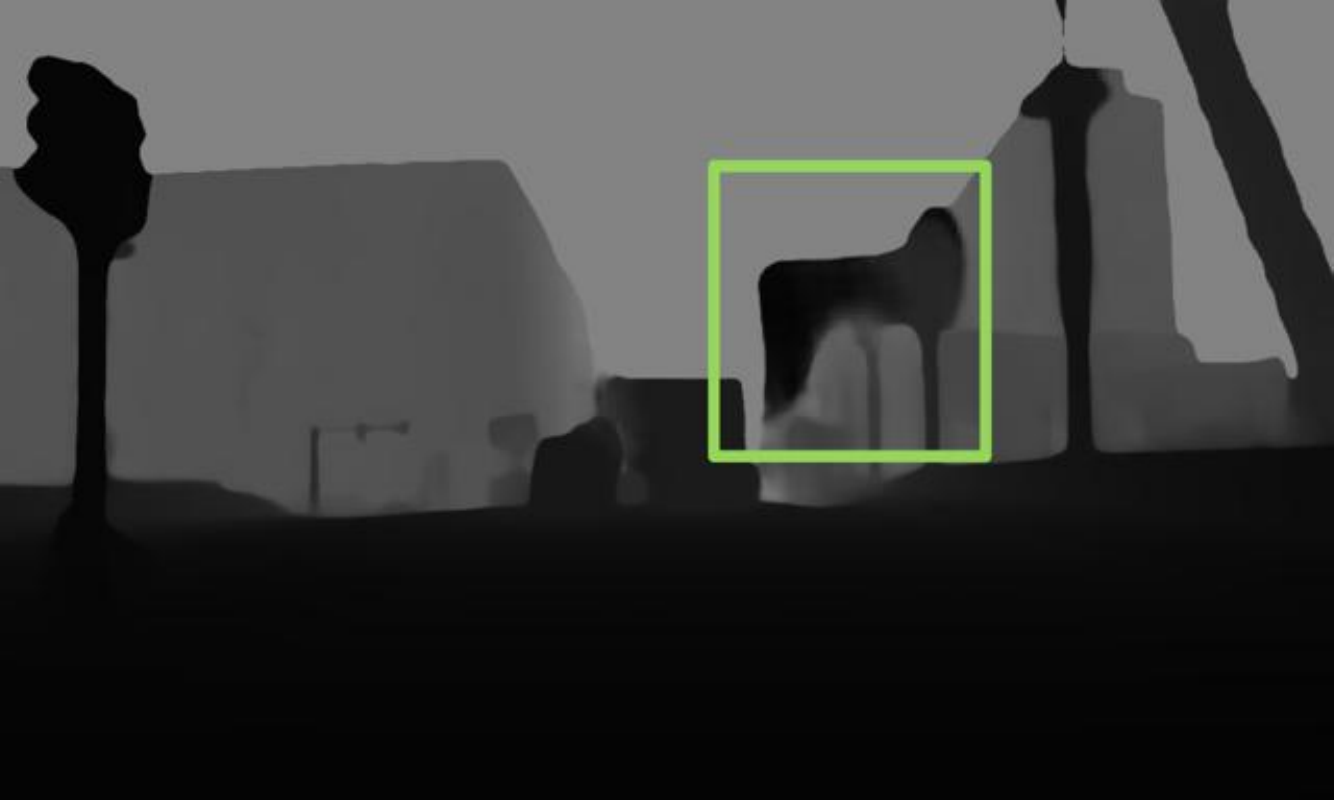}
\end{subfigure} &
\begin{subfigure}[b]{0.3\textwidth}
   \includegraphics[width=\textwidth]{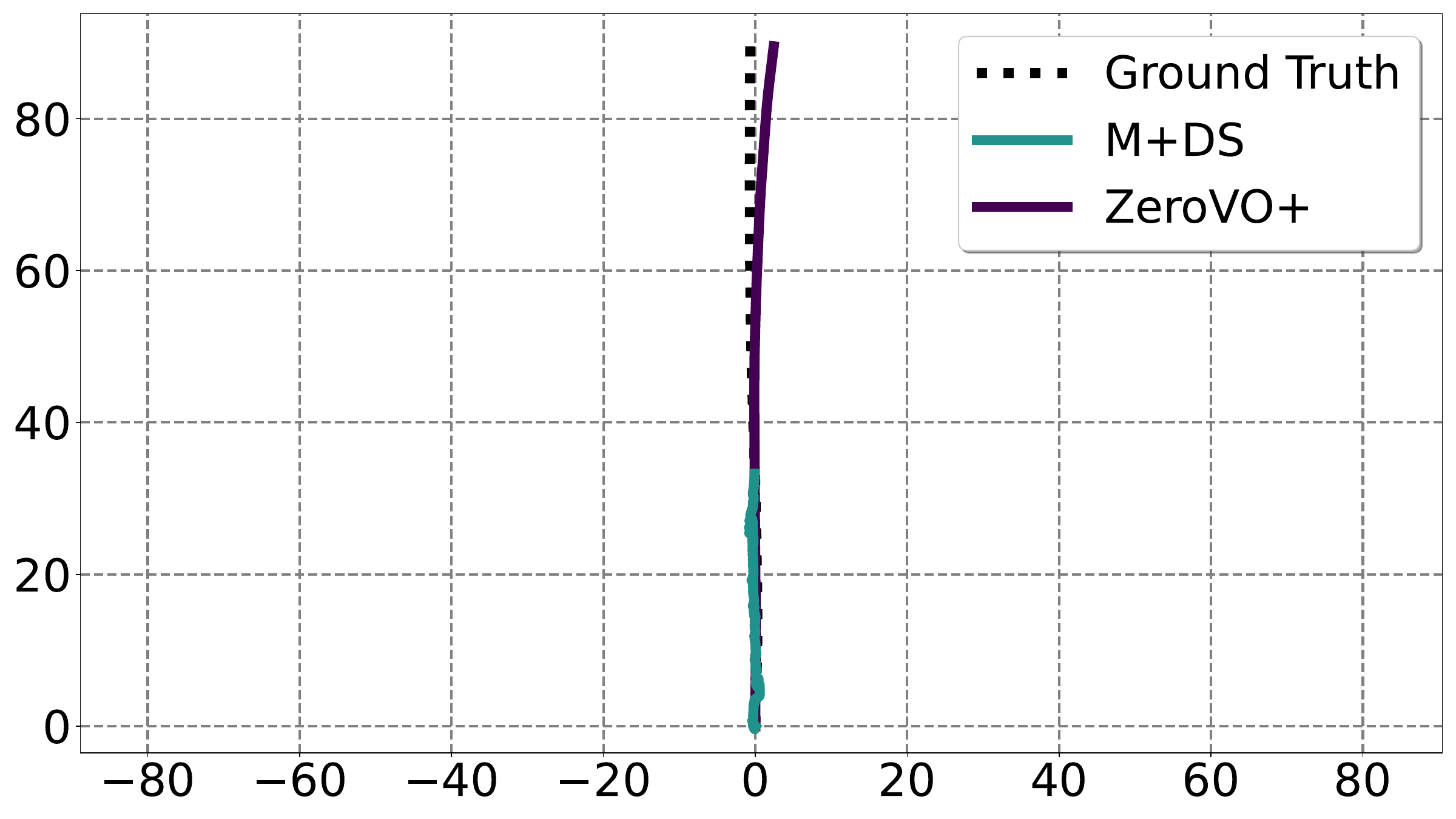}
\end{subfigure} \\

\hline

\hline
\begin{subfigure}[b]{0.3\textwidth}
   \includegraphics[width=\textwidth]{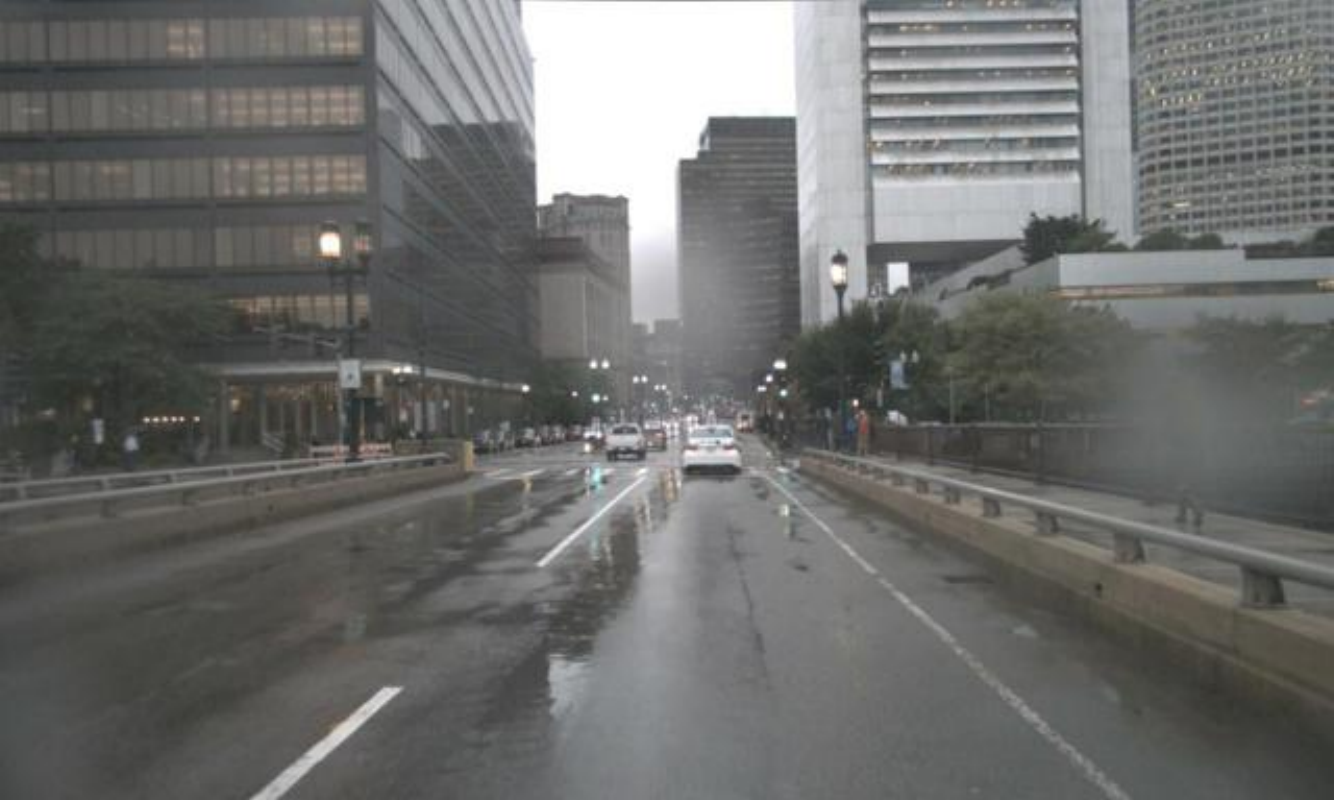}
\end{subfigure} &
\begin{subfigure}[b]{0.3\textwidth}
   \includegraphics[width=\textwidth]{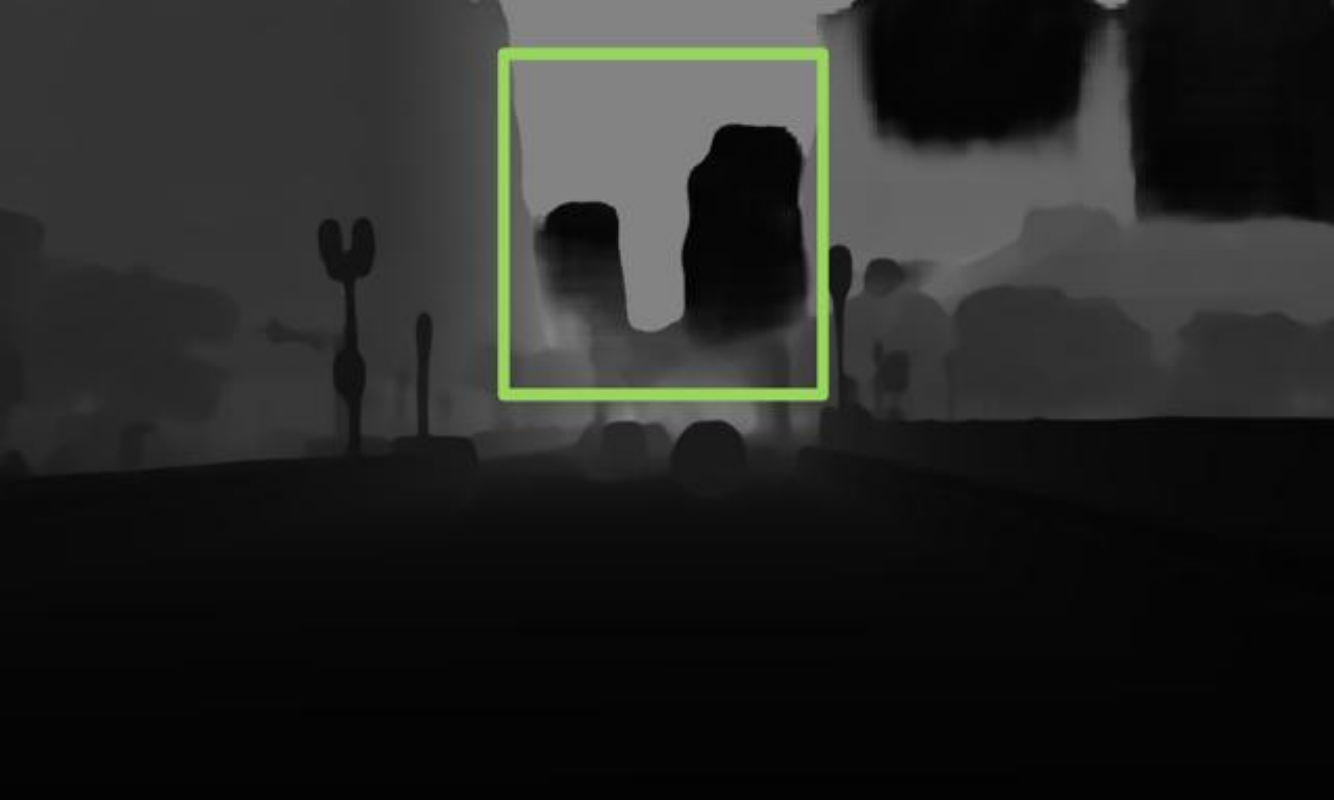}
\end{subfigure} &
\begin{subfigure}[b]{0.3\textwidth}
   \includegraphics[width=\textwidth]{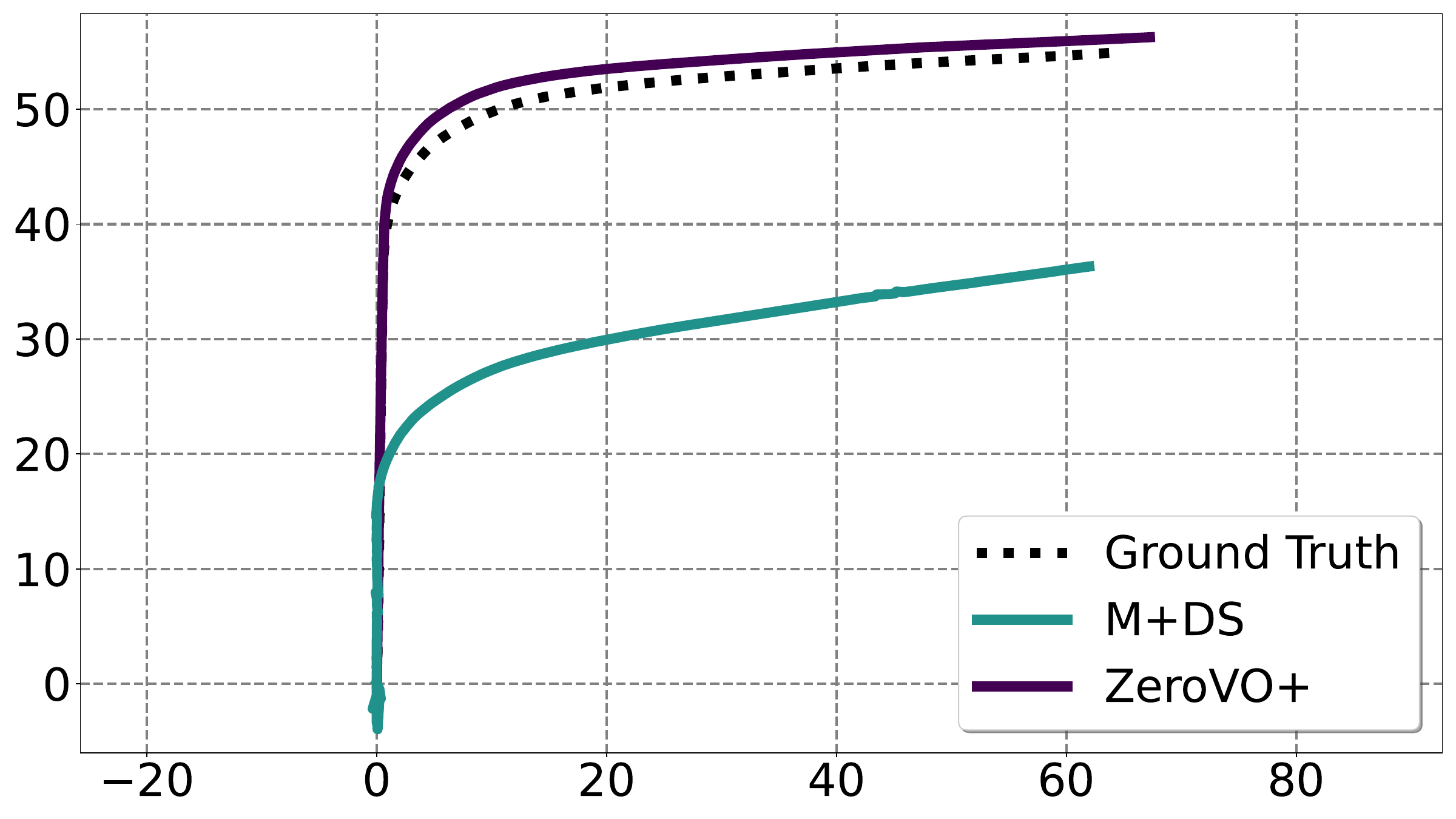}
\end{subfigure} \\
\hline

\hline
\begin{subfigure}[b]{0.3\textwidth}
   \includegraphics[width=\textwidth]{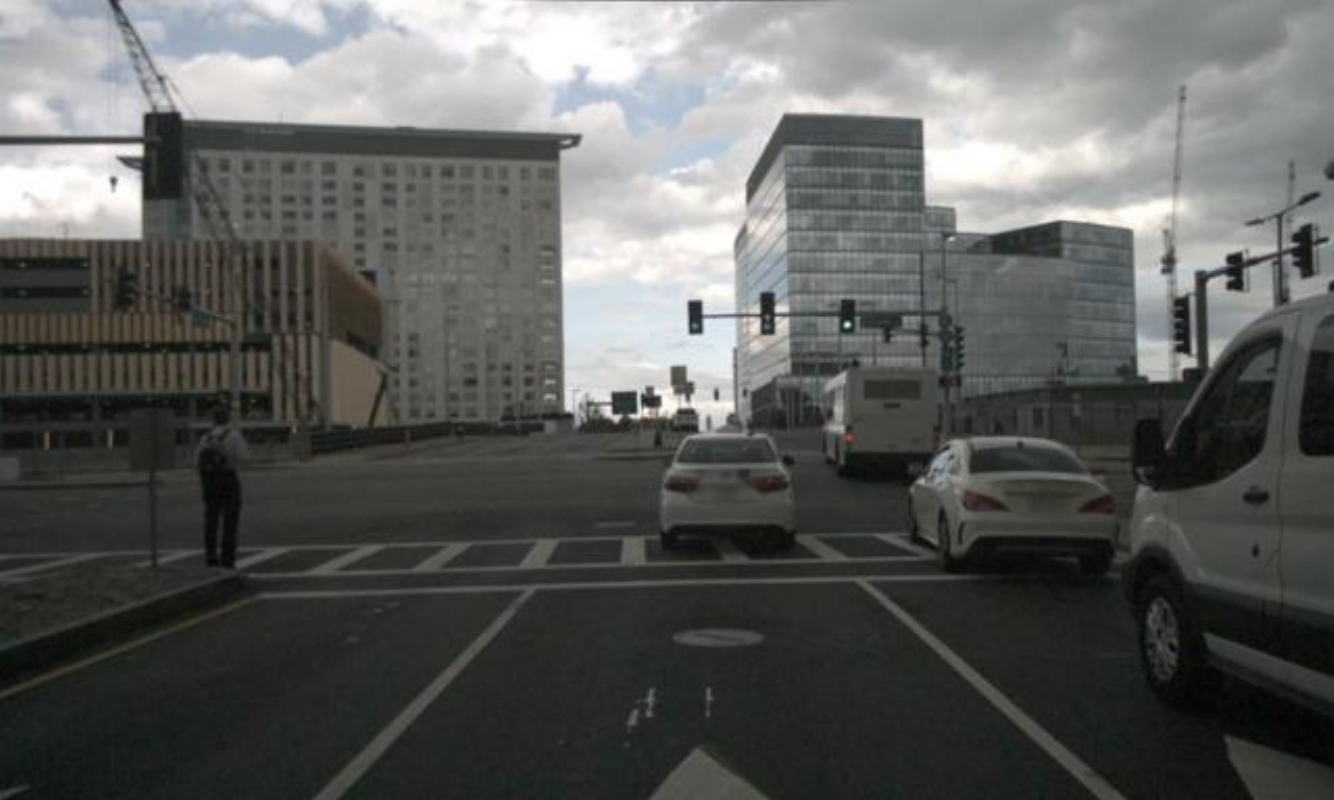}
\end{subfigure} &
\begin{subfigure}[b]{0.3\textwidth}
   \includegraphics[width=\textwidth]{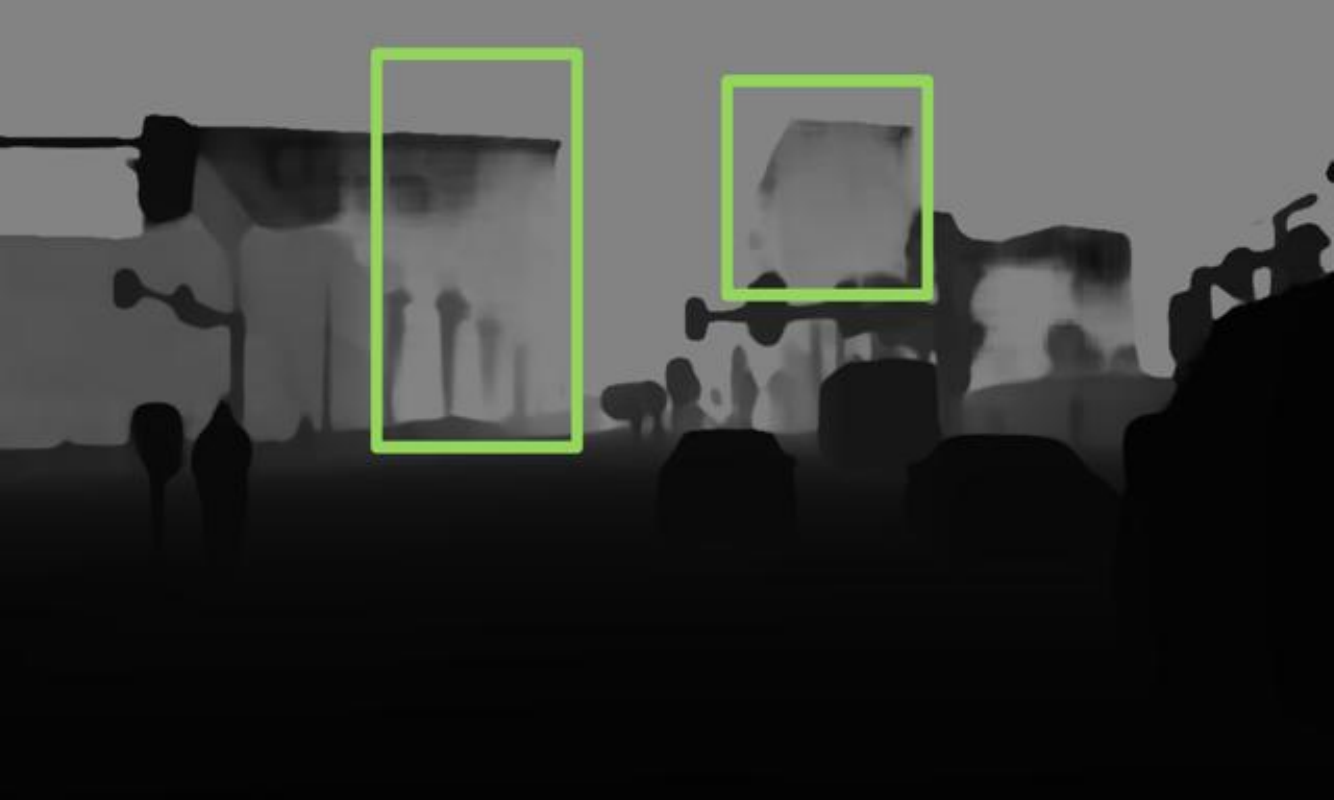}
\end{subfigure} &
\begin{subfigure}[b]{0.3\textwidth}
   \includegraphics[width=\textwidth]{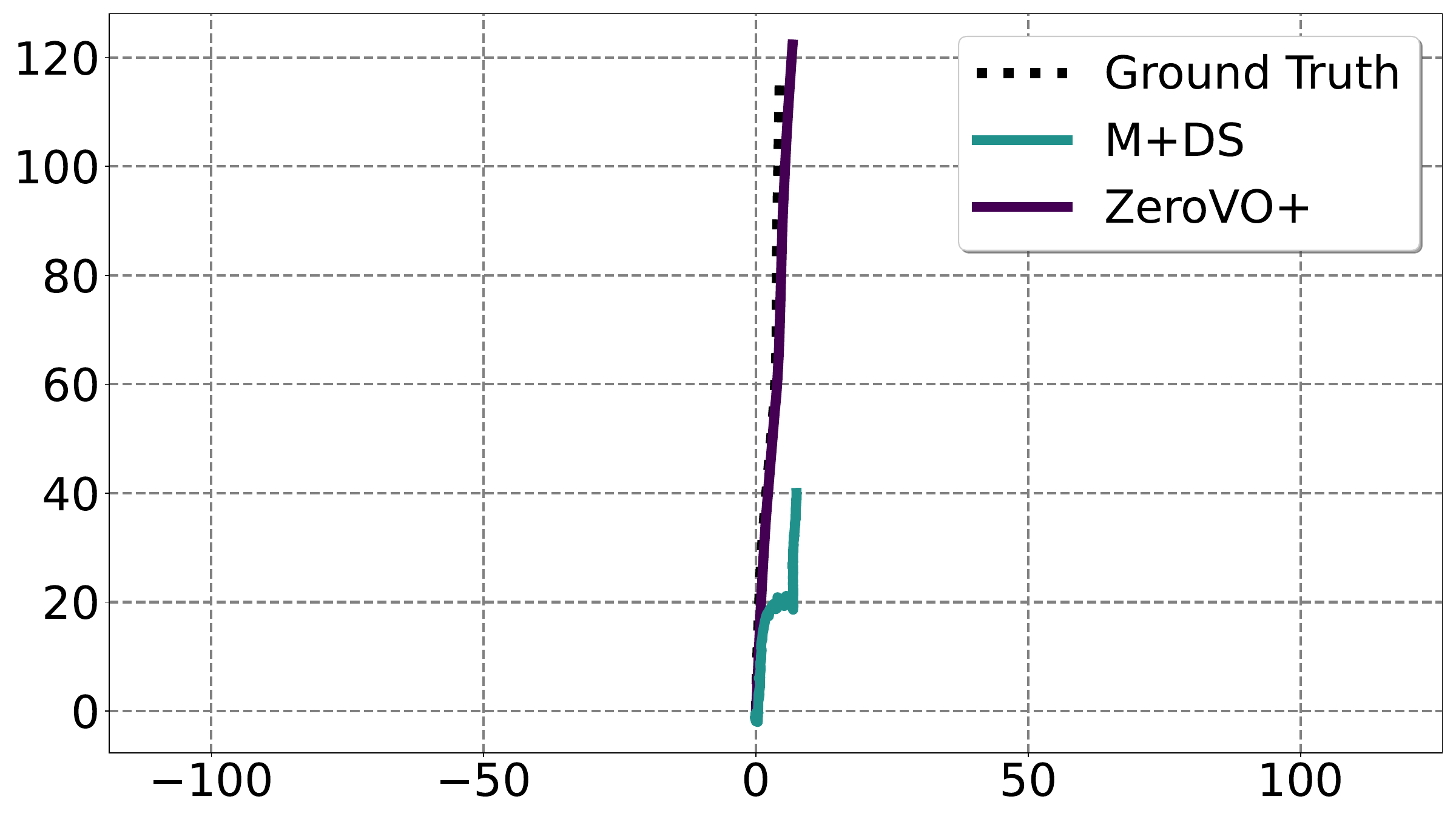}
\end{subfigure} \\
\hline

\hline
\begin{subfigure}[b]{0.3\textwidth}
   \includegraphics[width=\textwidth]{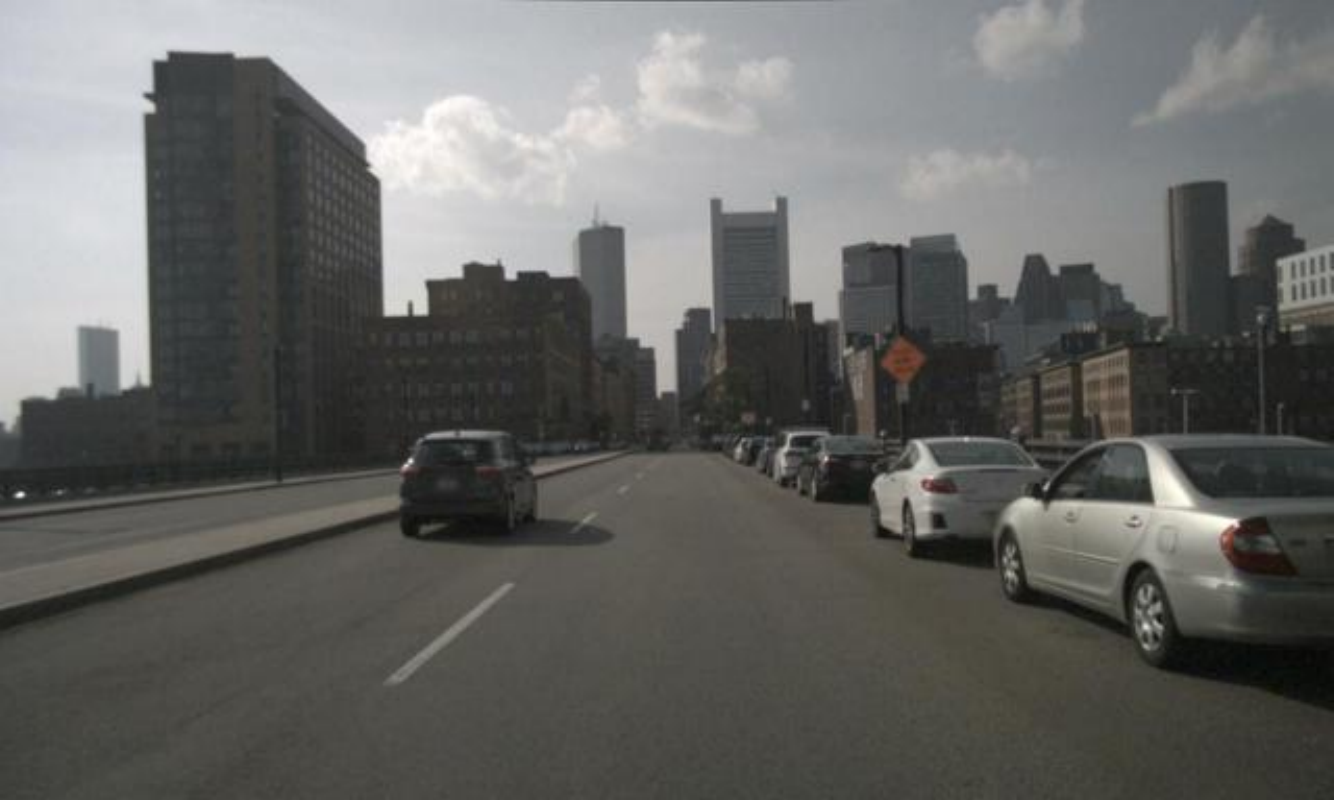}
\end{subfigure} &
\begin{subfigure}[b]{0.3\textwidth}
   \includegraphics[width=\textwidth]{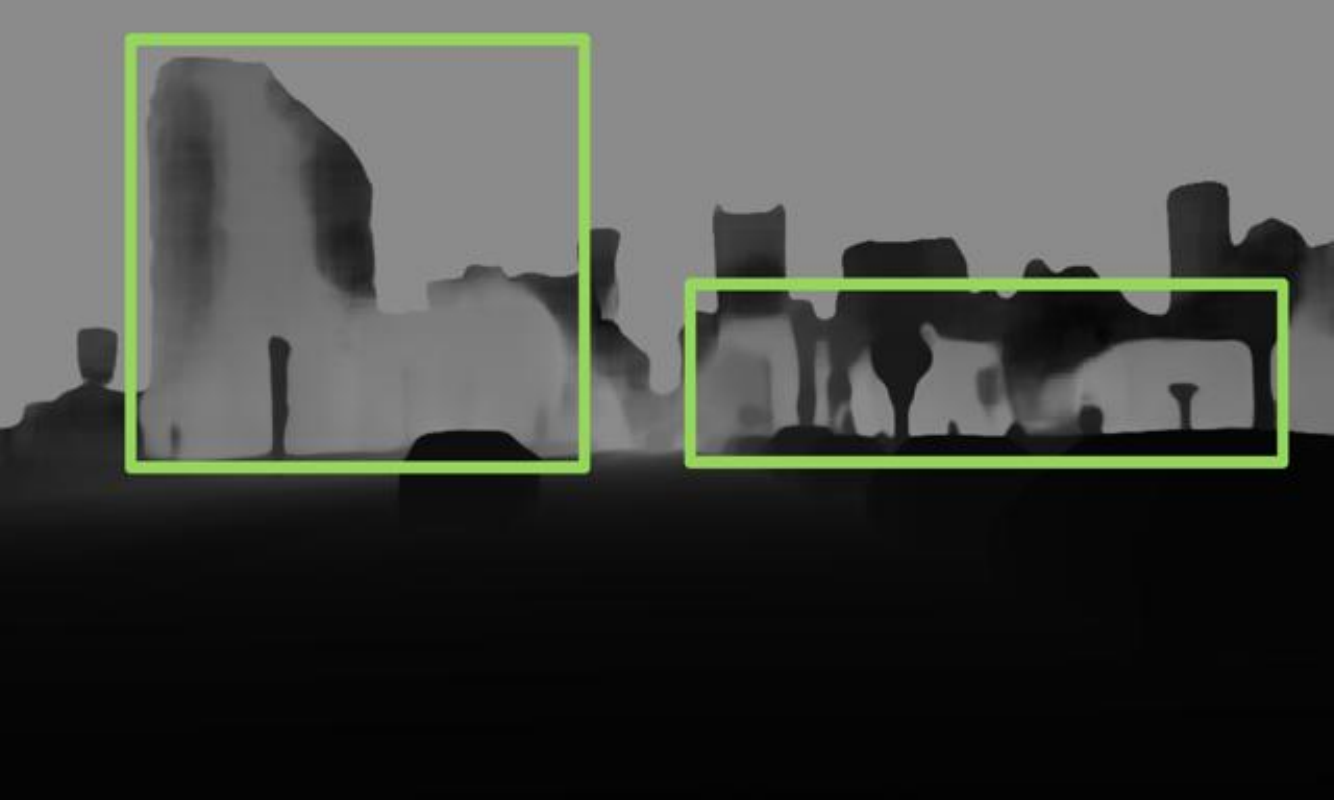}
\end{subfigure} &
\begin{subfigure}[b]{0.3\textwidth}
   \includegraphics[width=\textwidth]{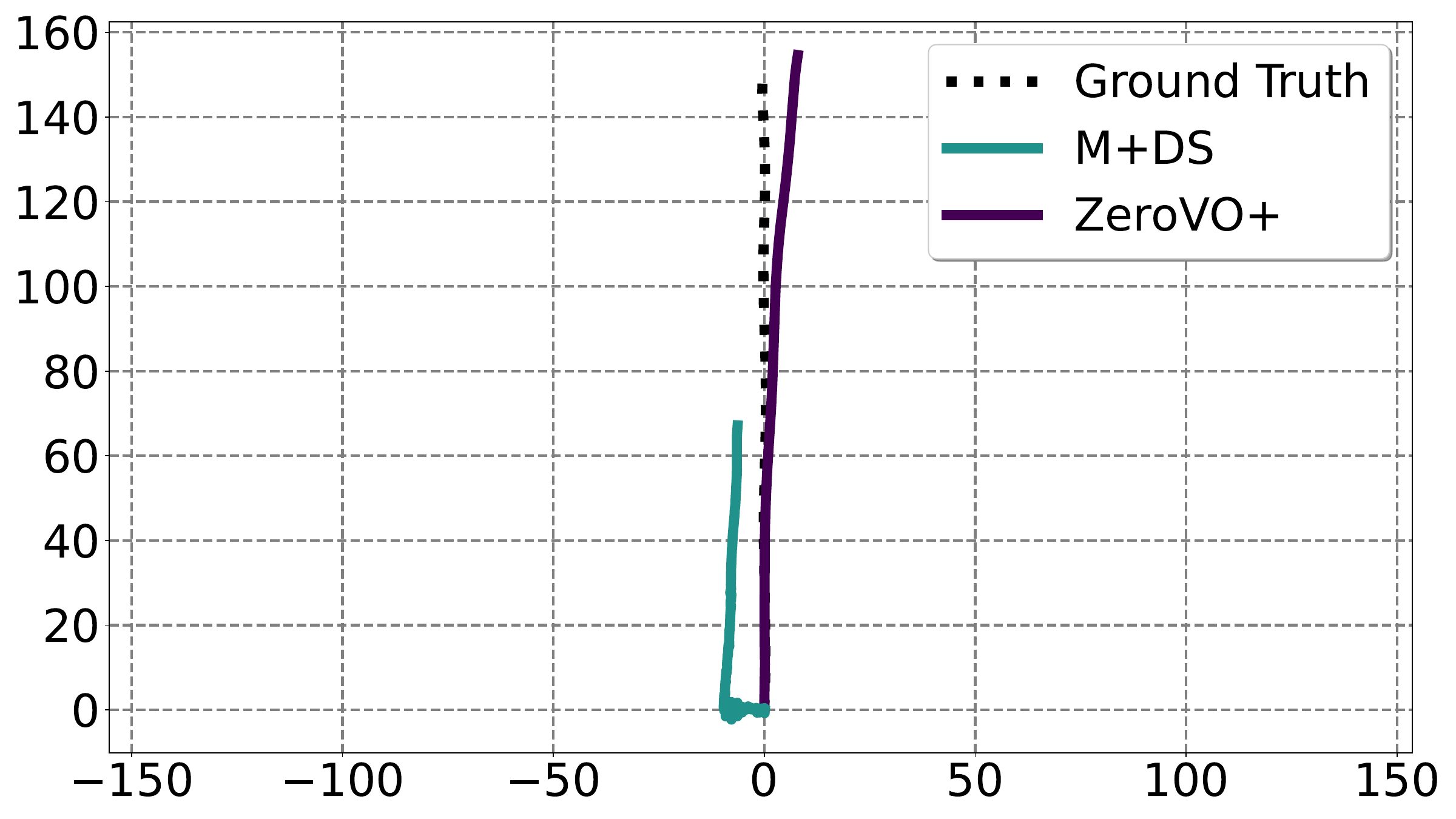}
\end{subfigure} \\
\hline

\hline
\begin{subfigure}[b]{0.3\textwidth}
   \includegraphics[width=\textwidth]{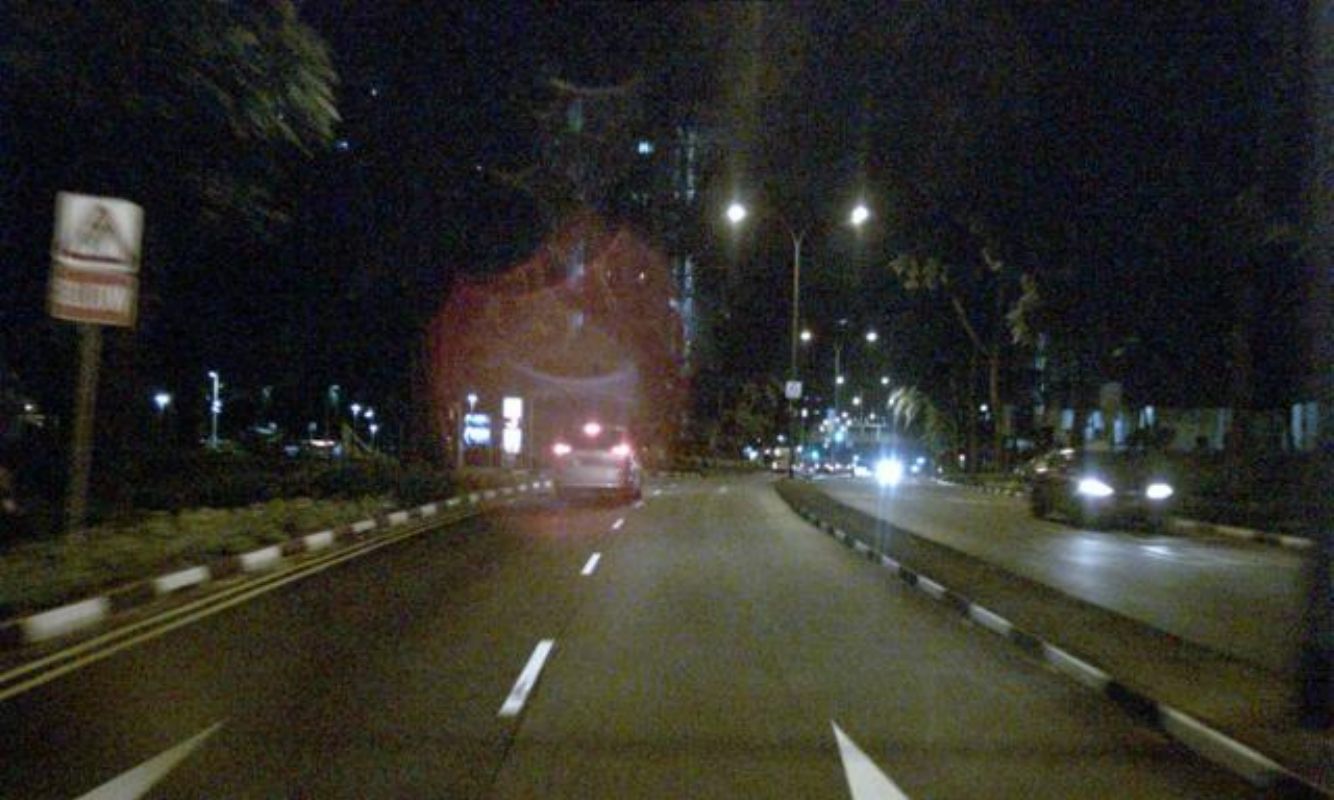}
\end{subfigure} &
\begin{subfigure}[b]{0.3\textwidth}
   \includegraphics[width=\textwidth]{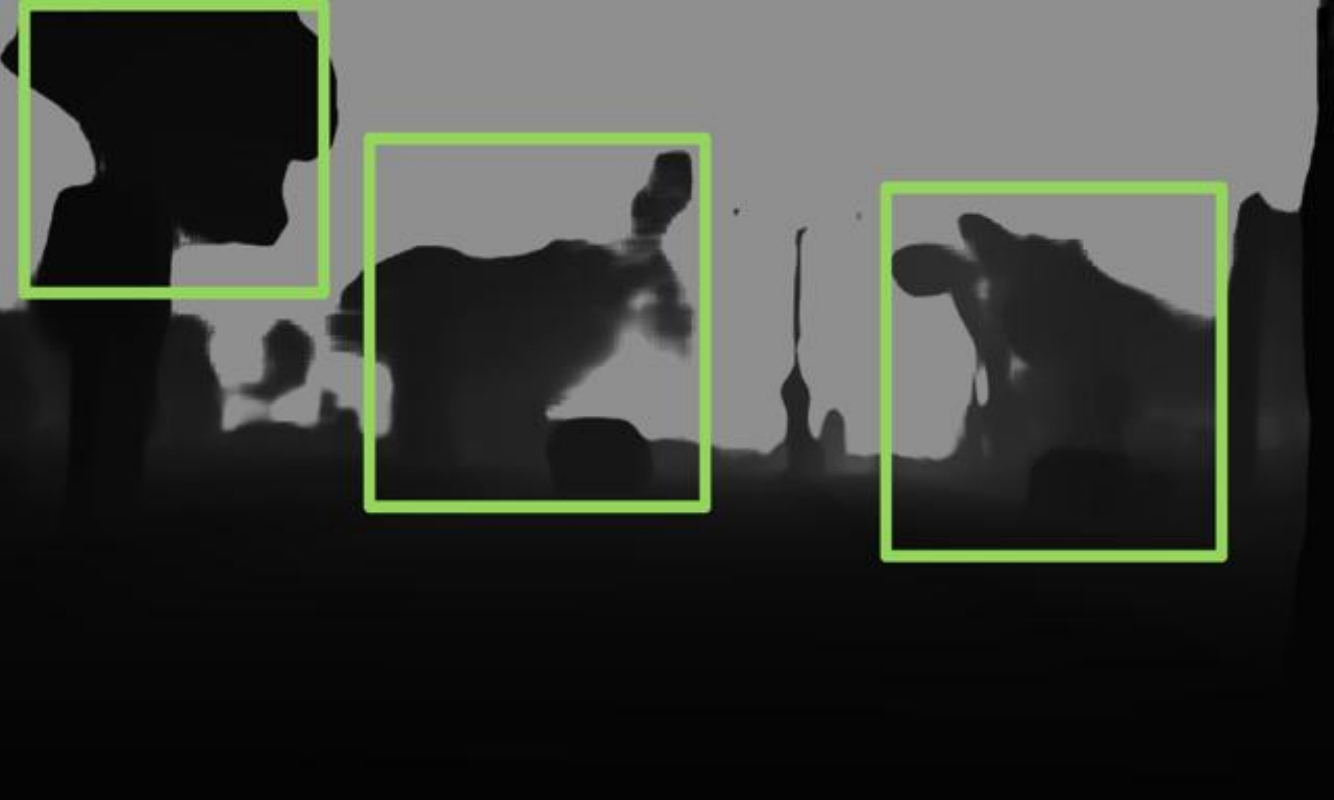}
\end{subfigure} &
\begin{subfigure}[b]{0.3\textwidth}
   \includegraphics[width=\textwidth]{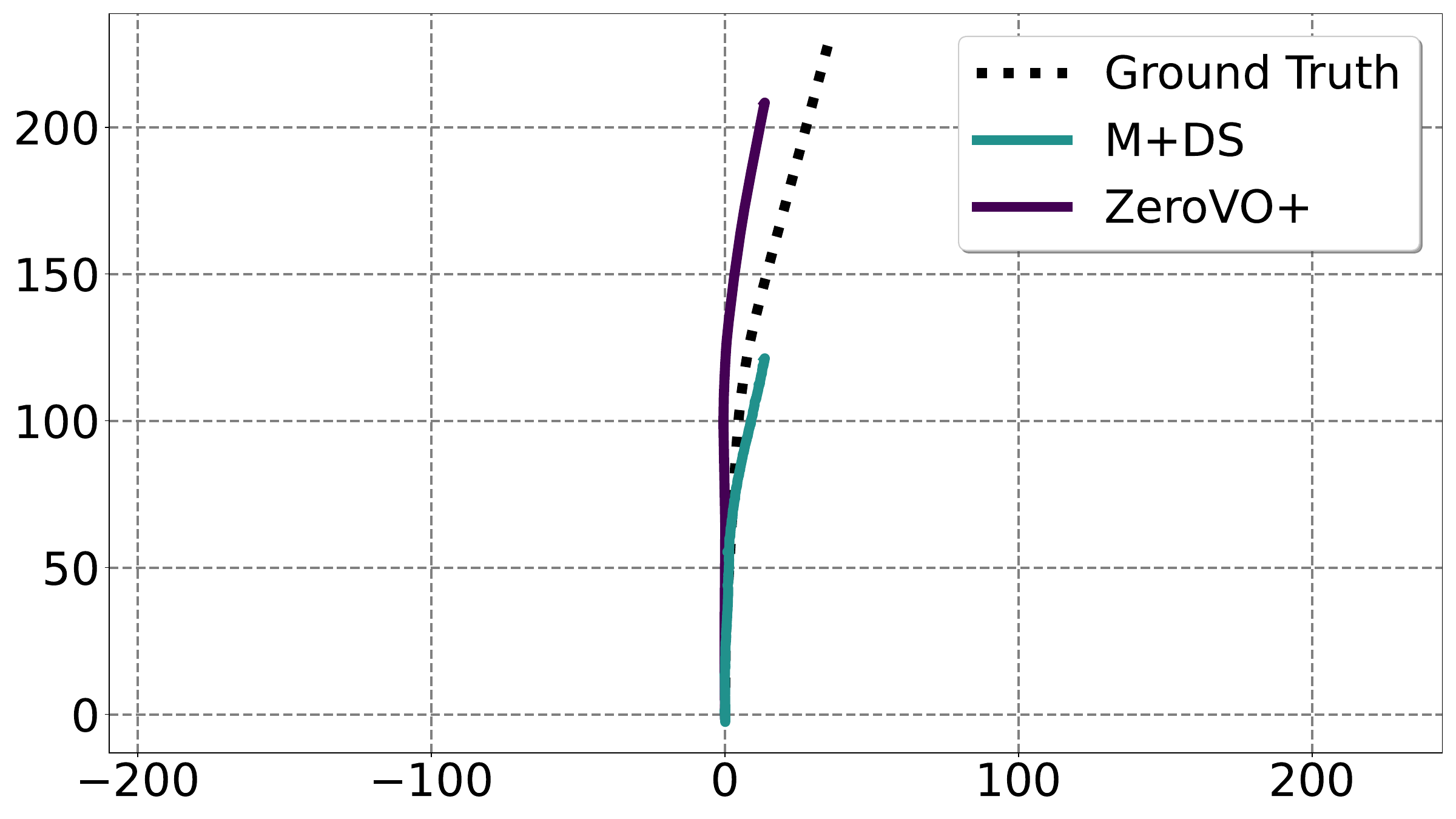}
\end{subfigure} \\
\hline

\end{tabular}

\end{table*}

\begin{table*}[!t]
\centering
\caption{\textbf{Failure Cases on GTA.} This figure illustrates qualitative results of failure cases on Argoverse with Metric3Dv2 combined with Droid-SLAM. The left column displays the RGB input images, the middle column shows the corresponding depth predictions (darker pixels indicate closer distances), and the right column compares trajectory estimations. The green boxes in the depth images highlight regions where Metric3Dv2 struggles due to environmental factors such as sky reflections, clouds, lens artifacts, and glass surfaces. These challenges lead to significant trajectory drifts. }
\label{tab:gta_fail}
\begin{tabular}{|c|c|c|}
\hline
\rowcolor{lightgray}
\textbf{RGB Image} & \textbf{Wrong Depth Image} & \textbf{Plotted Trajectory} \\

\hline
\begin{subfigure}[b]{0.3\textwidth}
   \includegraphics[width=\textwidth]{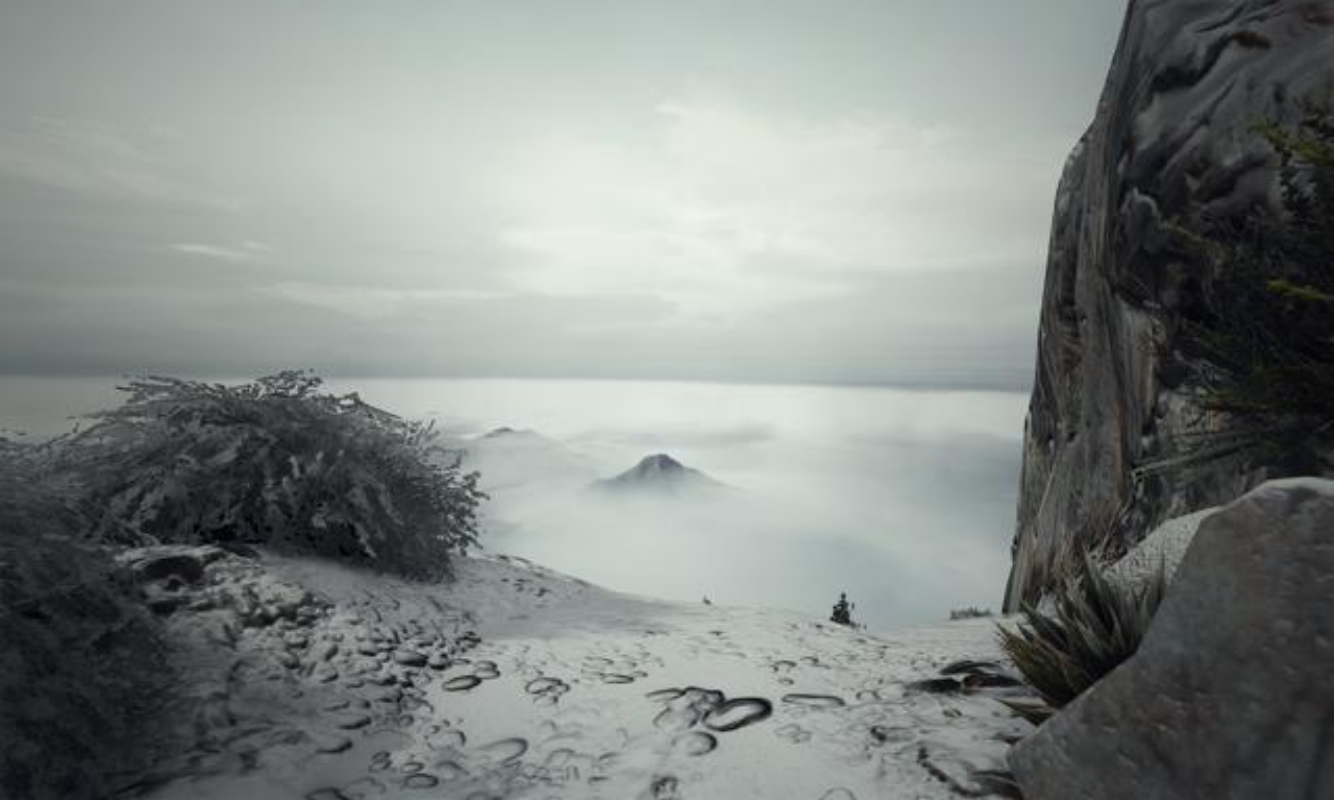}
\end{subfigure} &
\begin{subfigure}[b]{0.3\textwidth}
   \includegraphics[width=\textwidth]{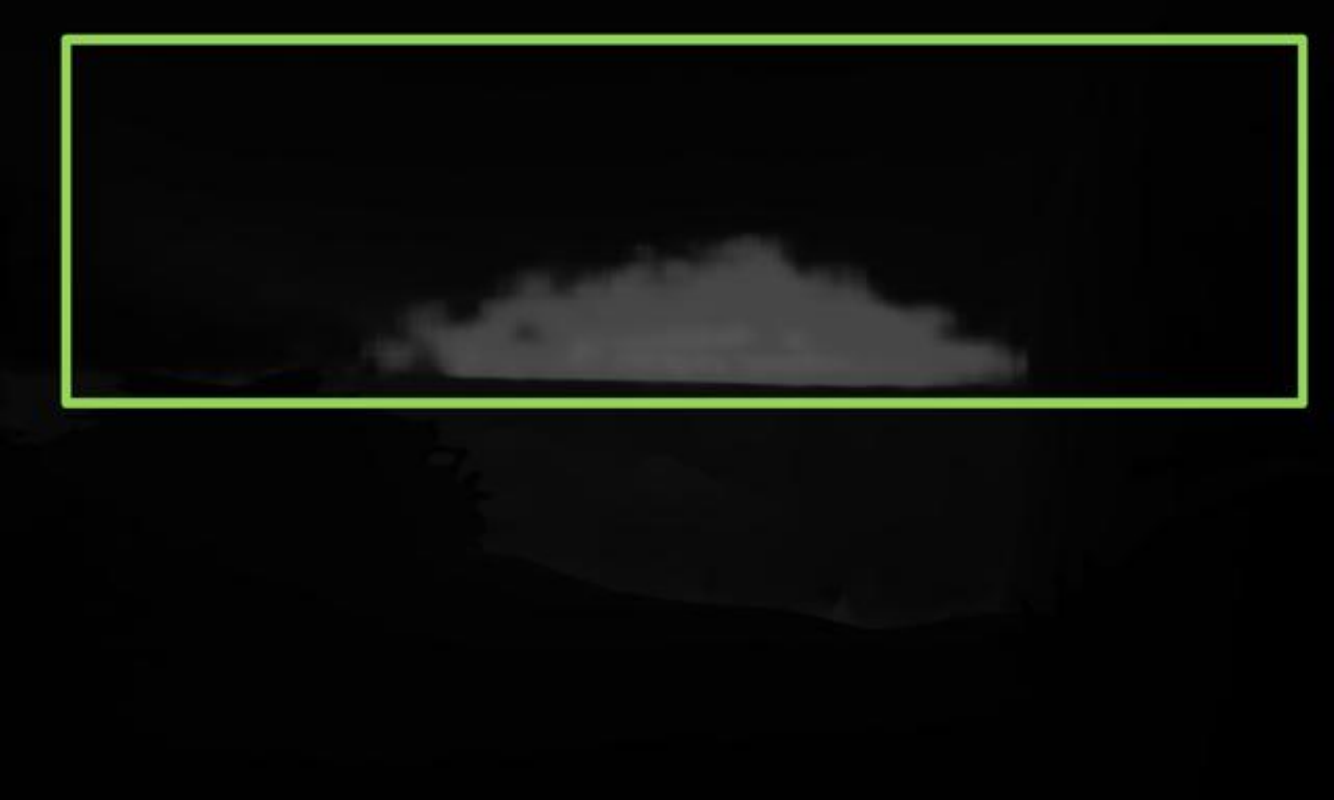}
\end{subfigure} &
\begin{subfigure}[b]{0.3\textwidth}
   \includegraphics[width=\textwidth]{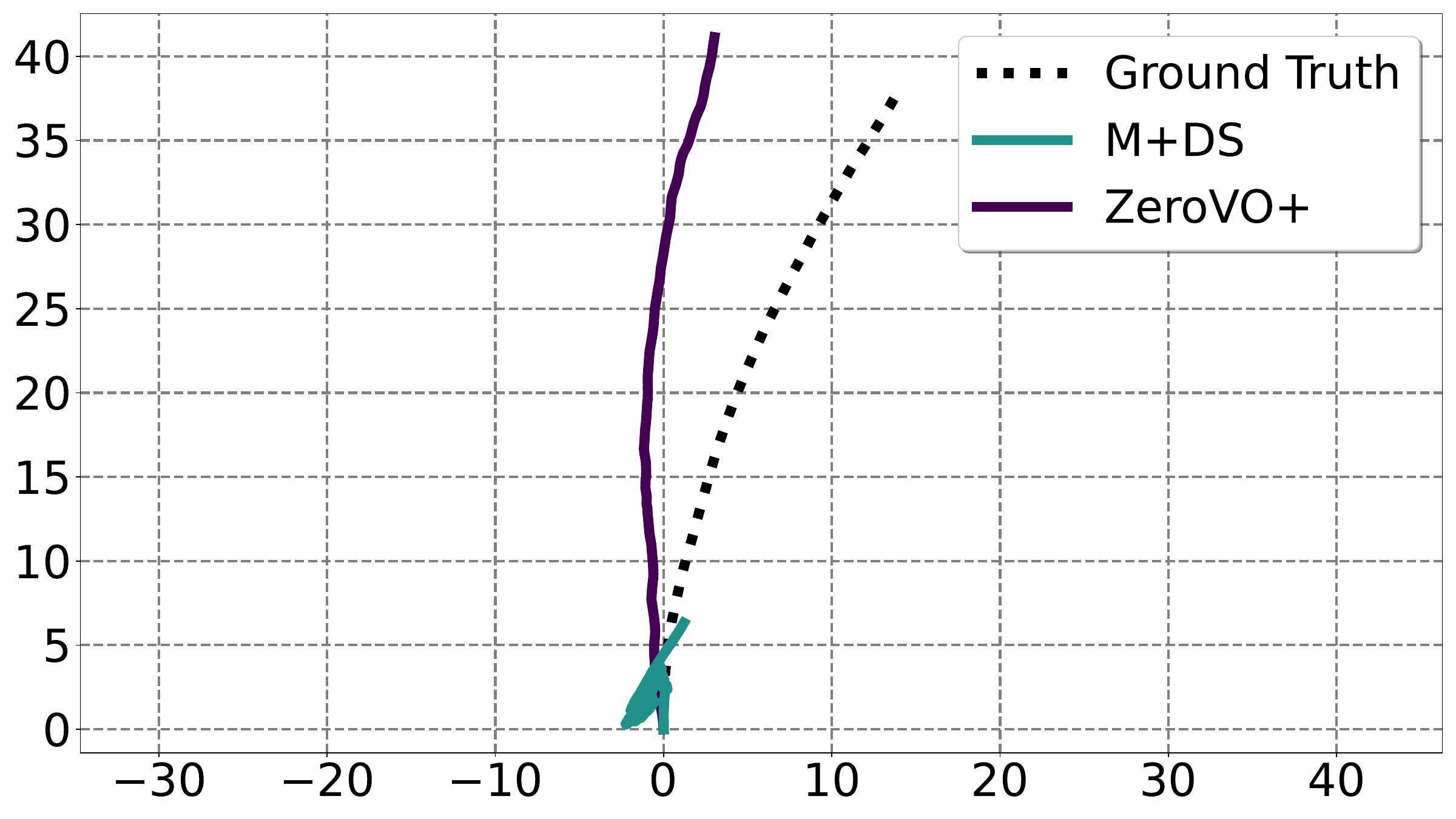}
\end{subfigure} \\
\hline

\hline
\begin{subfigure}[b]{0.3\textwidth}
   \includegraphics[width=\textwidth]{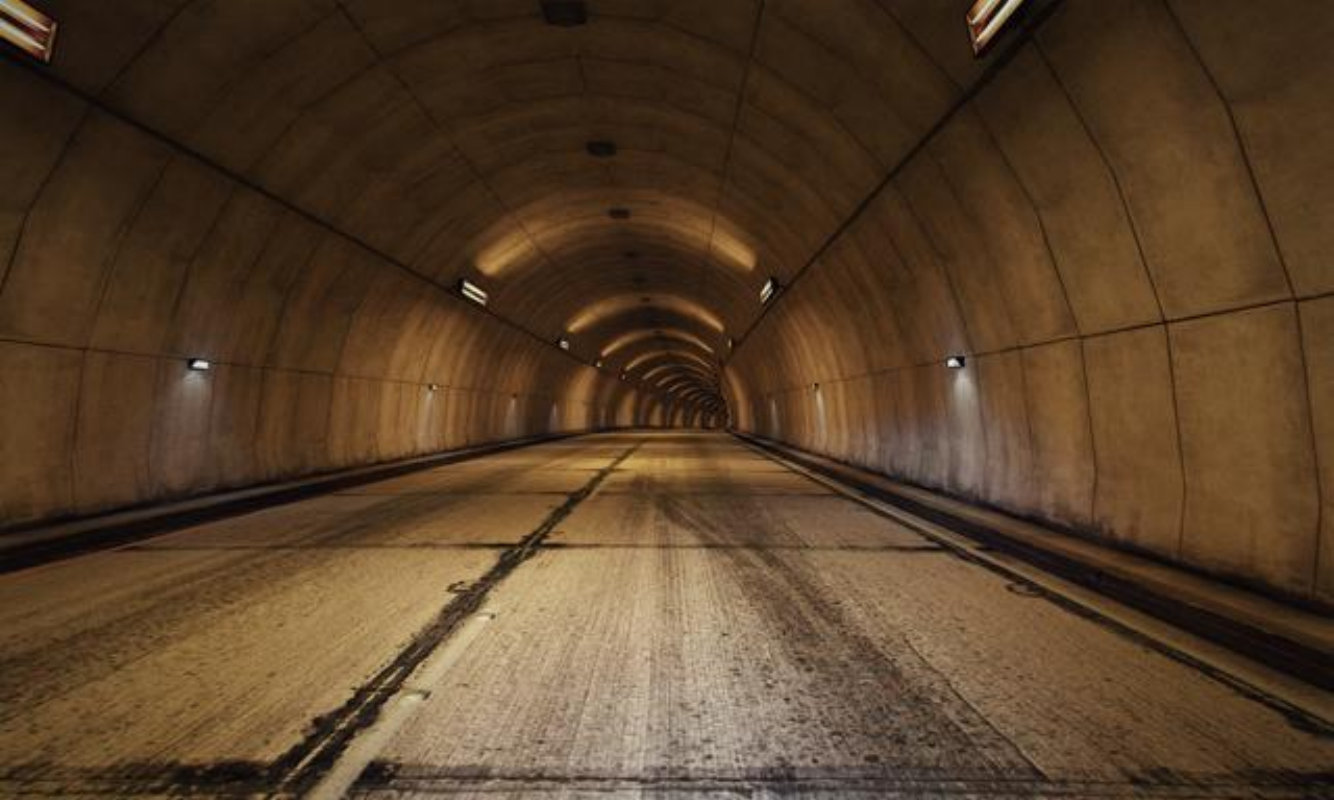}
\end{subfigure} &
\begin{subfigure}[b]{0.3\textwidth}
   \includegraphics[width=\textwidth]{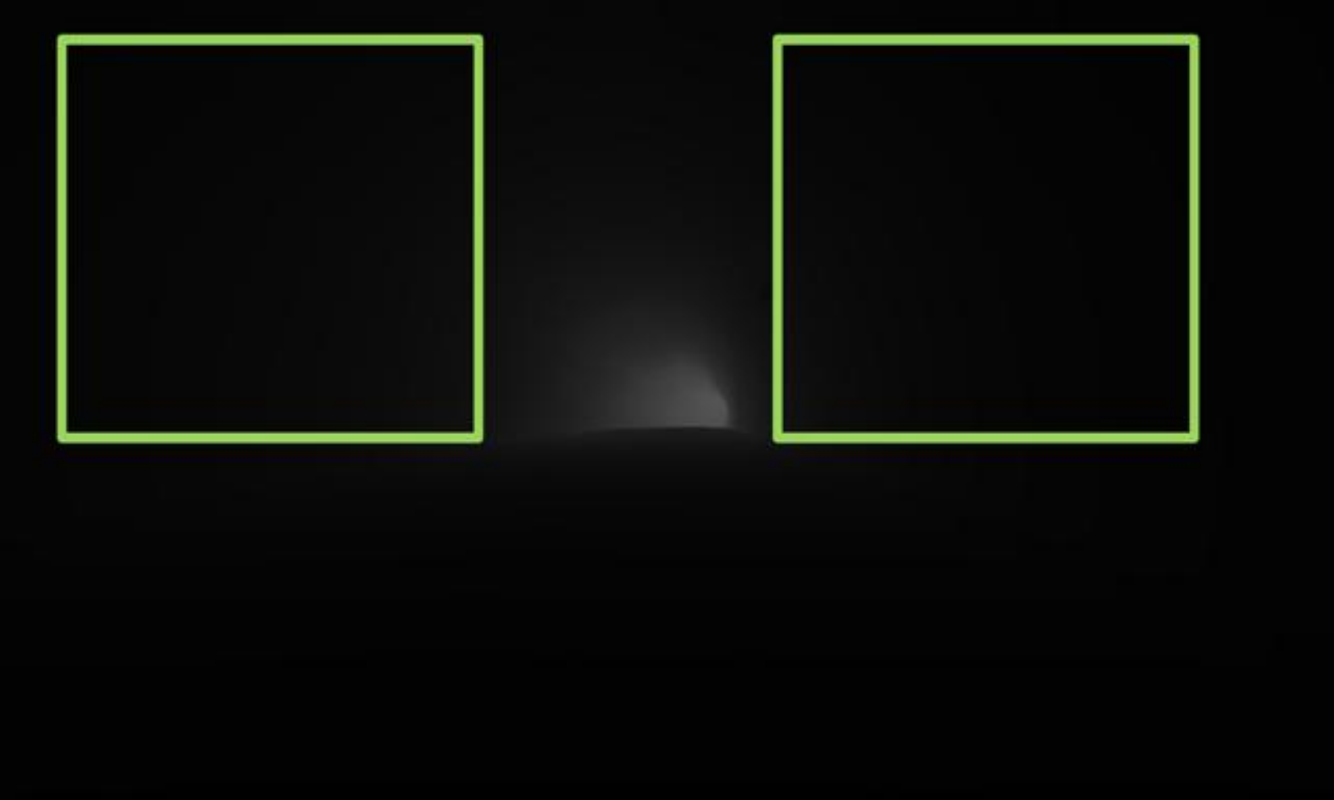}
\end{subfigure} &
\begin{subfigure}[b]{0.3\textwidth}
   \includegraphics[width=\textwidth]{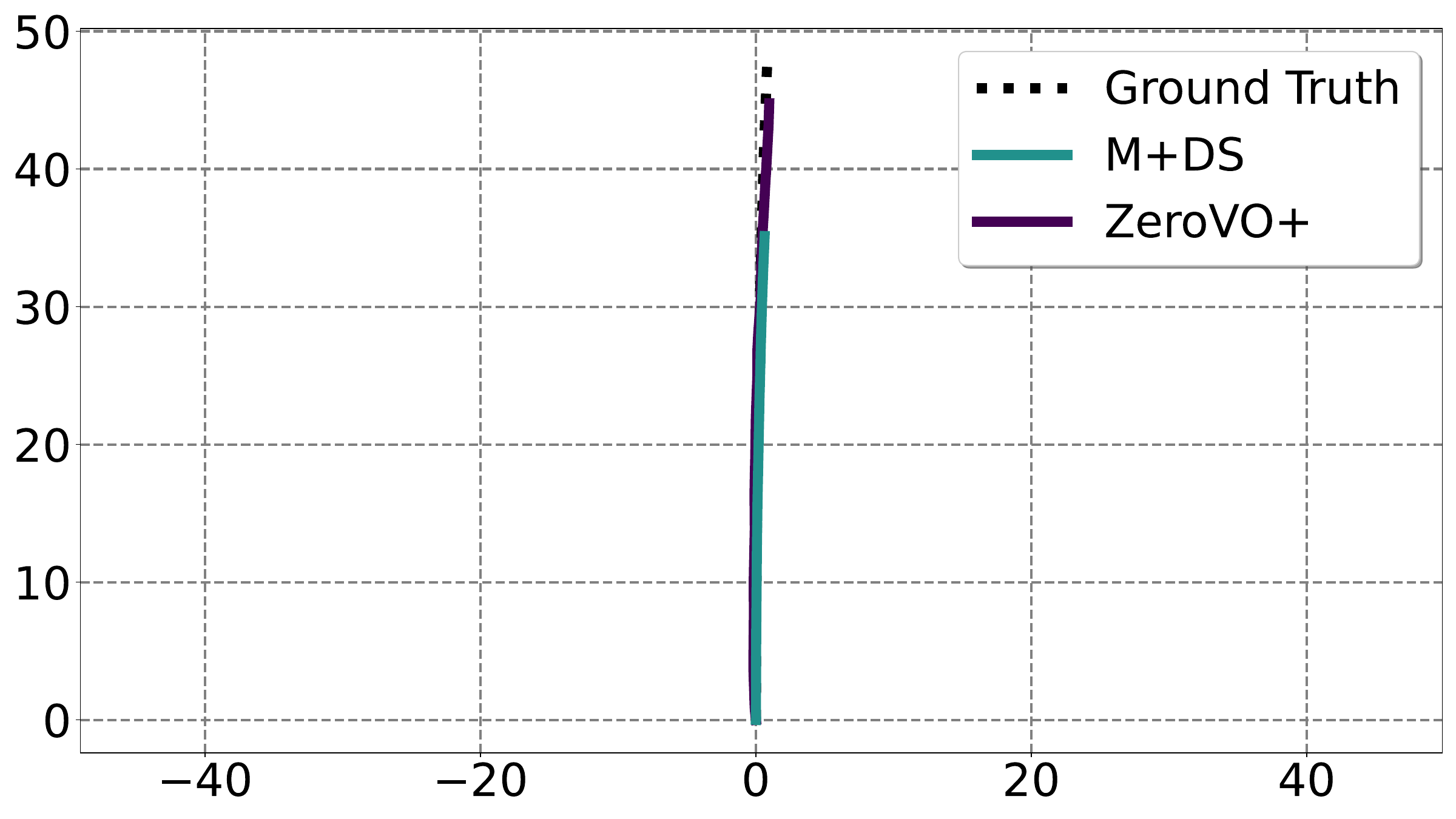}
\end{subfigure} \\

\hline

\hline
\begin{subfigure}[b]{0.3\textwidth}
   \includegraphics[width=\textwidth]{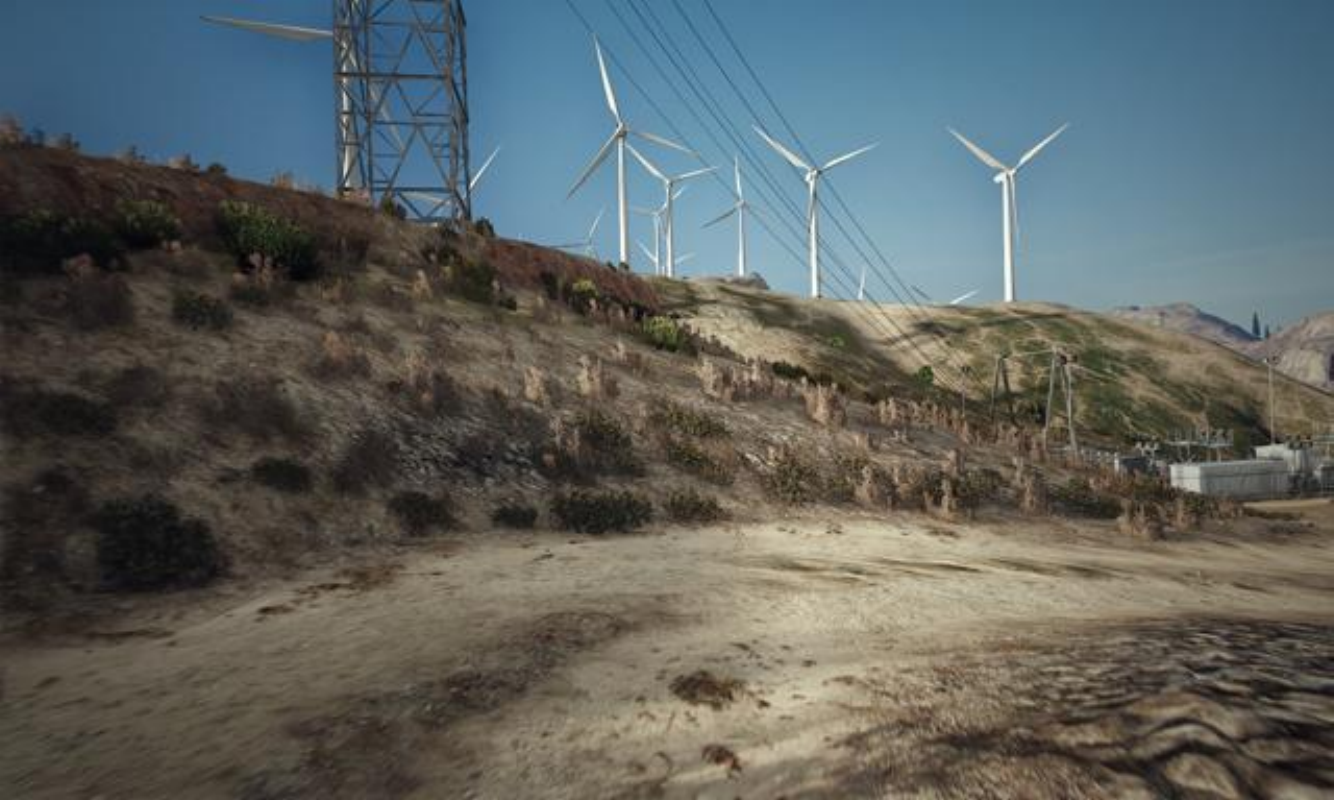}
\end{subfigure} &
\begin{subfigure}[b]{0.3\textwidth}
   \includegraphics[width=\textwidth]{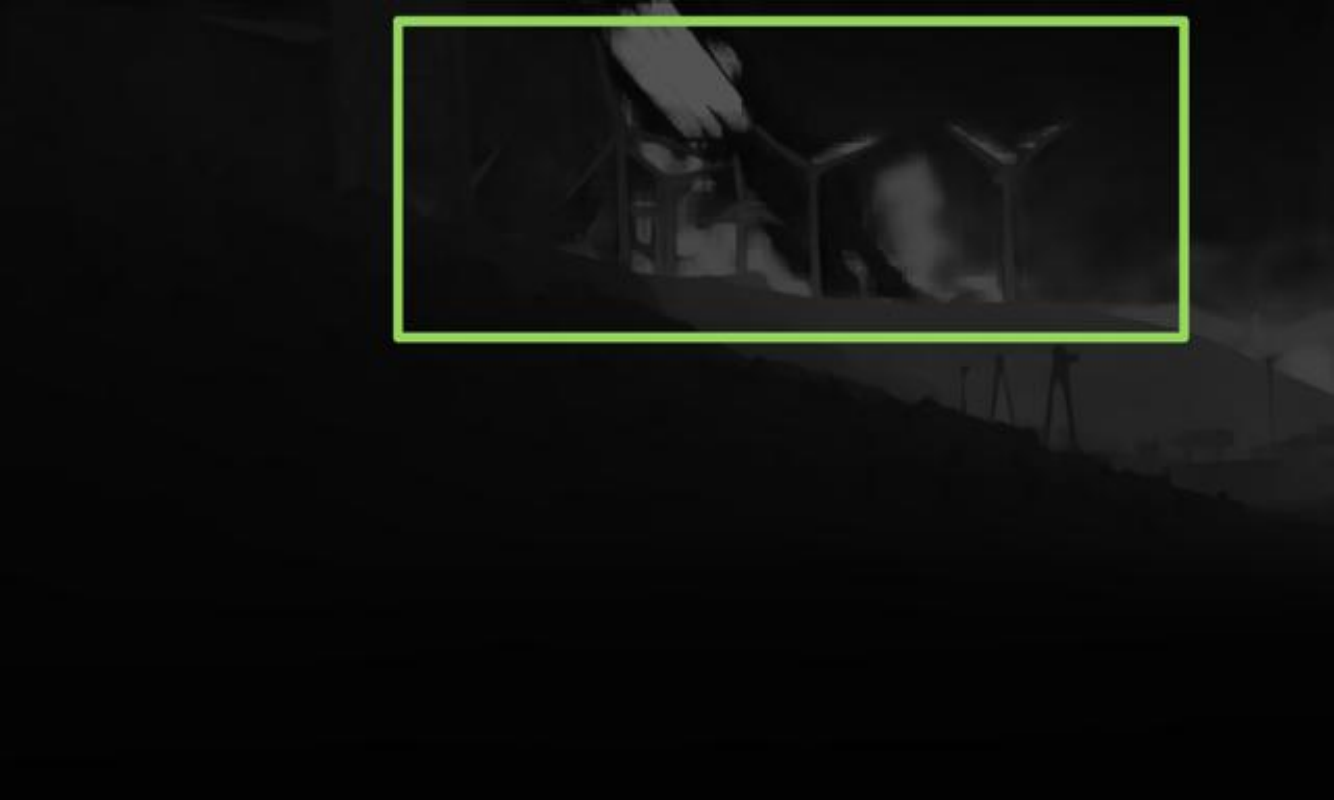}
\end{subfigure} &
\begin{subfigure}[b]{0.3\textwidth}
   \includegraphics[width=\textwidth]{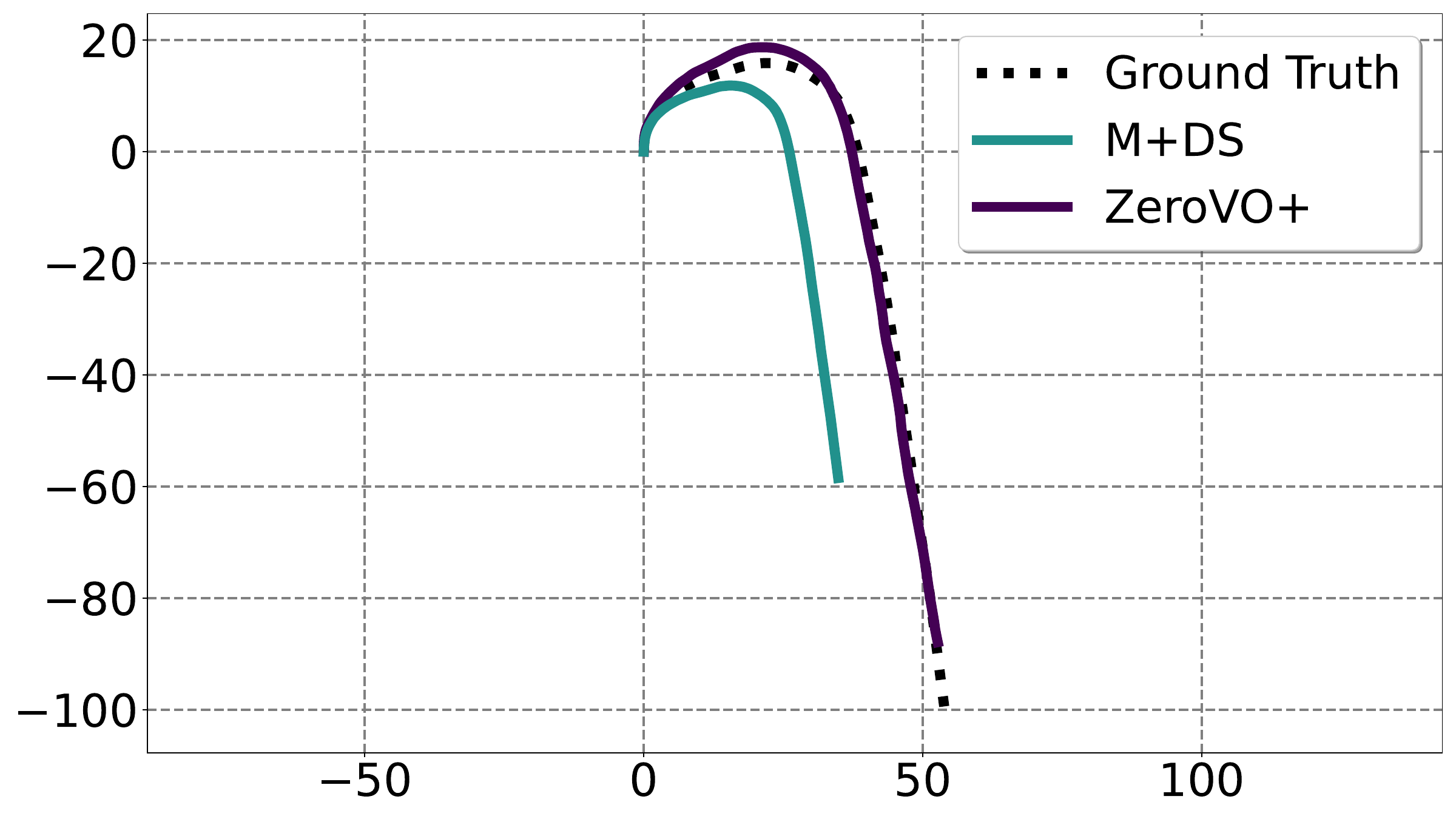}
\end{subfigure} \\
\hline

\hline
\begin{subfigure}[b]{0.3\textwidth}
   \includegraphics[width=\textwidth]{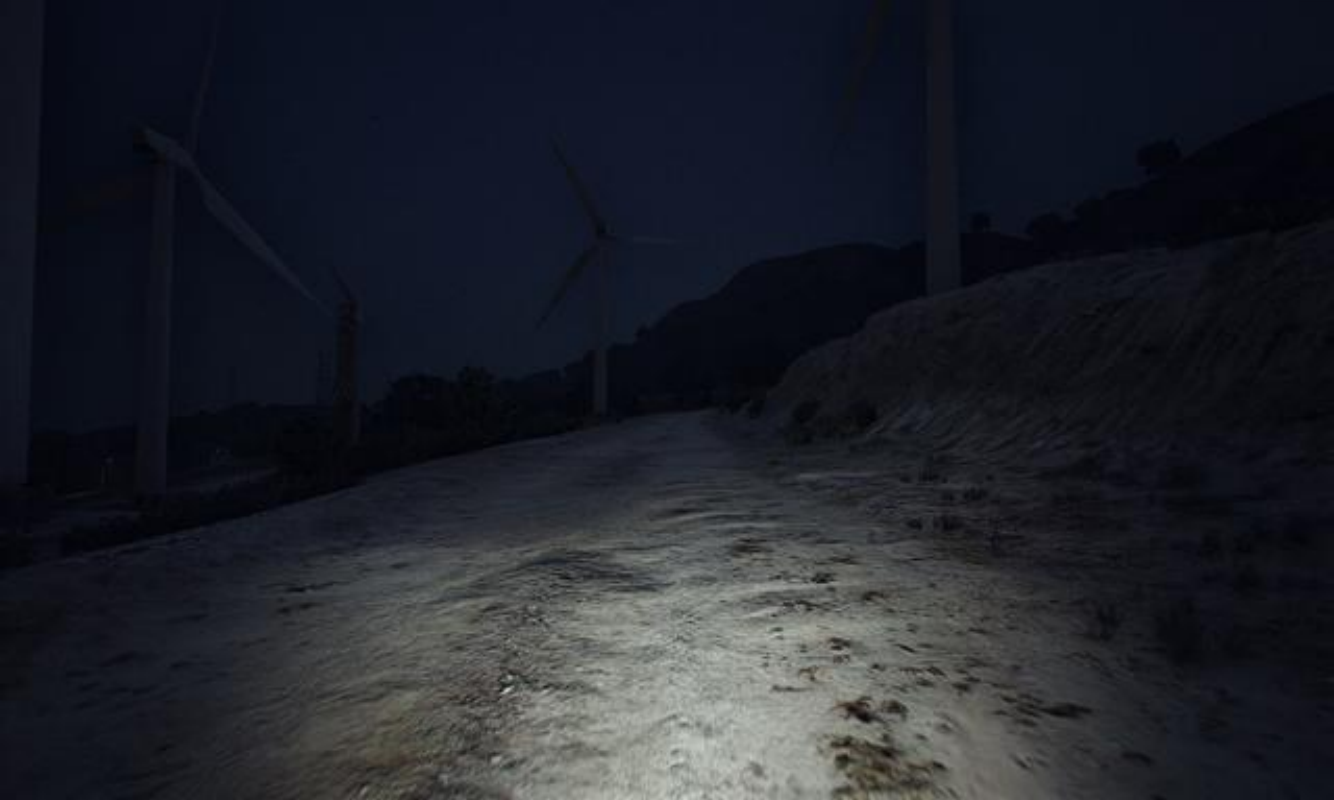}
\end{subfigure} &
\begin{subfigure}[b]{0.3\textwidth}
   \includegraphics[width=\textwidth]{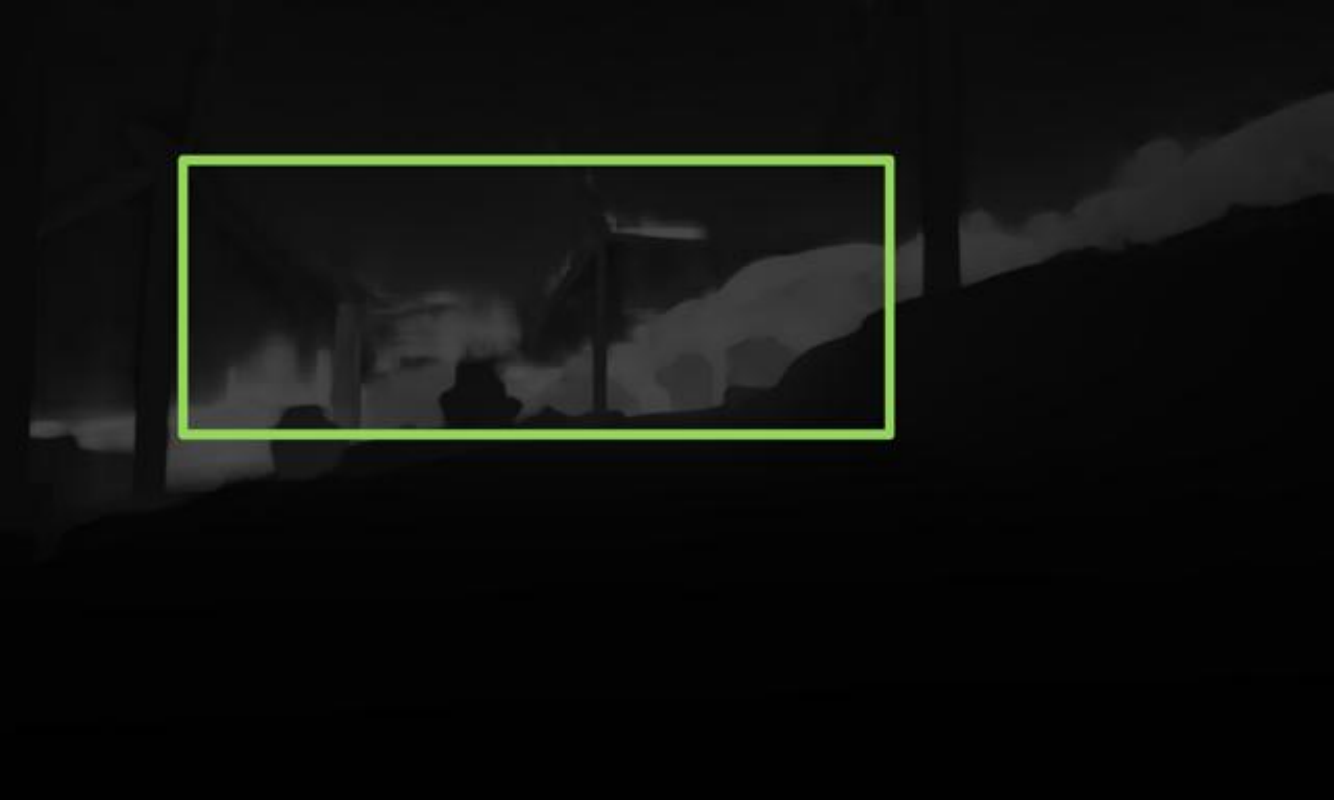}
\end{subfigure} &
\begin{subfigure}[b]{0.3\textwidth}
   \includegraphics[width=\textwidth]{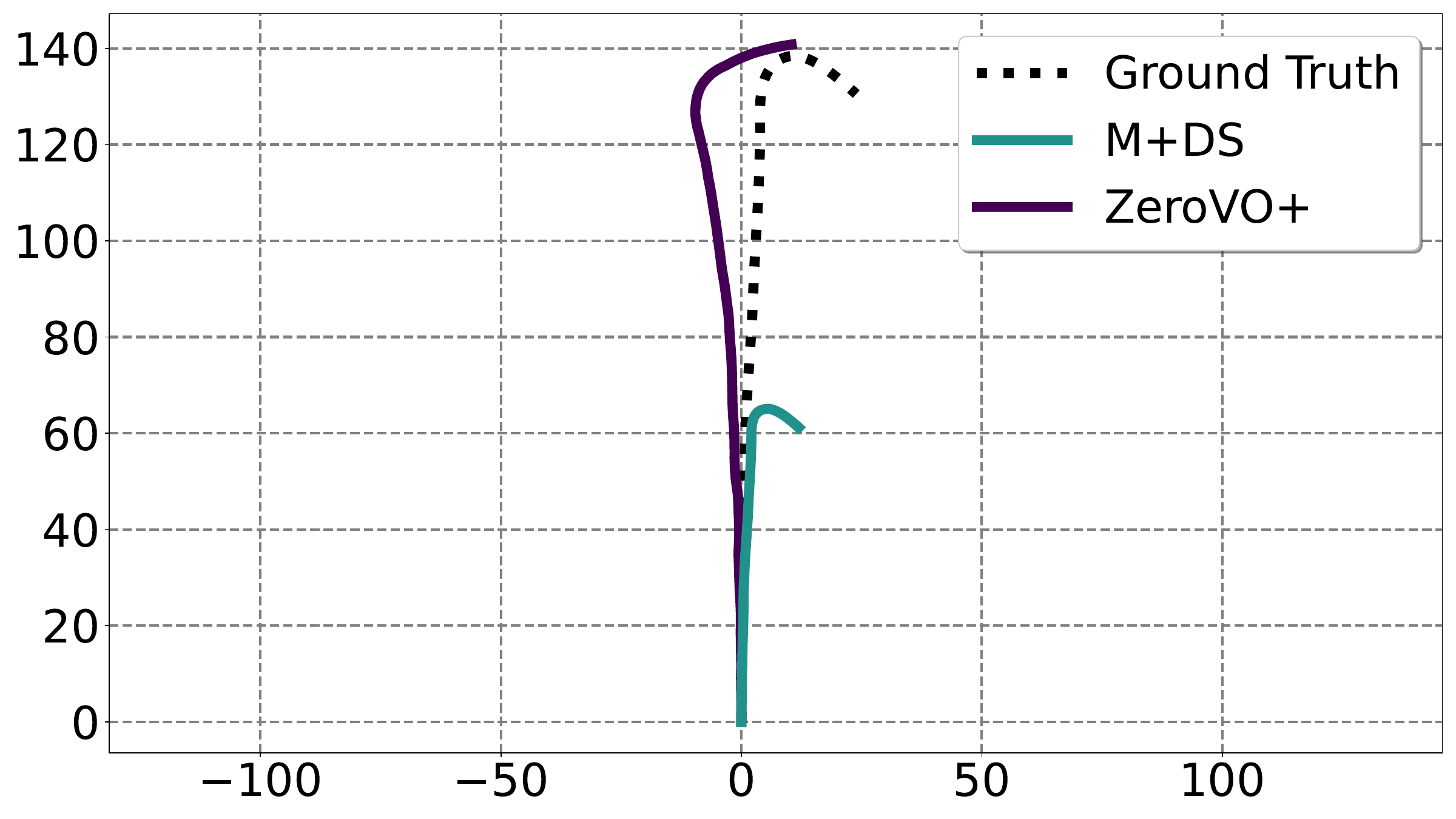}
\end{subfigure} \\
\hline

\hline
\begin{subfigure}[b]{0.3\textwidth}
   \includegraphics[width=\textwidth]{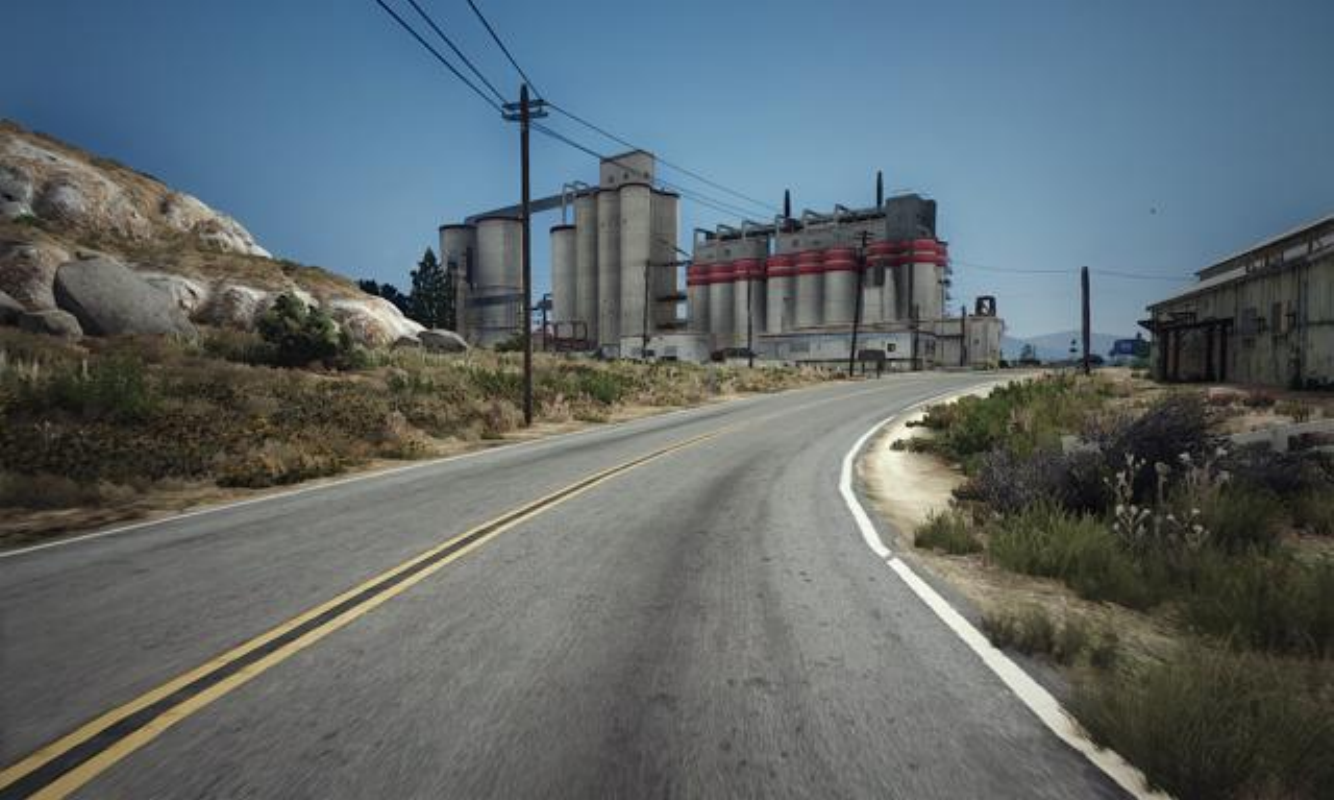}
\end{subfigure} &
\begin{subfigure}[b]{0.3\textwidth}
   \includegraphics[width=\textwidth]{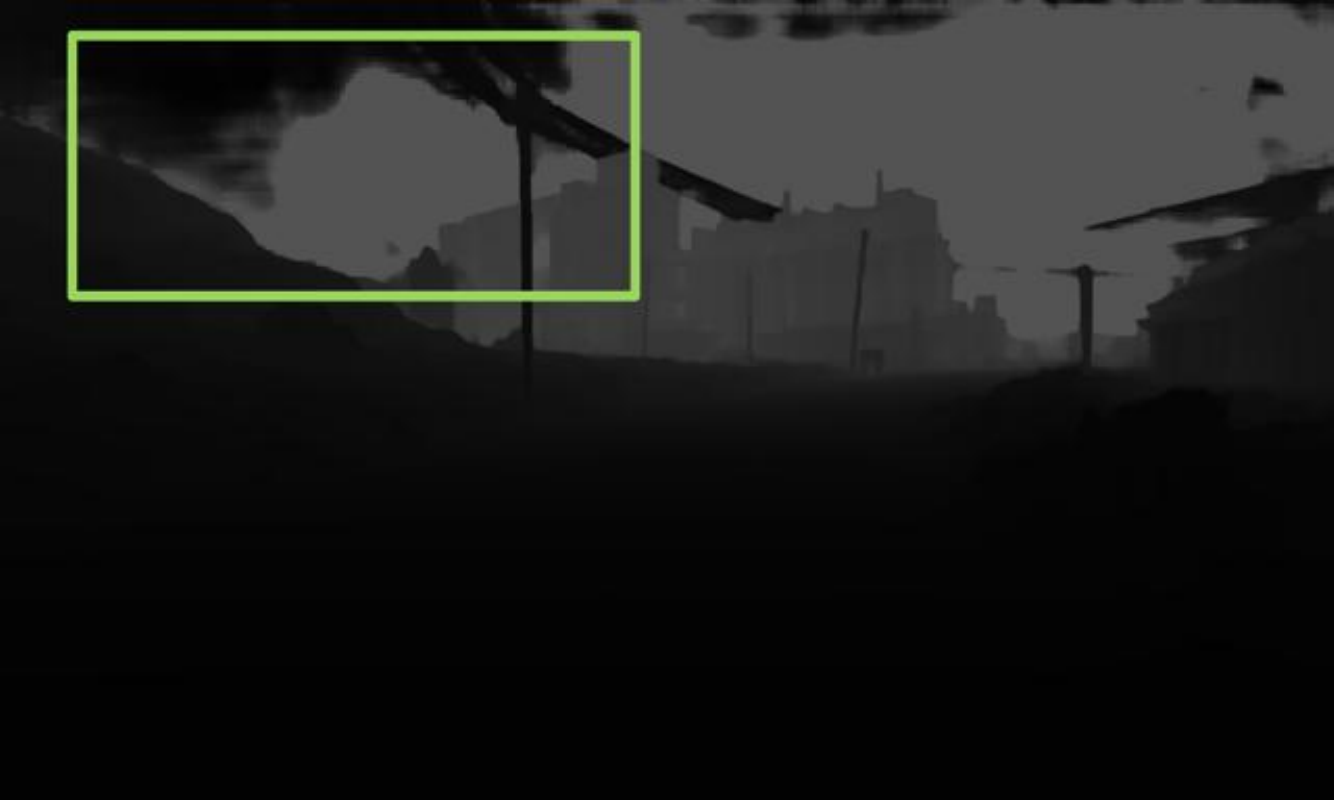}
\end{subfigure} &
\begin{subfigure}[b]{0.3\textwidth}
   \includegraphics[width=\textwidth]{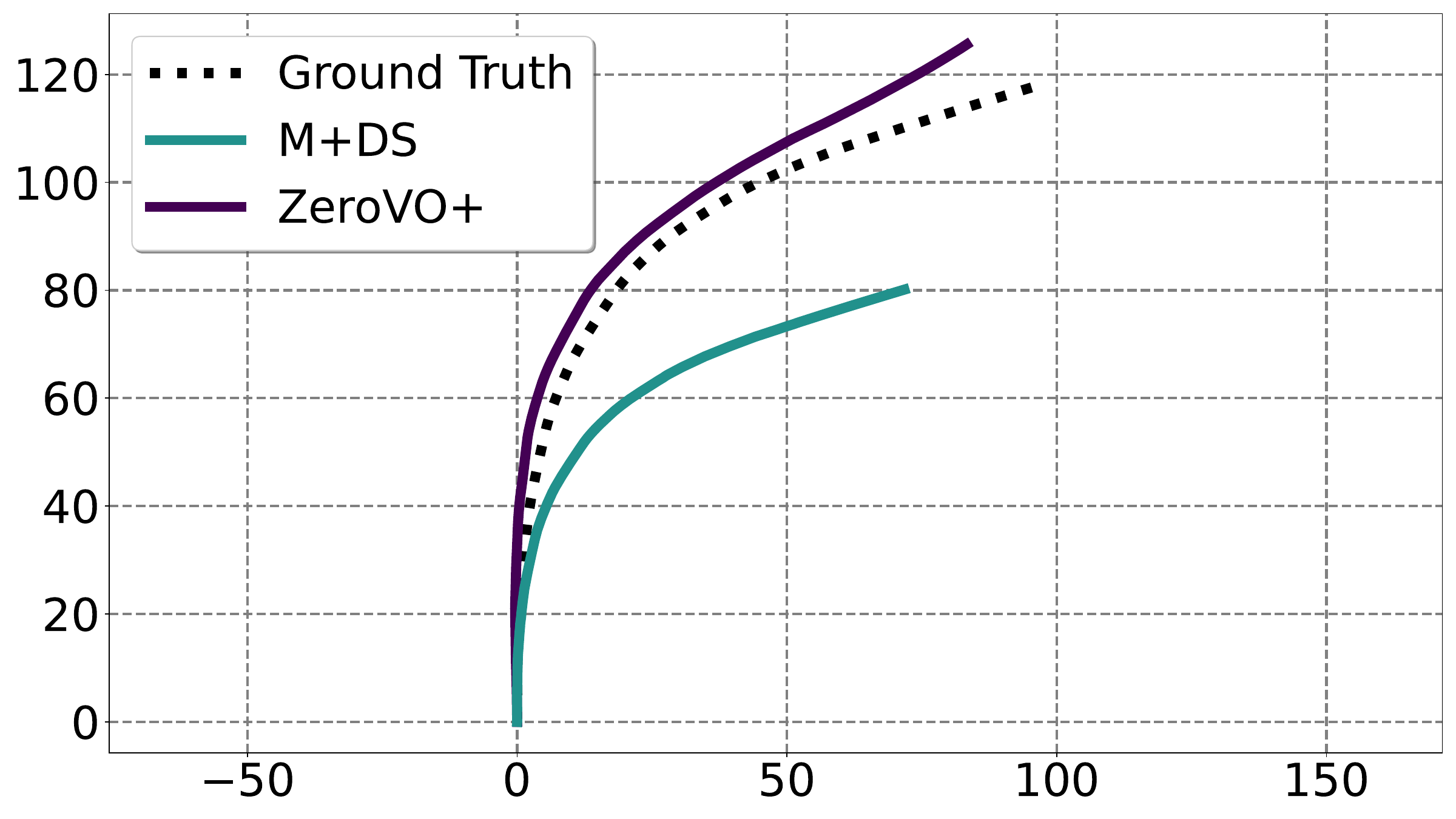}
\end{subfigure} \\
\hline

\hline
\begin{subfigure}[b]{0.3\textwidth}
   \includegraphics[width=\textwidth]{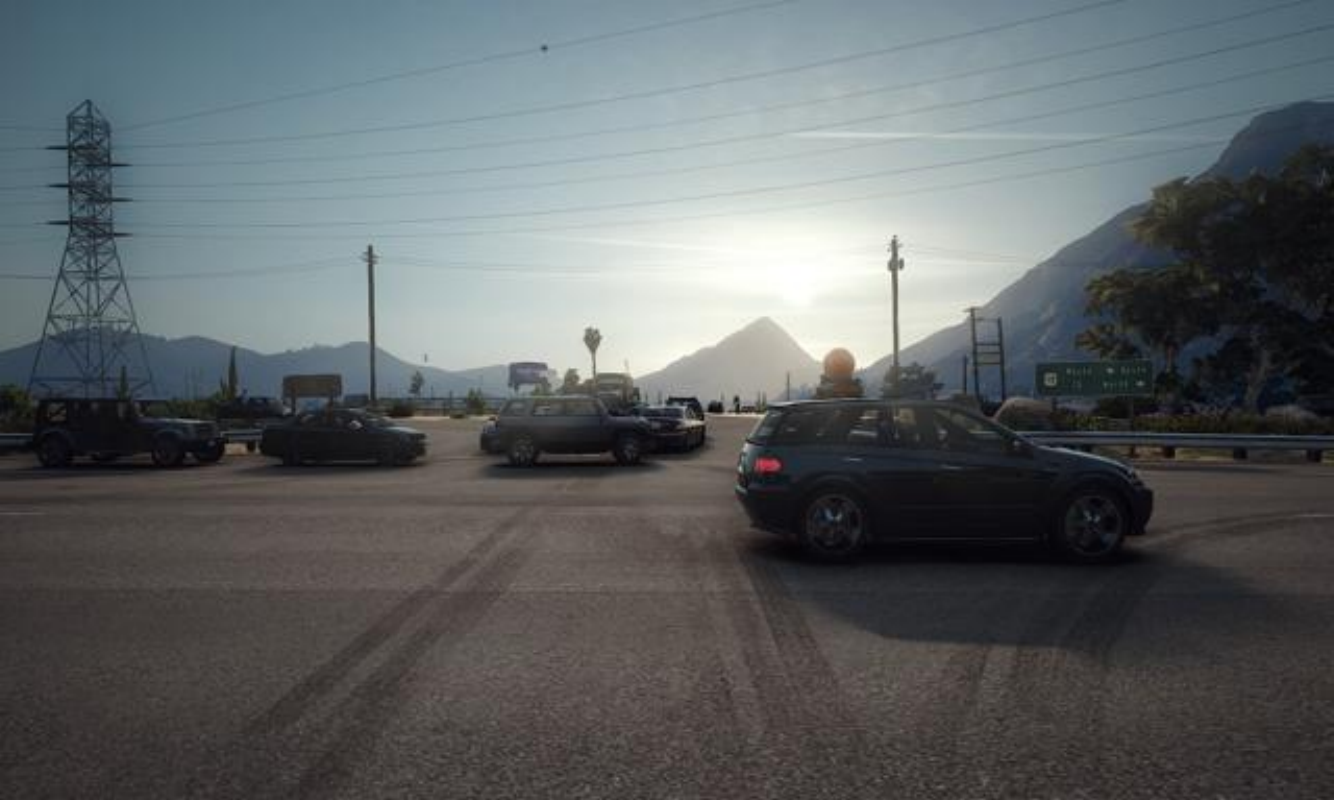}
\end{subfigure} &
\begin{subfigure}[b]{0.3\textwidth}
   \includegraphics[width=\textwidth]{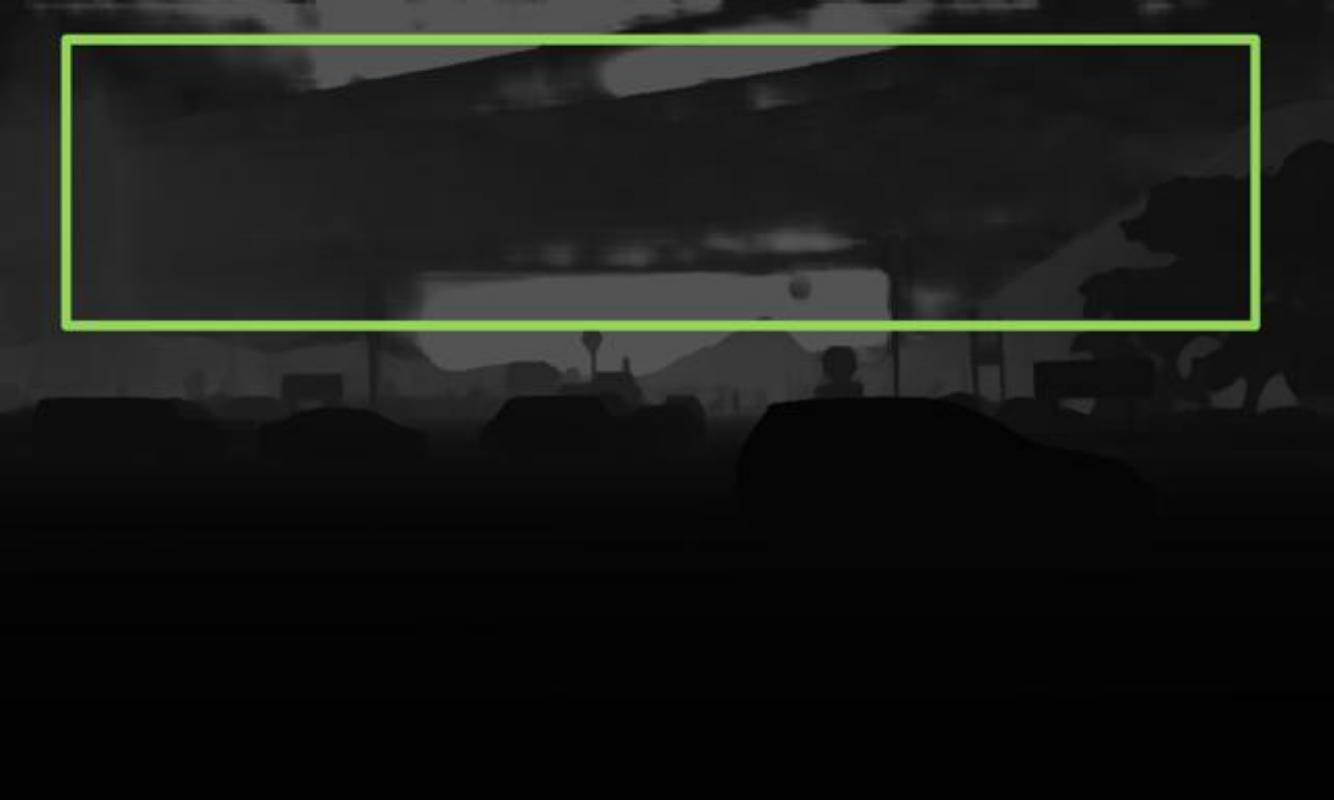}
\end{subfigure} &
\begin{subfigure}[b]{0.3\textwidth}
   \includegraphics[width=\textwidth]{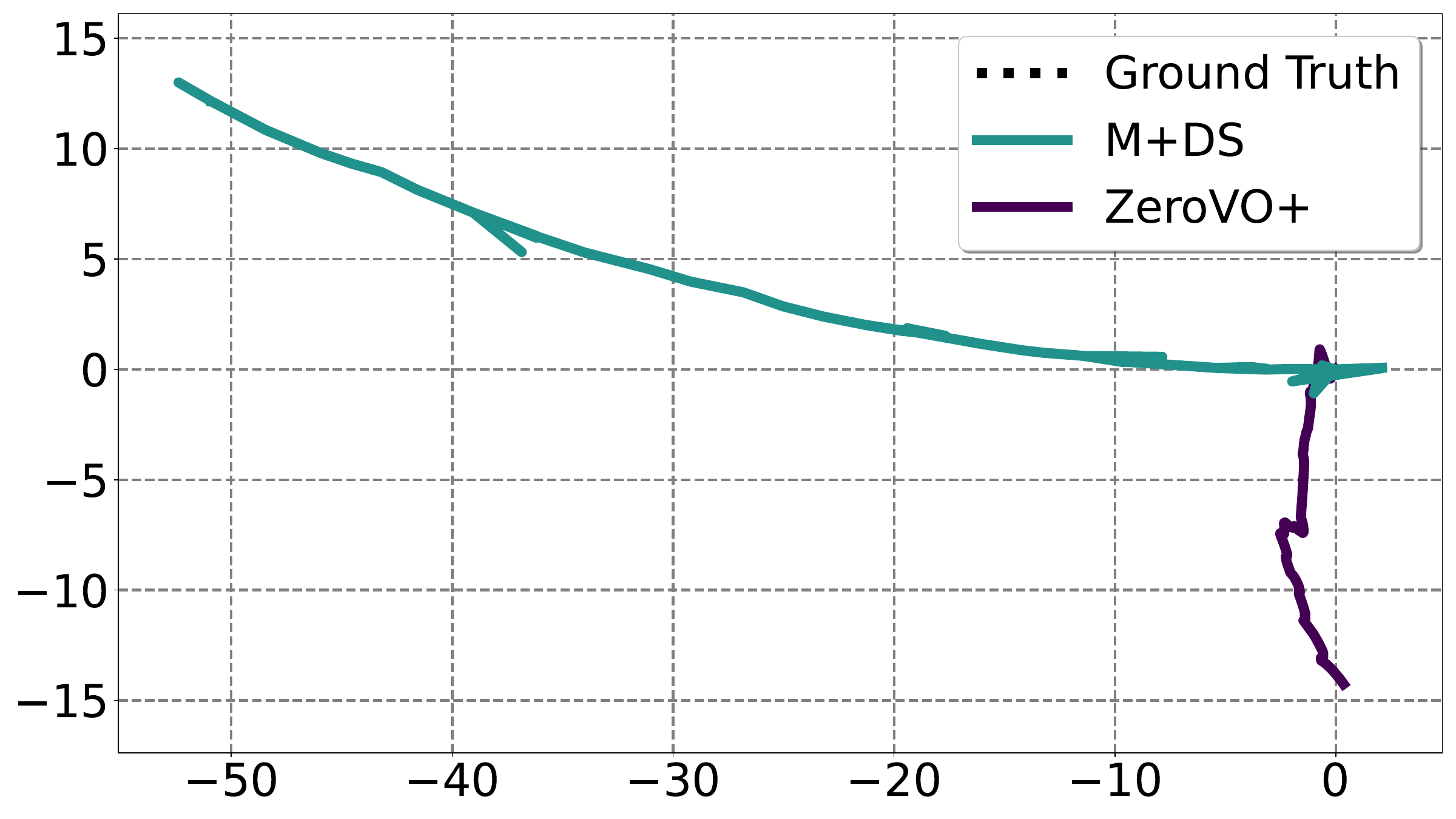}
\end{subfigure} \\
\hline

\end{tabular}

\end{table*}

\begin{table}[!t]
\centering
\caption{\textbf{Qualitative Failure Cases on Argoverse.} This figure illustrates qualitative results of failure cases on Argoverse with Metric3Dv2 combined with Droid-SLAM. The left column displays the RGB input images, the middle column shows the corresponding depth predictions (darker pixels indicate closer distances), and the right column compares trajectory estimations. The green boxes in the depth images highlight regions where Metric3Dv2 struggles due to environmental factors such as sky reflections, clouds, lens artifacts, and glass surfaces. These challenges lead to significant trajectory drifts. }
\label{tab:argo_fail1}
\begin{tabular}{|c|c|c|}
\hline
\rowcolor{lightgray}
\textbf{RGB Image} & \textbf{Wrong Depth Image} & \textbf{Plotted Trajectory} \\

\hline
\begin{subfigure}[b]{0.3\textwidth}
   \includegraphics[width=\textwidth]{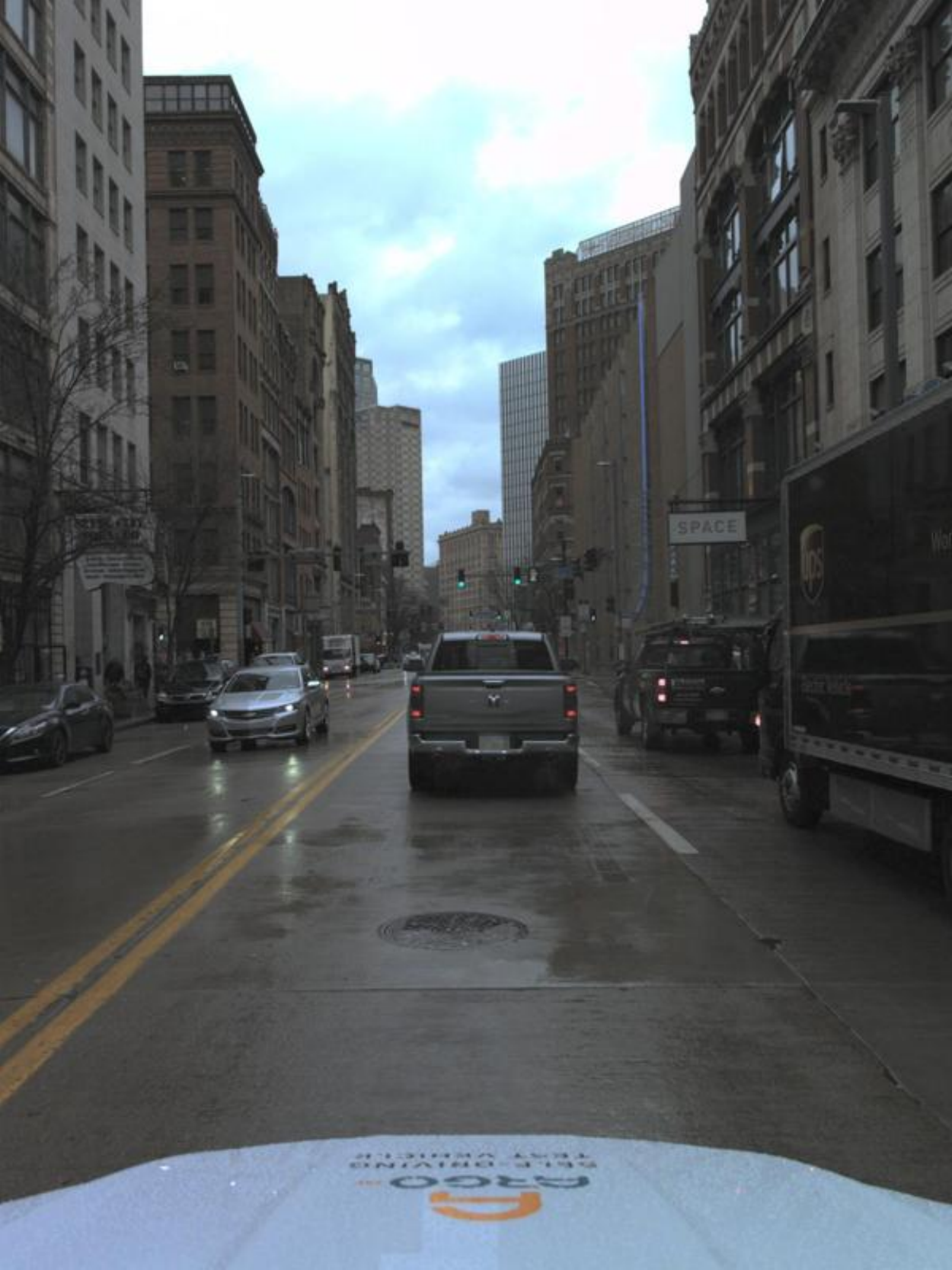}
\end{subfigure} &
\begin{subfigure}[b]{0.3\textwidth}
   \includegraphics[width=\textwidth]{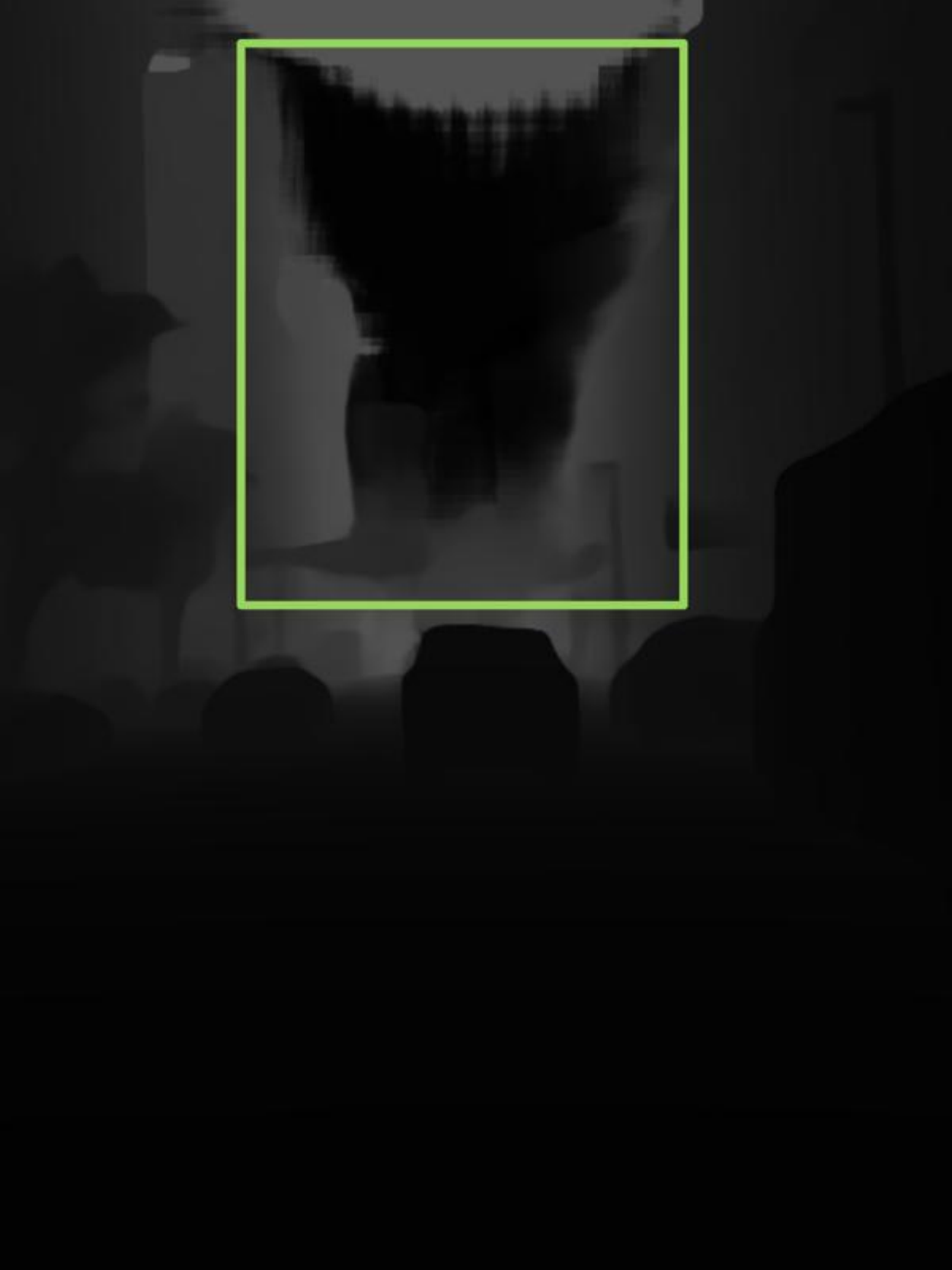}
\end{subfigure} &
\begin{subfigure}[b]{0.3\textwidth}
   \includegraphics[width=\textwidth]{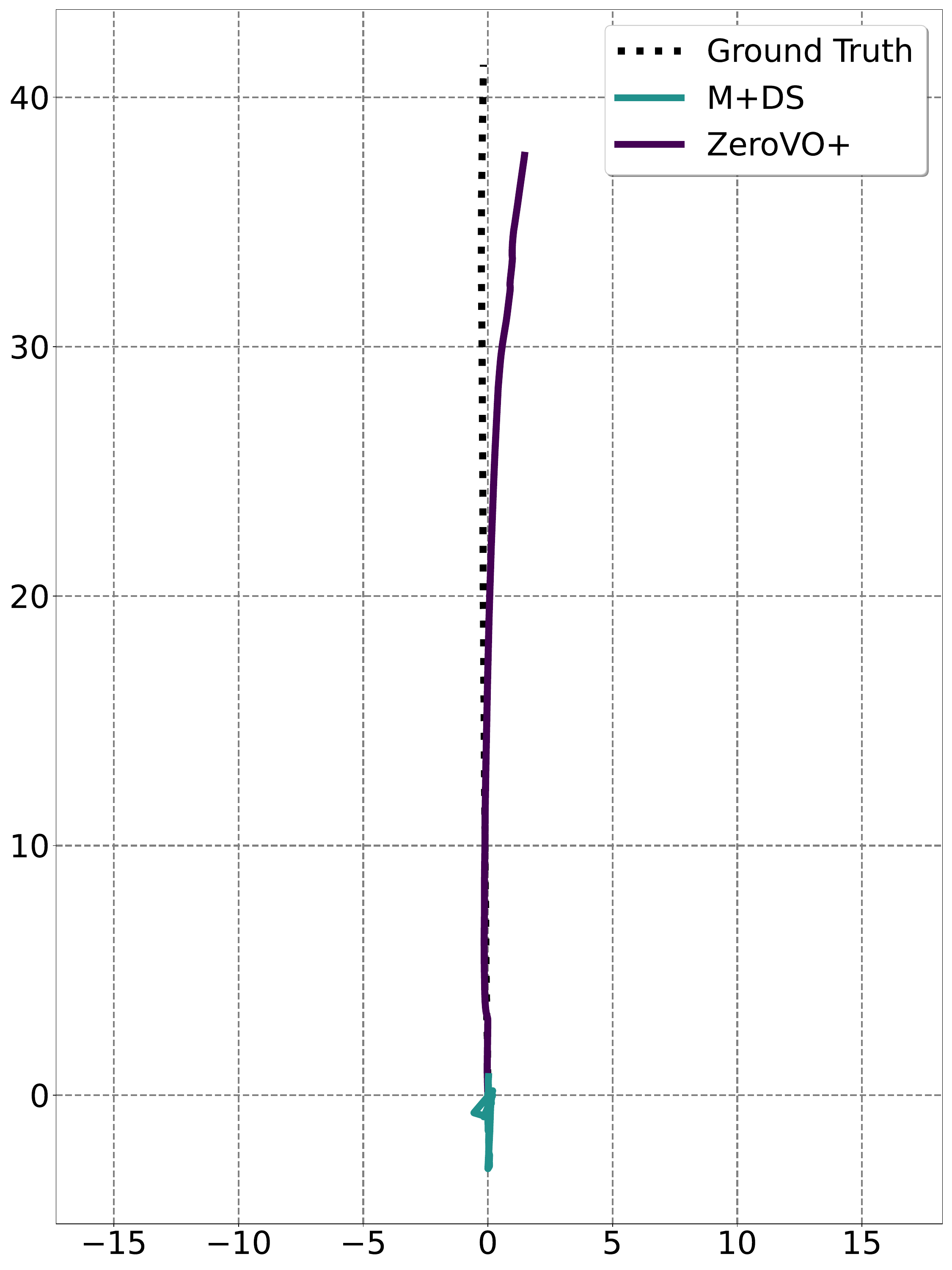}
\end{subfigure} \\
\hline

\hline
\begin{subfigure}[b]{0.3\textwidth}
   \includegraphics[width=\textwidth]{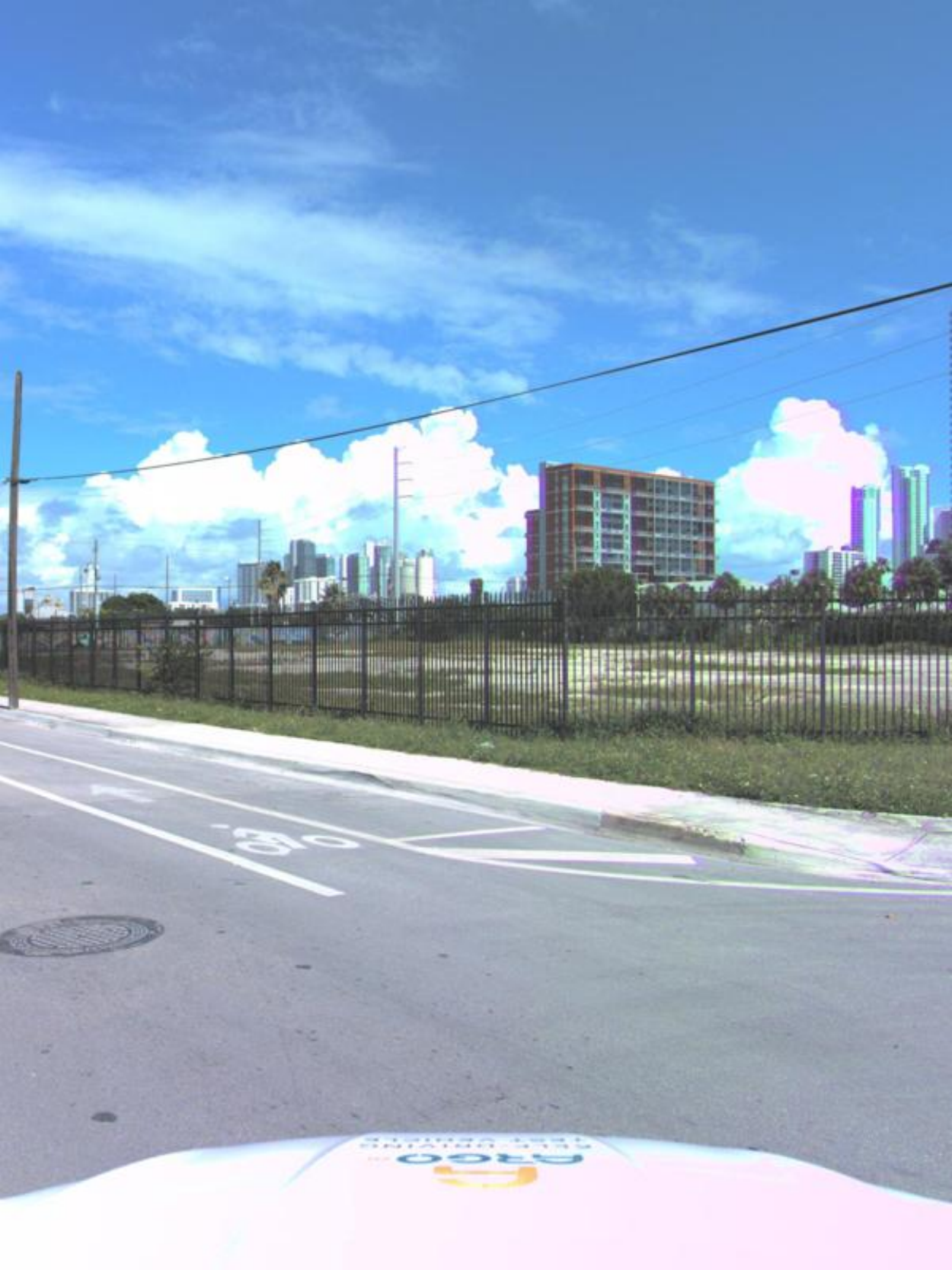}
\end{subfigure} &
\begin{subfigure}[b]{0.3\textwidth}
   \includegraphics[width=\textwidth]{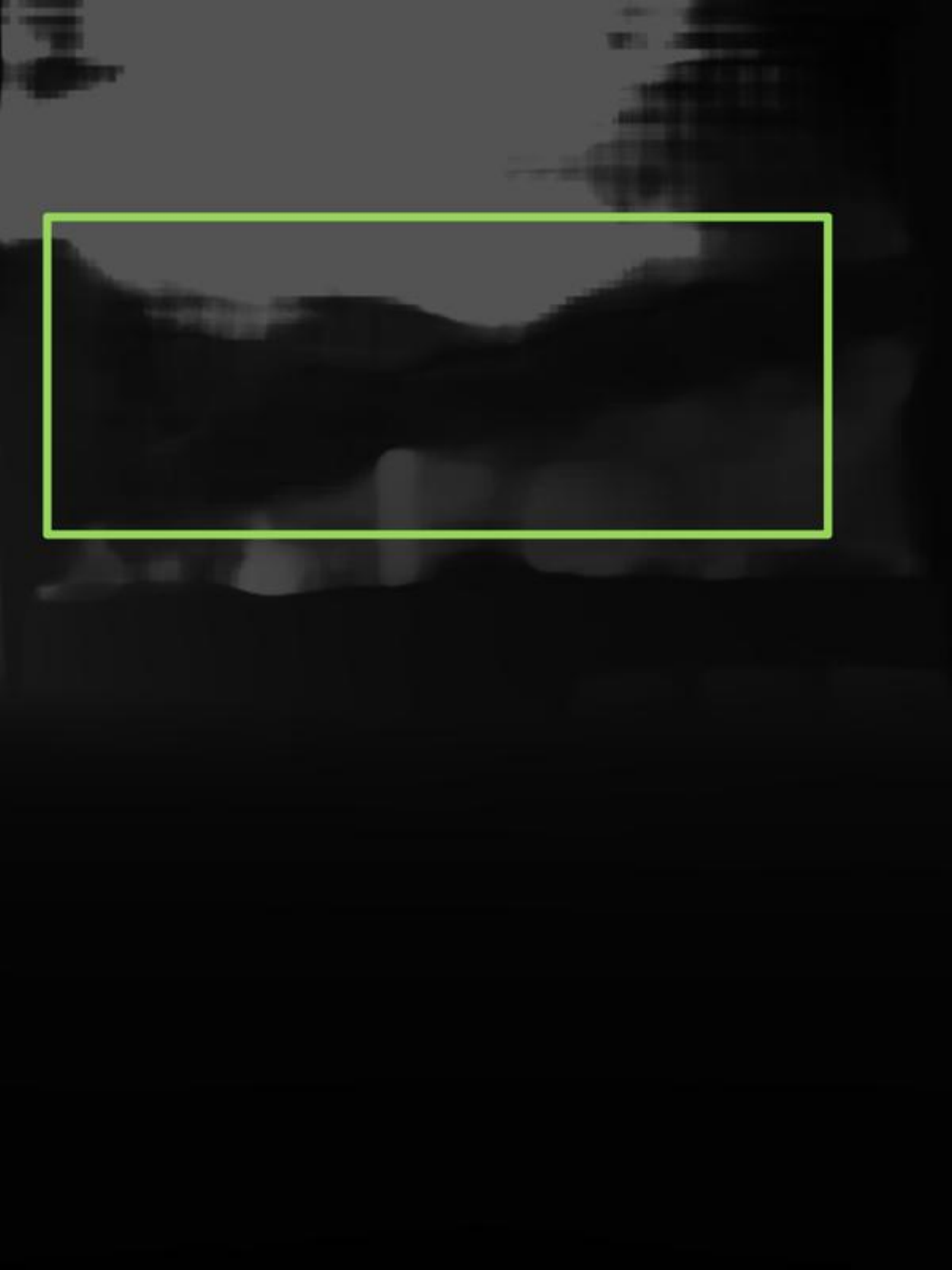}
\end{subfigure} &
\begin{subfigure}[b]{0.3\textwidth}
   \includegraphics[width=\textwidth]{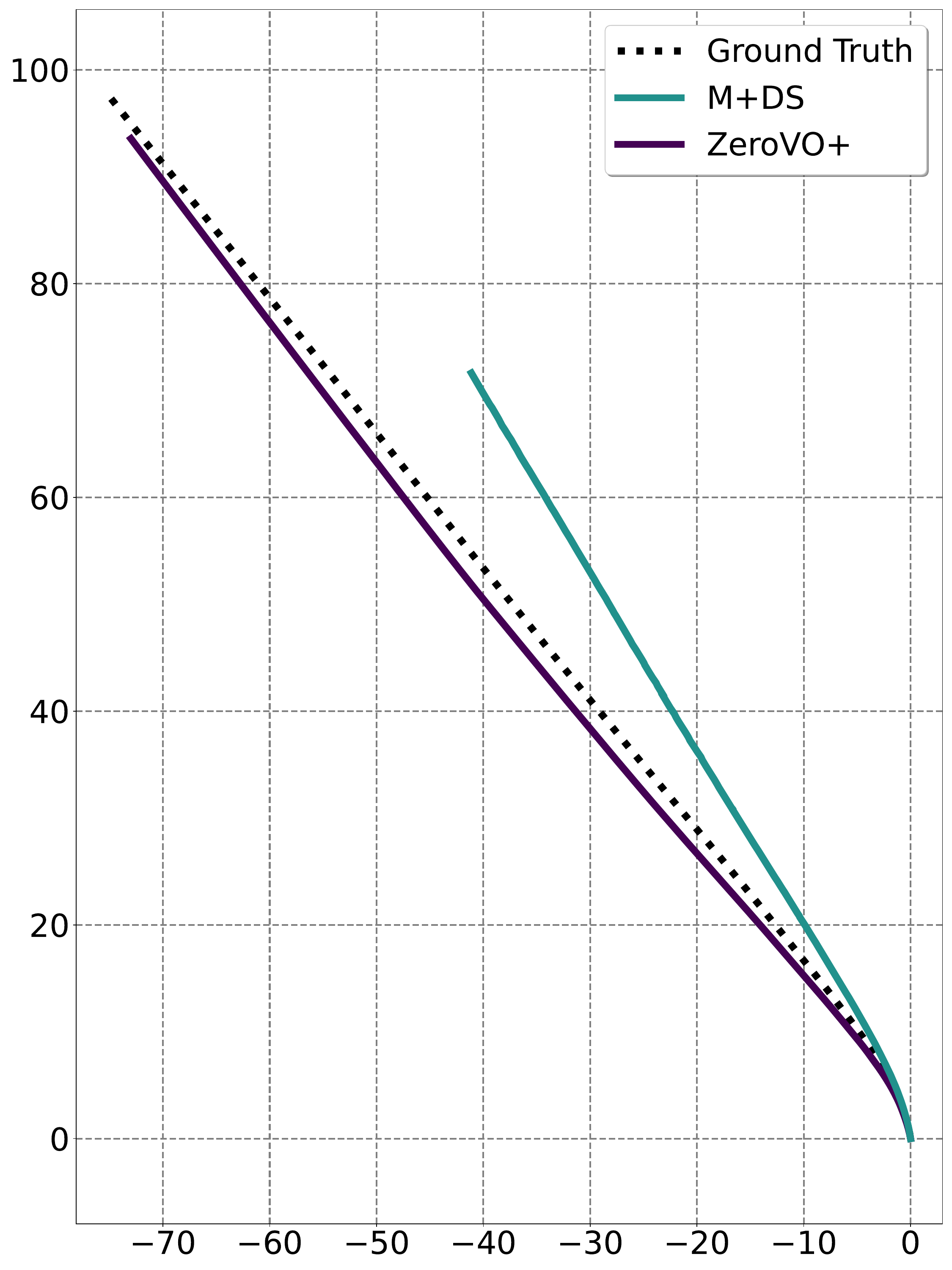}
\end{subfigure} \\

\hline

\hline
\begin{subfigure}[b]{0.3\textwidth}
   \includegraphics[width=\textwidth]{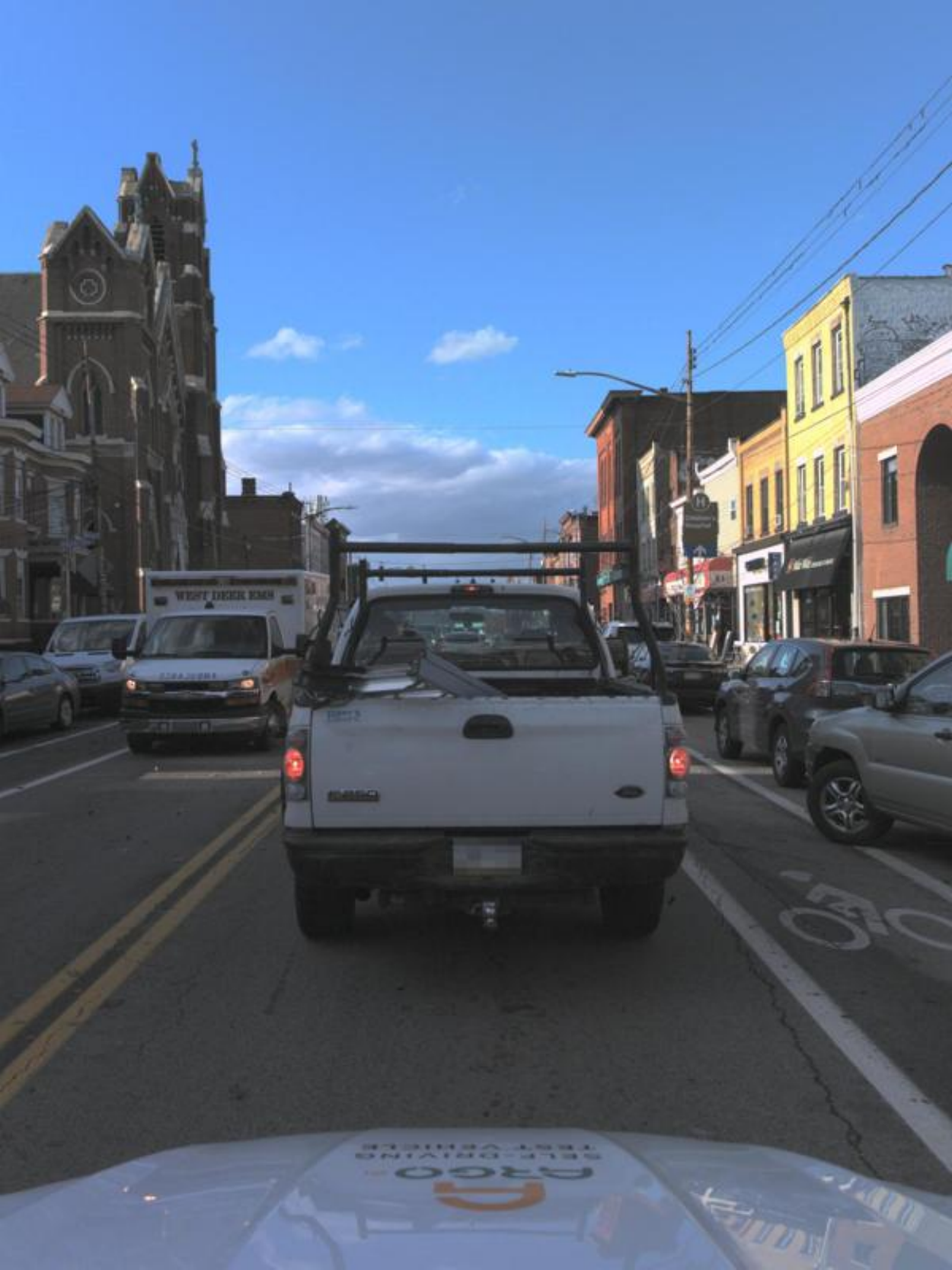}
\end{subfigure} &
\begin{subfigure}[b]{0.3\textwidth}
   \includegraphics[width=\textwidth]{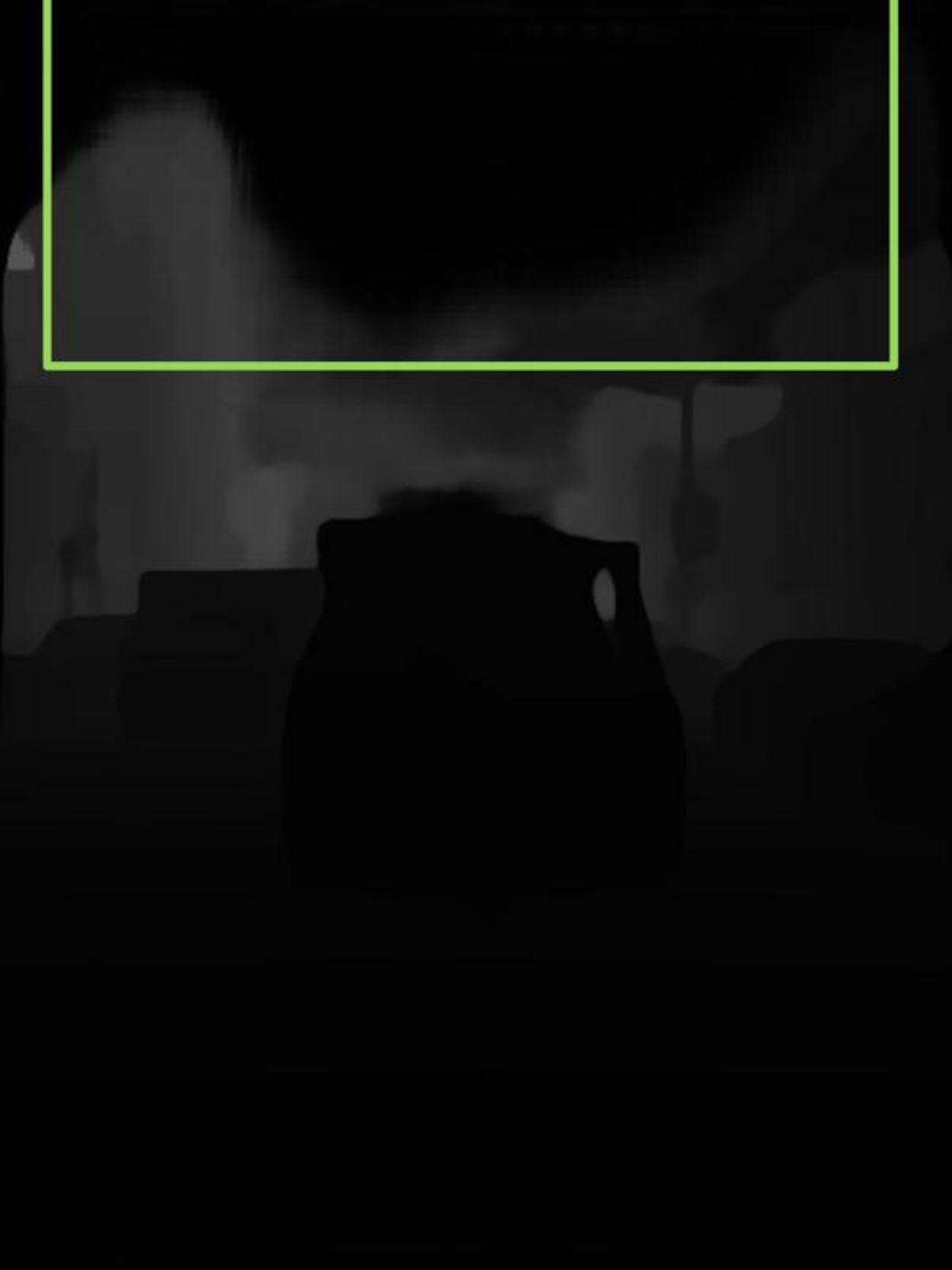}
\end{subfigure} &
\begin{subfigure}[b]{0.3\textwidth}
   \includegraphics[width=\textwidth]{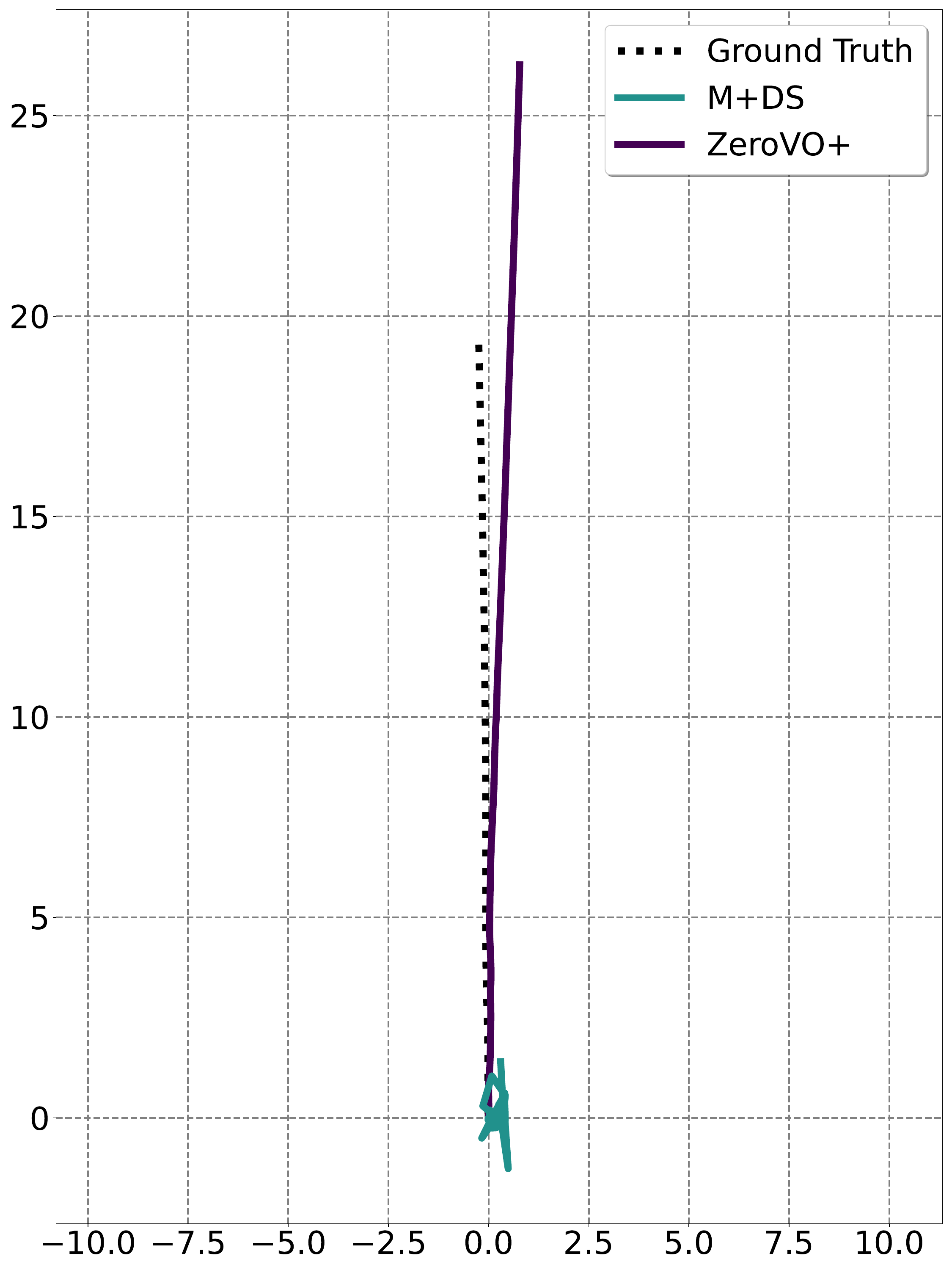}
\end{subfigure} \\
\hline
\end{tabular}

\end{table}

\begin{table}[!t]
\centering
\caption{\textbf{Qualitative Failure Cases on Argoverse.} This figure illustrates qualitative results of failure cases on Argoverse with Metric3Dv2 combined with Droid-SLAM. The left column displays the RGB input images, the middle column shows the corresponding depth predictions (darker pixels indicate closer distances), and the right column compares trajectory estimations. The green boxes in the depth images highlight regions where Metric3Dv2 struggles due to environmental factors such as sky reflections, clouds, lens artifacts, and glass surfaces. These challenges lead to significant trajectory drifts. }
\label{tab:argo_fail2}
\begin{tabular}{|c|c|c|}
\hline
\rowcolor{lightgray}
\textbf{RGB Image} & \textbf{Wrong Depth Image} & \textbf{Plotted Trajectory} \\

\hline
\begin{subfigure}[b]{0.3\textwidth}
   \includegraphics[width=\textwidth]{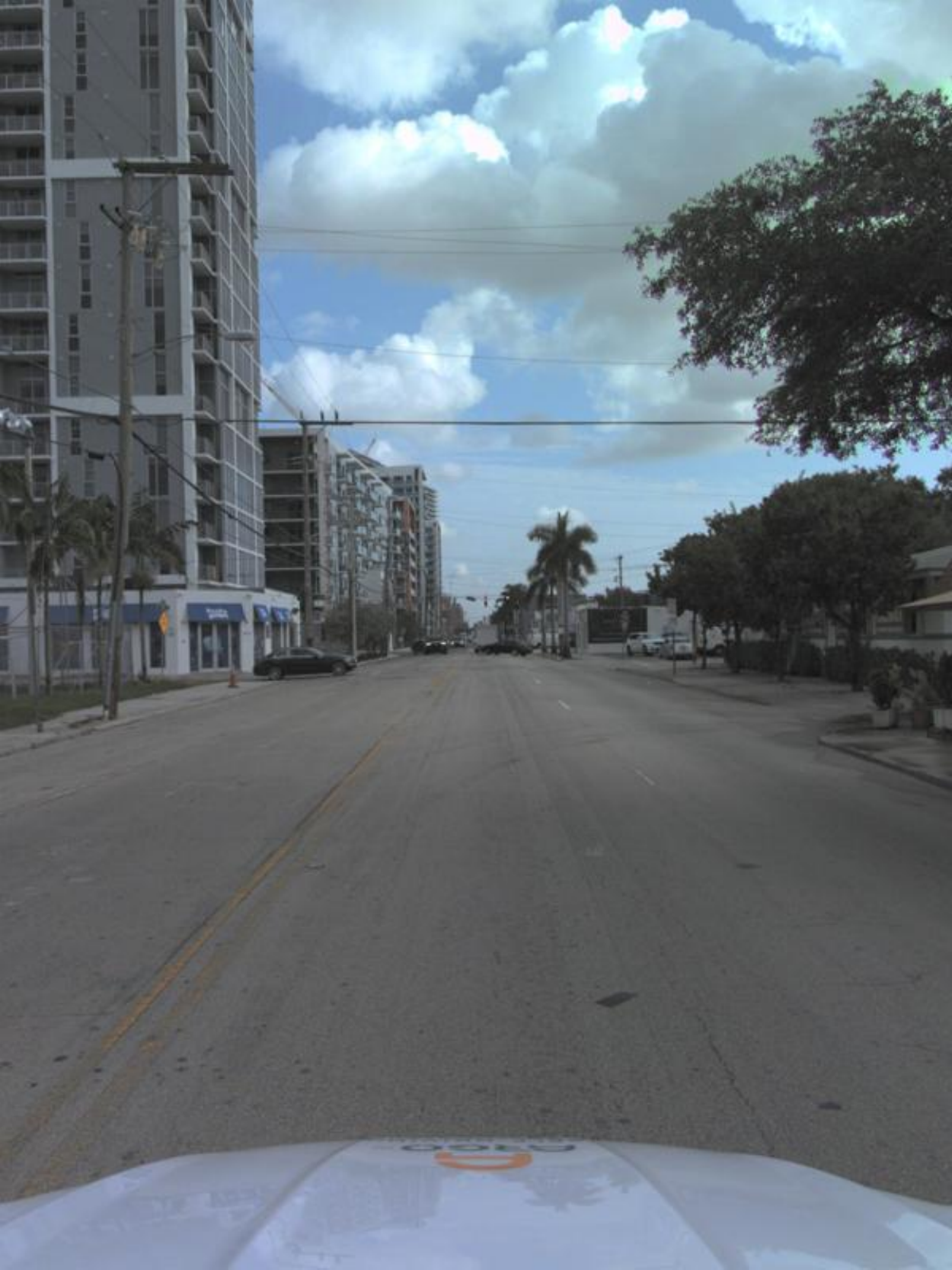}
\end{subfigure} &
\begin{subfigure}[b]{0.3\textwidth}
   \includegraphics[width=\textwidth]{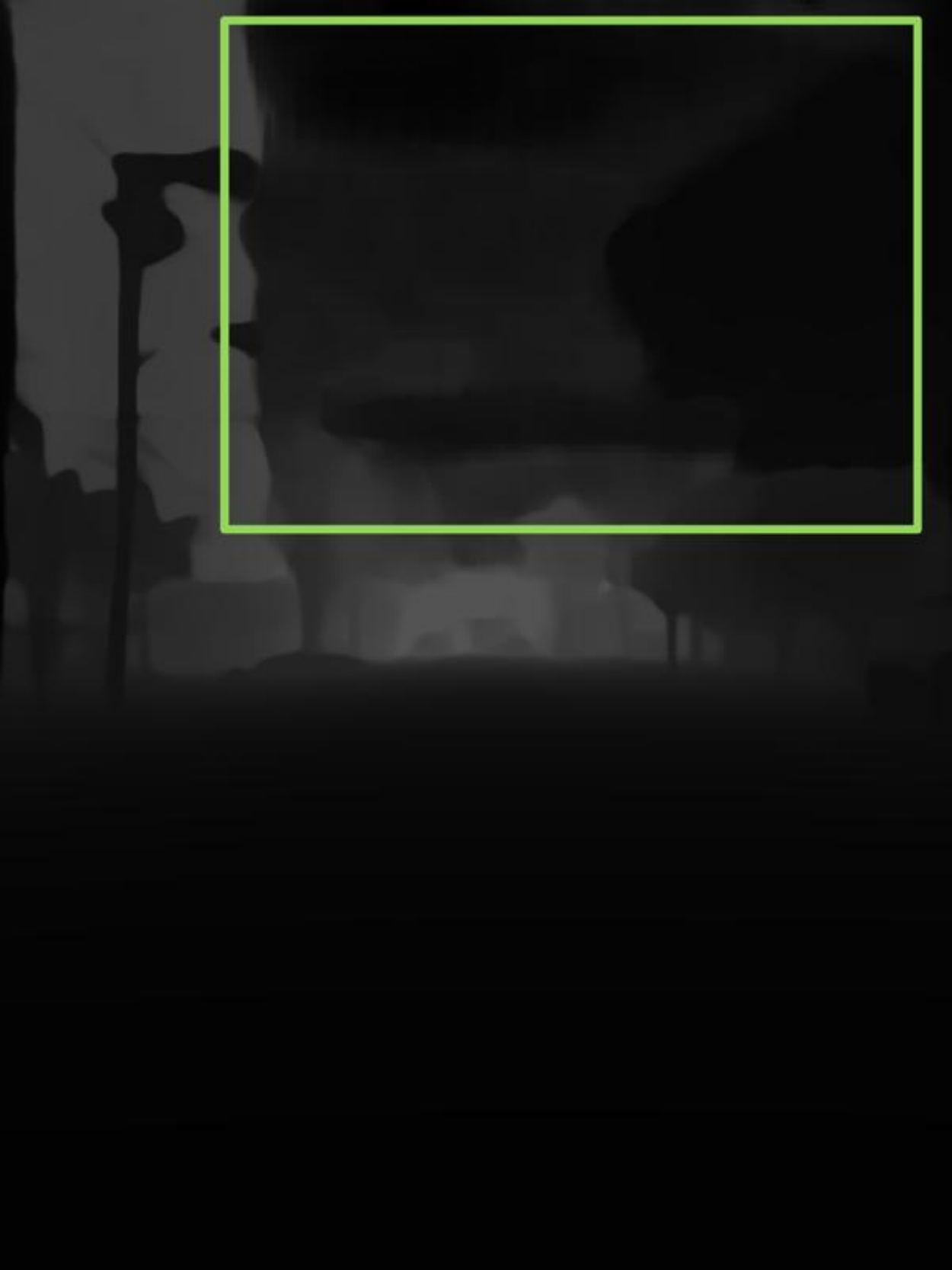}
\end{subfigure} &
\begin{subfigure}[b]{0.3\textwidth}
   \includegraphics[width=\textwidth]{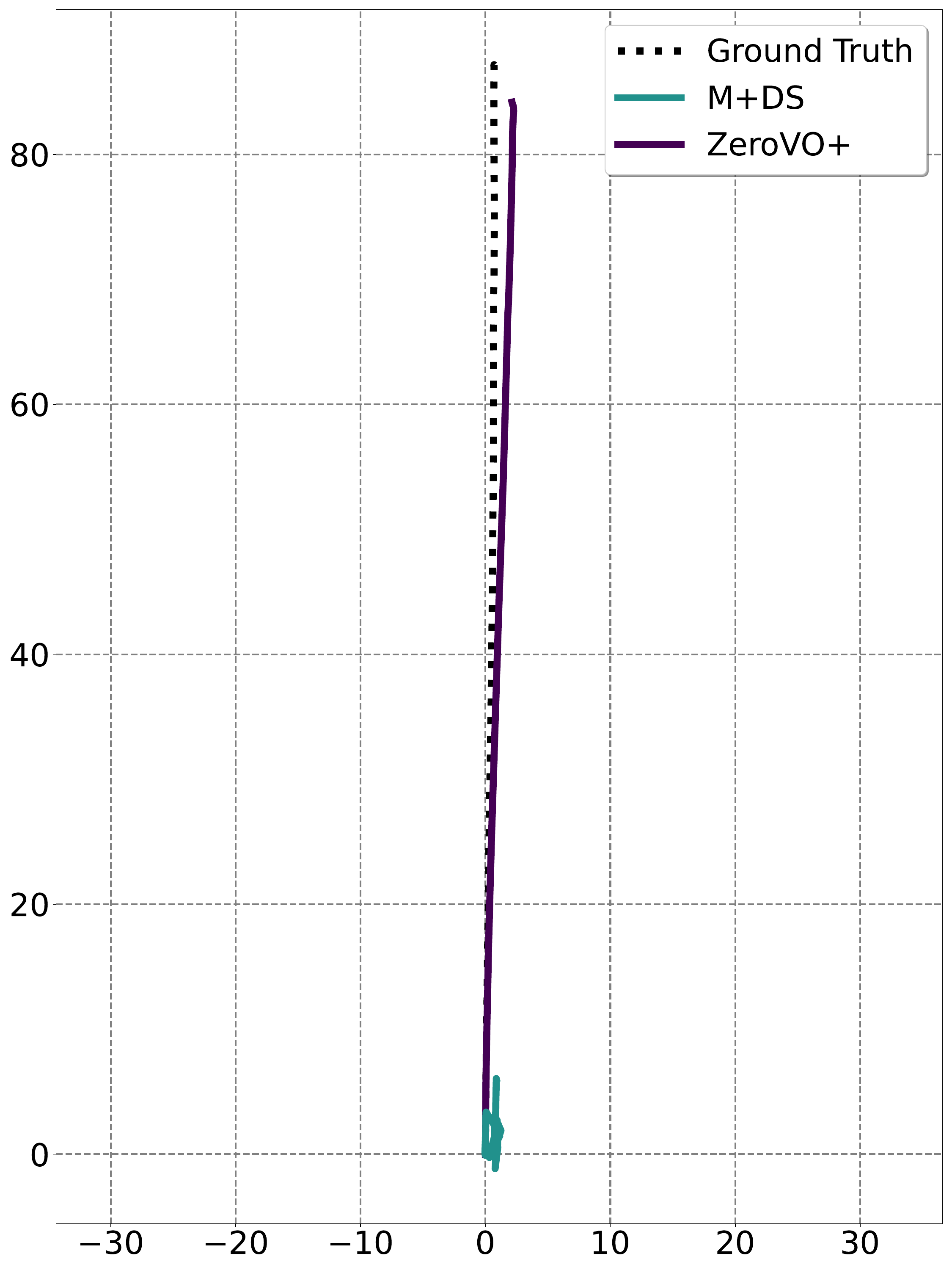}
\end{subfigure} \\
\hline

\hline
\begin{subfigure}[b]{0.3\textwidth}
   \includegraphics[width=\textwidth]{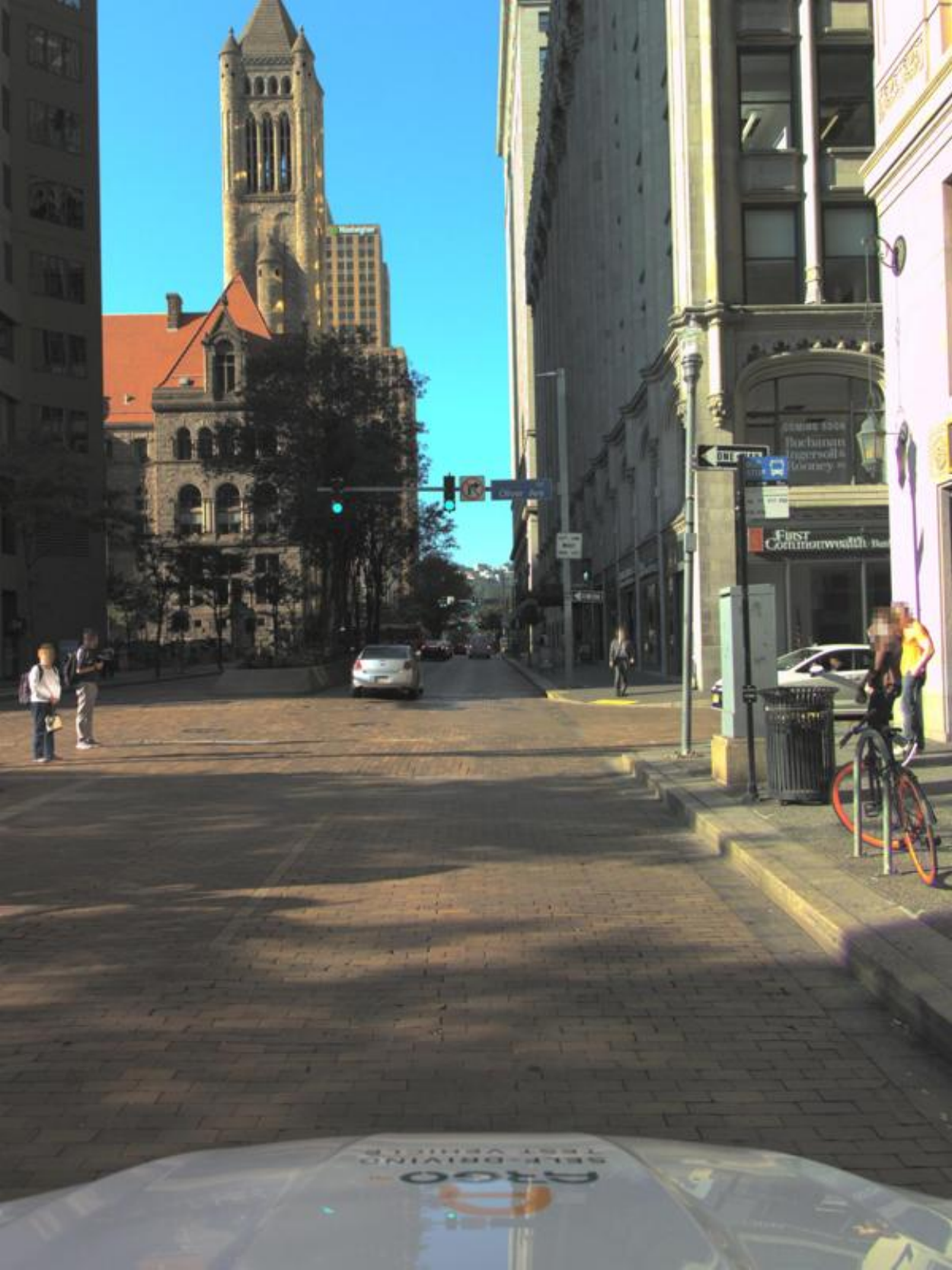}
\end{subfigure} &
\begin{subfigure}[b]{0.3\textwidth}
   \includegraphics[width=\textwidth]{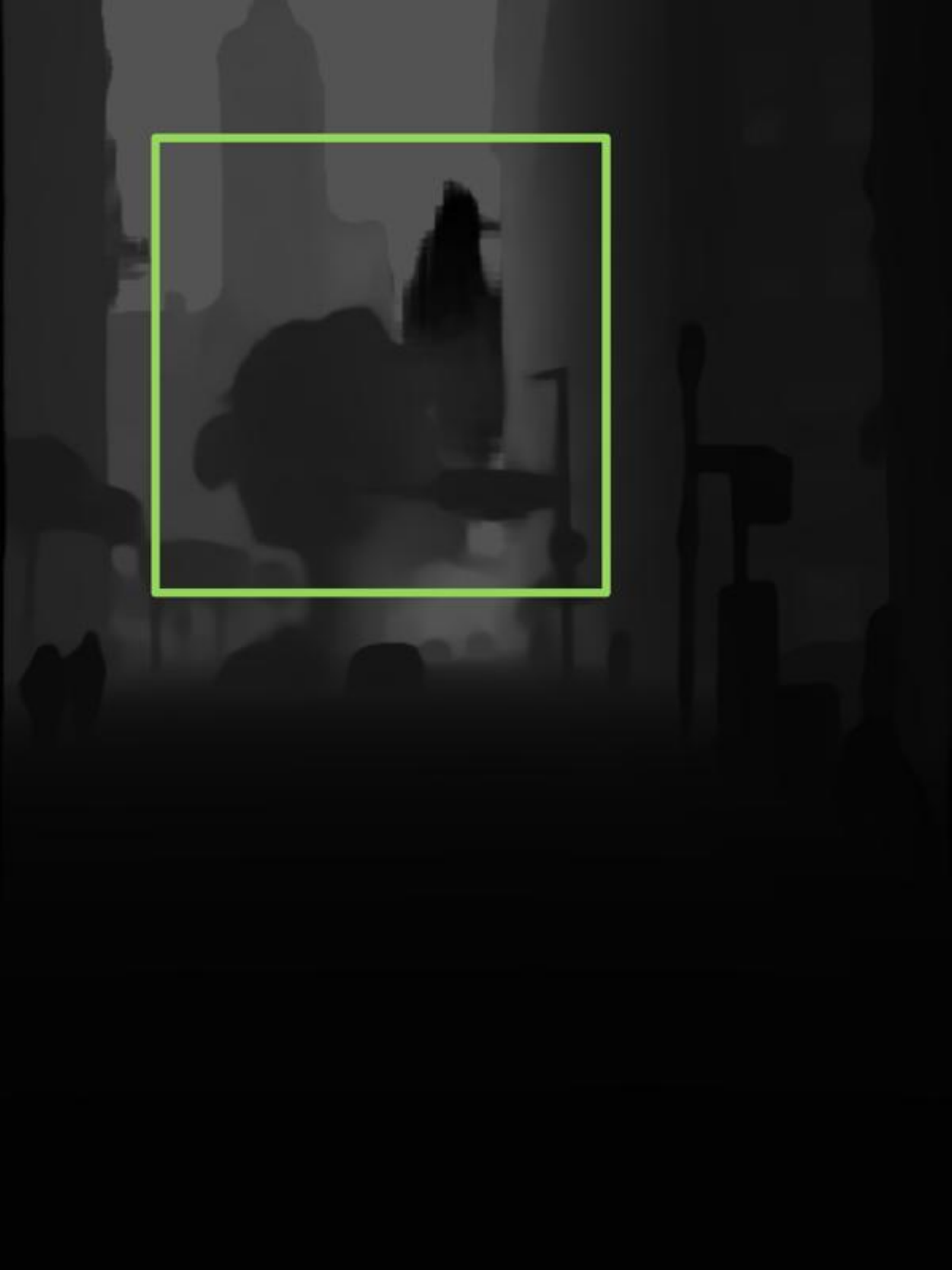}
\end{subfigure} &
\begin{subfigure}[b]{0.3\textwidth}
   \includegraphics[width=\textwidth]{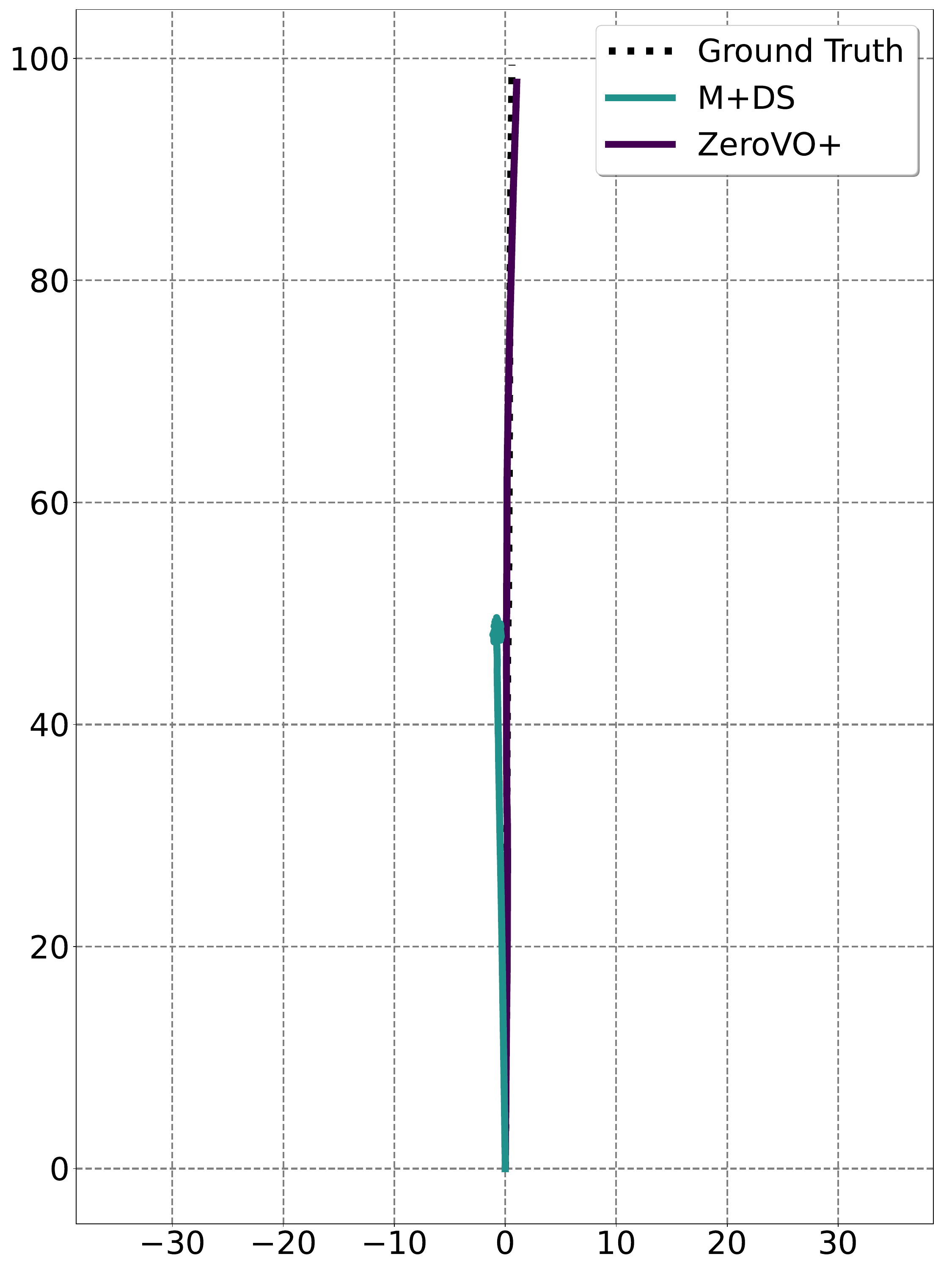}
\end{subfigure} \\

\hline

\hline
\begin{subfigure}[b]{0.3\textwidth}
   \includegraphics[width=\textwidth]{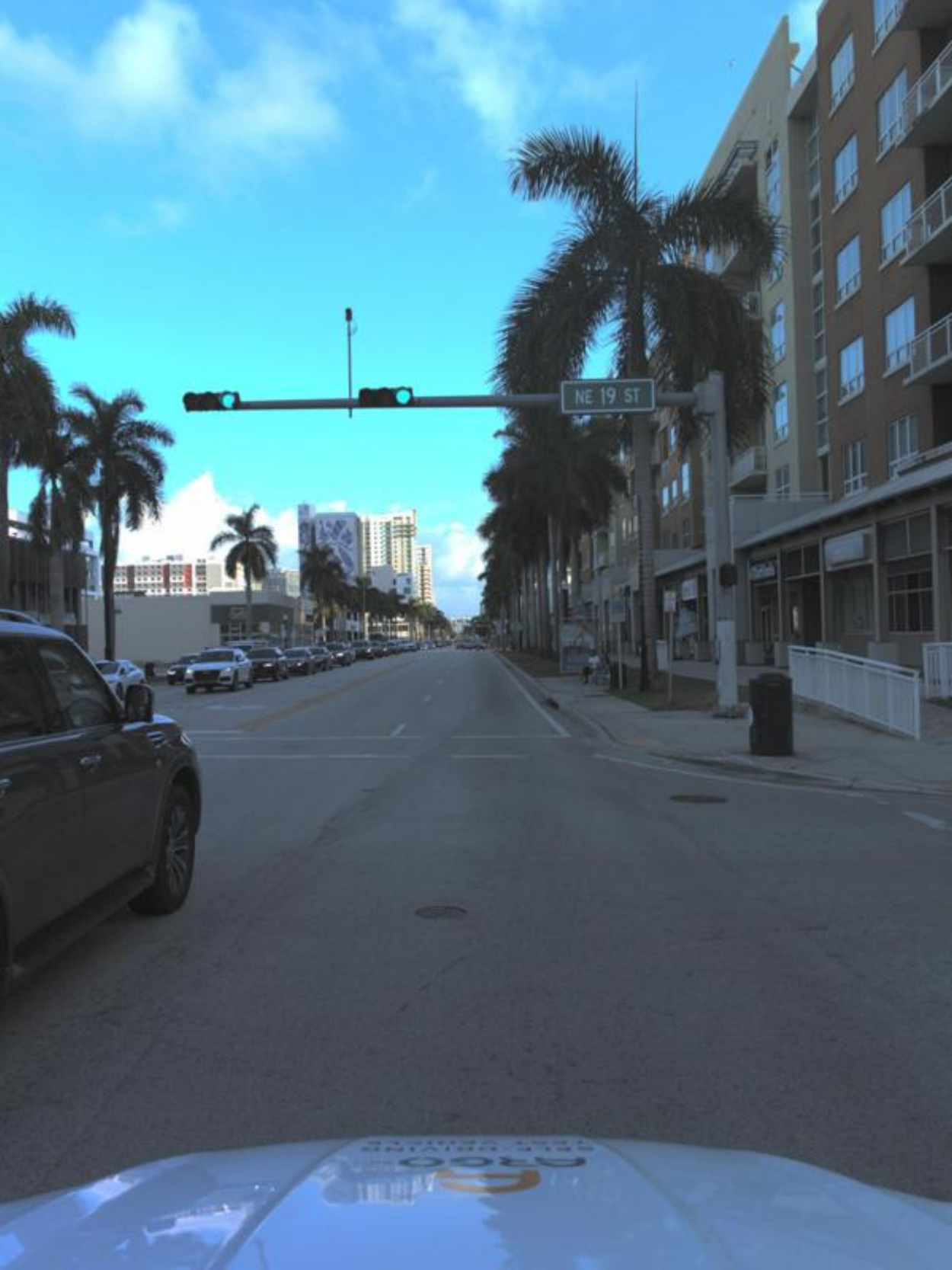}
\end{subfigure} &
\begin{subfigure}[b]{0.3\textwidth}
   \includegraphics[width=\textwidth]{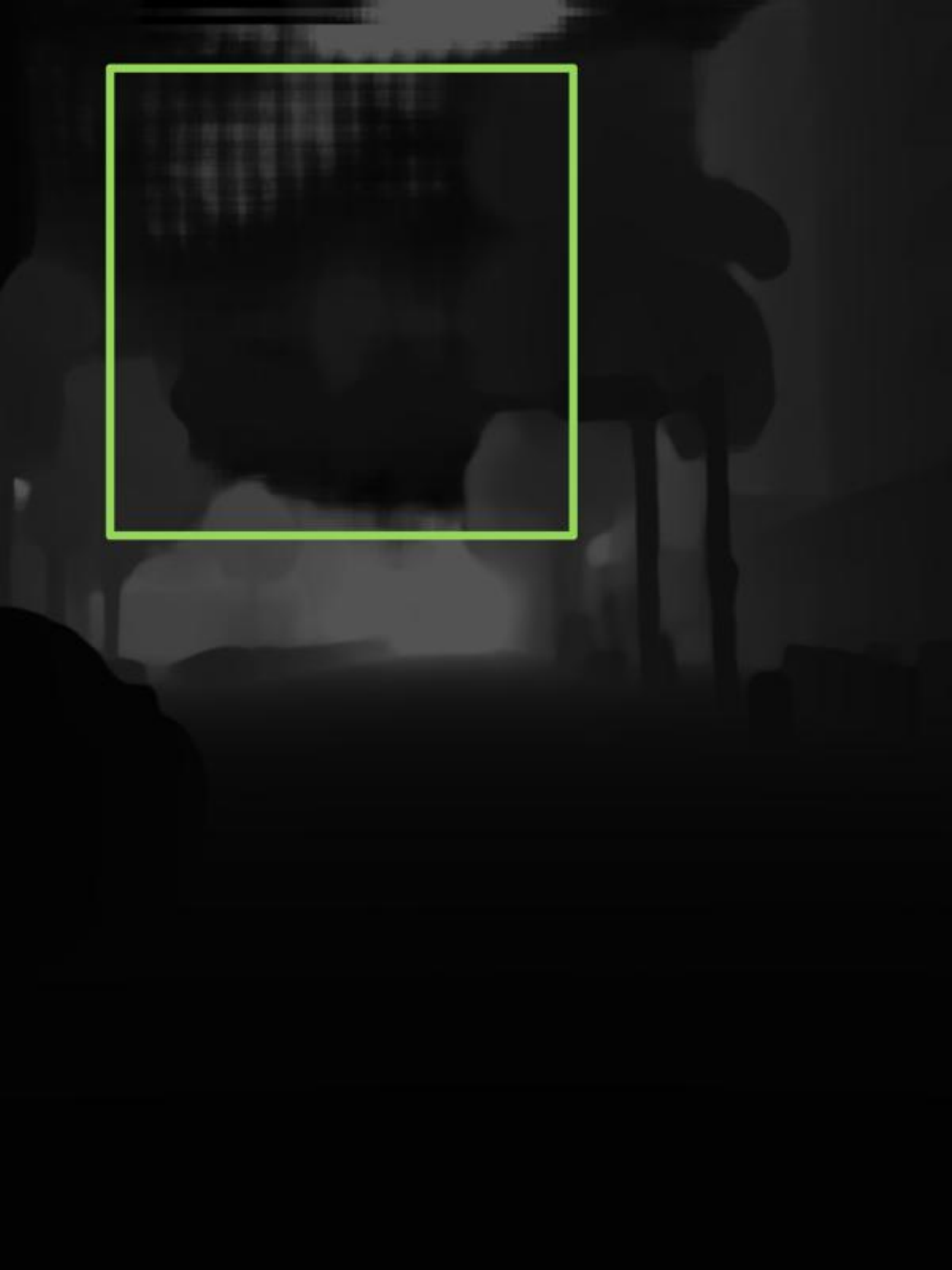}
\end{subfigure} &
\begin{subfigure}[b]{0.3\textwidth}
   \includegraphics[width=\textwidth]{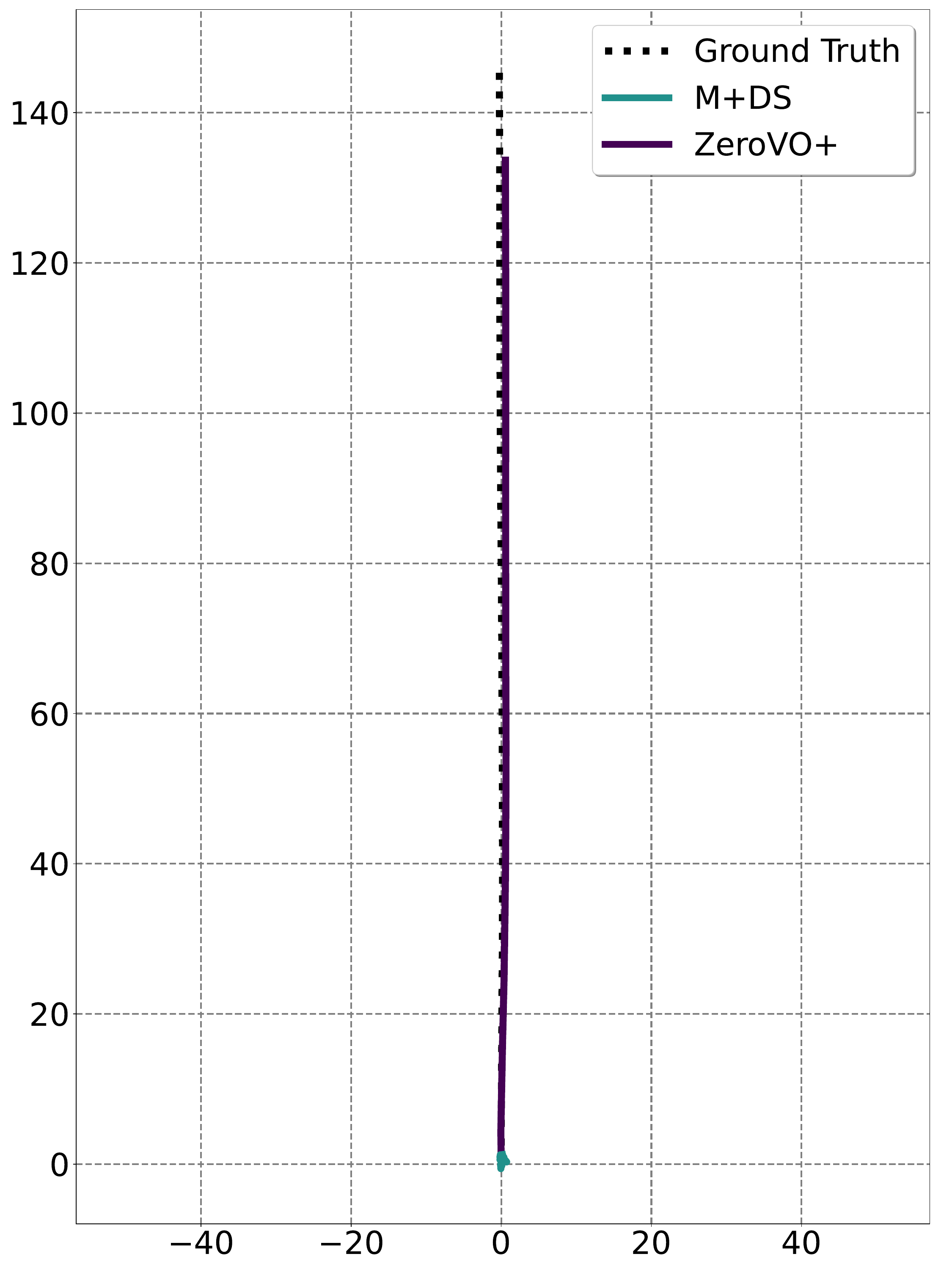}
\end{subfigure} \\
\hline
\end{tabular}

\end{table}

\begin{table}[!t]
\centering
\caption{\textbf{Qualitative Failure Cases on Argoverse.} This figure illustrates qualitative results of failure cases on Argoverse with Metric3Dv2 combined with Droid-SLAM. The left column displays the RGB input images, the middle column shows the corresponding depth predictions (darker pixels indicate closer distances), and the right column compares trajectory estimations. The green boxes in the depth images highlight regions where Metric3Dv2 struggles due to environmental factors such as sky reflections, clouds, lens artifacts, and glass surfaces. These challenges lead to significant trajectory drifts. }
\label{tab:argo_fail3}
\begin{tabular}{|c|c|c|}
\hline
\rowcolor{lightgray}
\textbf{RGB Image} & \textbf{Wrong Depth Image} & \textbf{Plotted Trajectory} \\

\hline
\begin{subfigure}[b]{0.3\textwidth}
   \includegraphics[width=\textwidth]{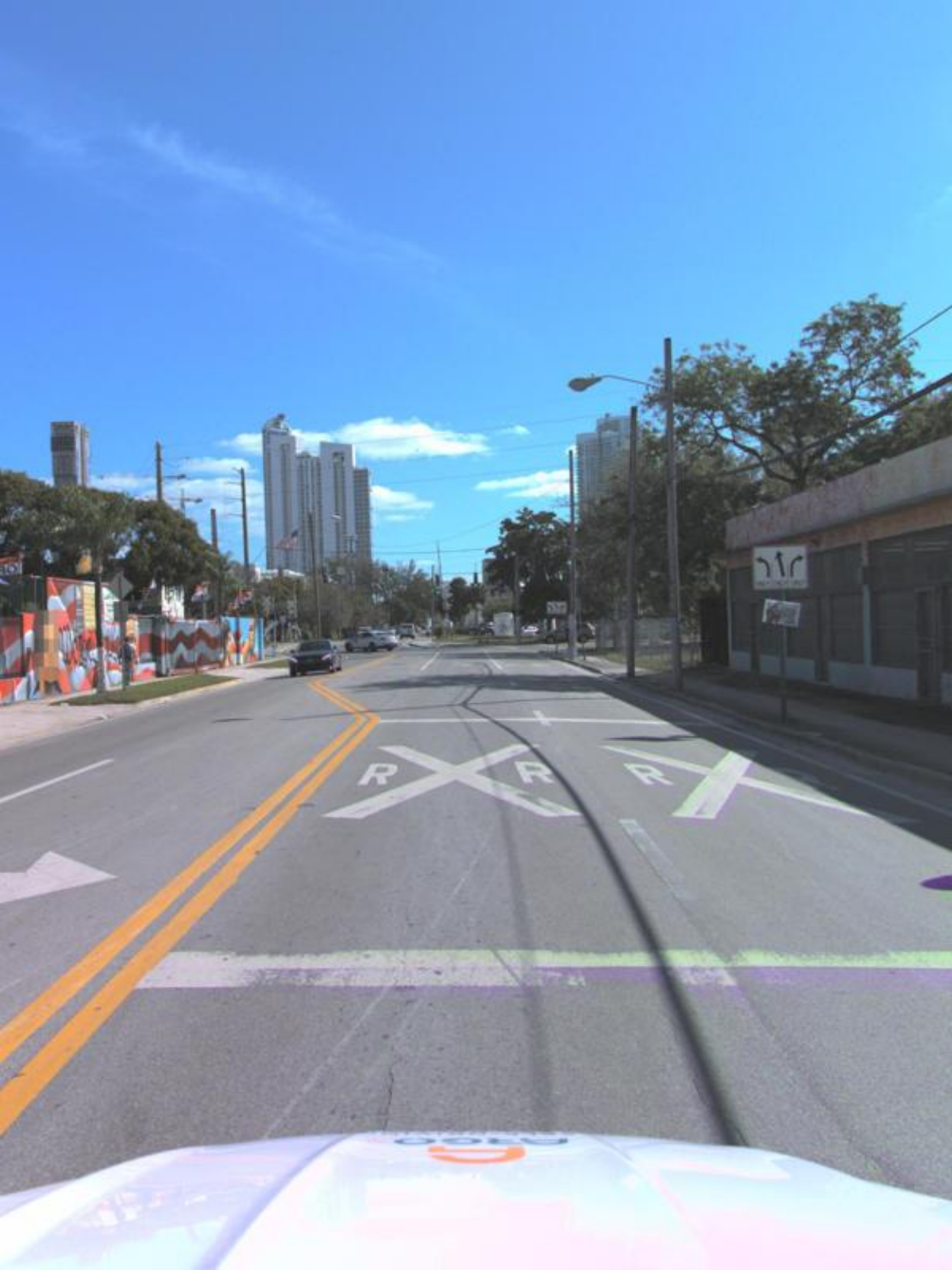}
\end{subfigure} &
\begin{subfigure}[b]{0.3\textwidth}
   \includegraphics[width=\textwidth]{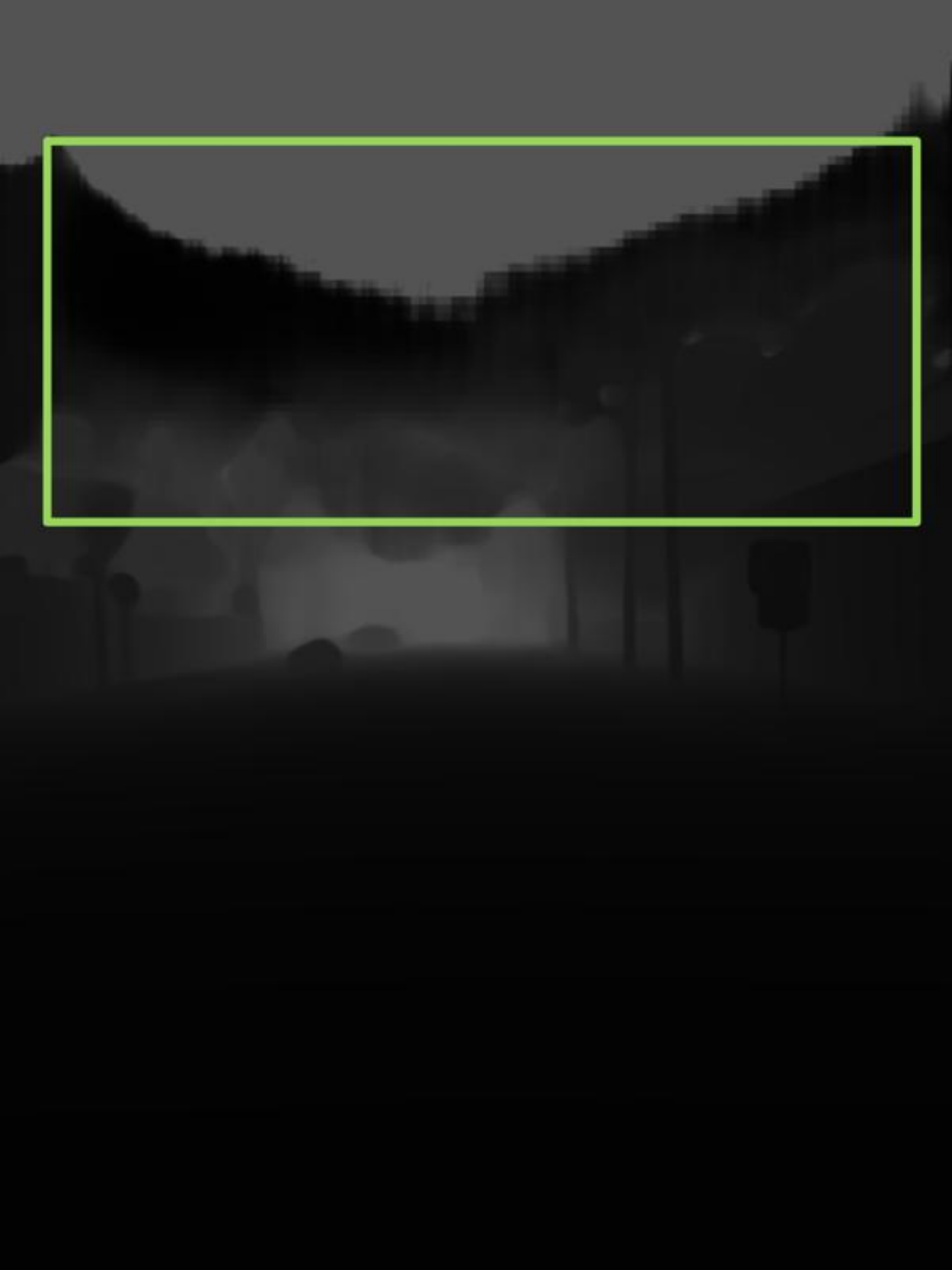}
\end{subfigure} &
\begin{subfigure}[b]{0.3\textwidth}
   \includegraphics[width=\textwidth]{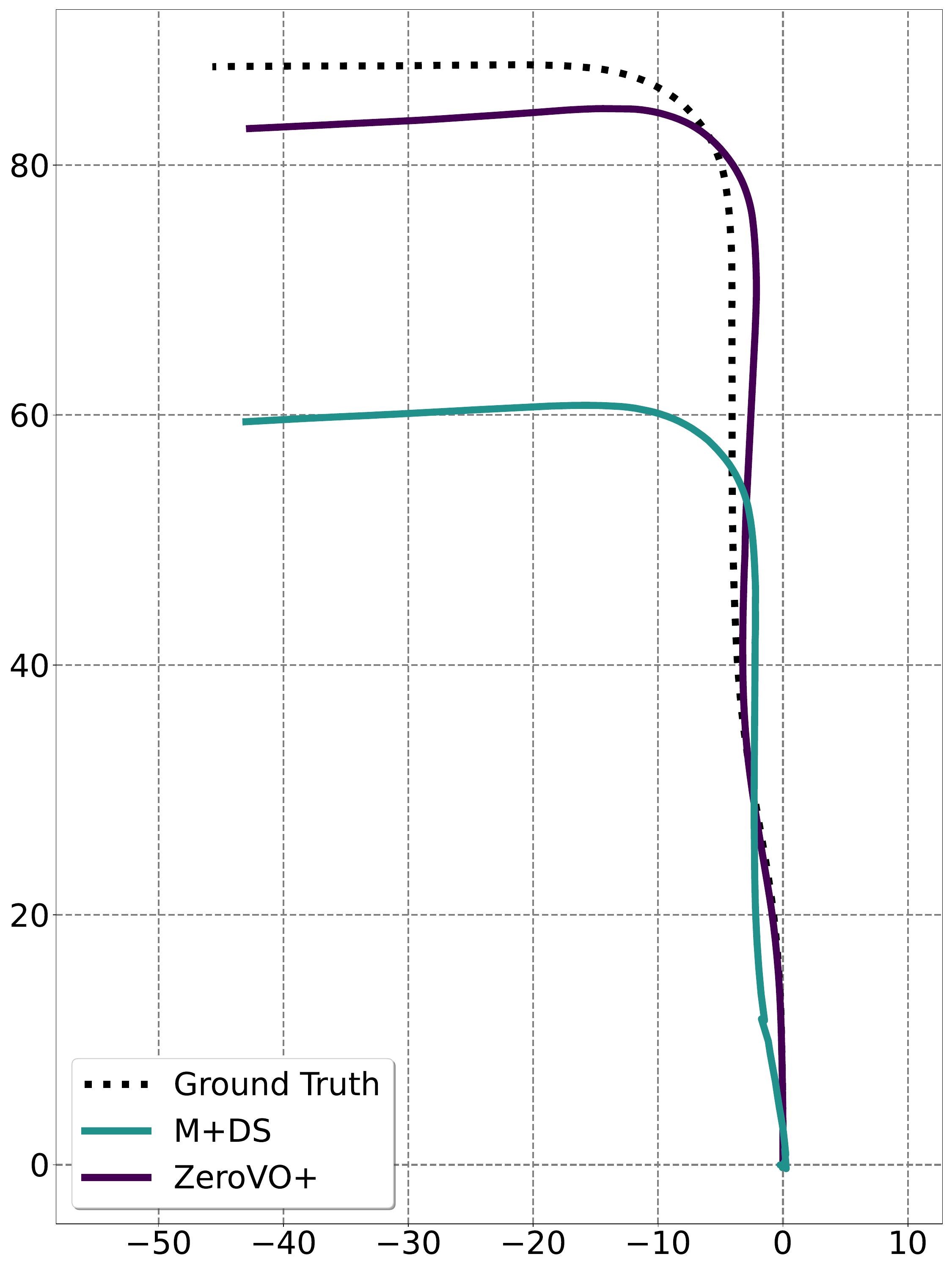}
\end{subfigure} \\
\hline

\hline
\begin{subfigure}[b]{0.3\textwidth}
   \includegraphics[width=\textwidth]{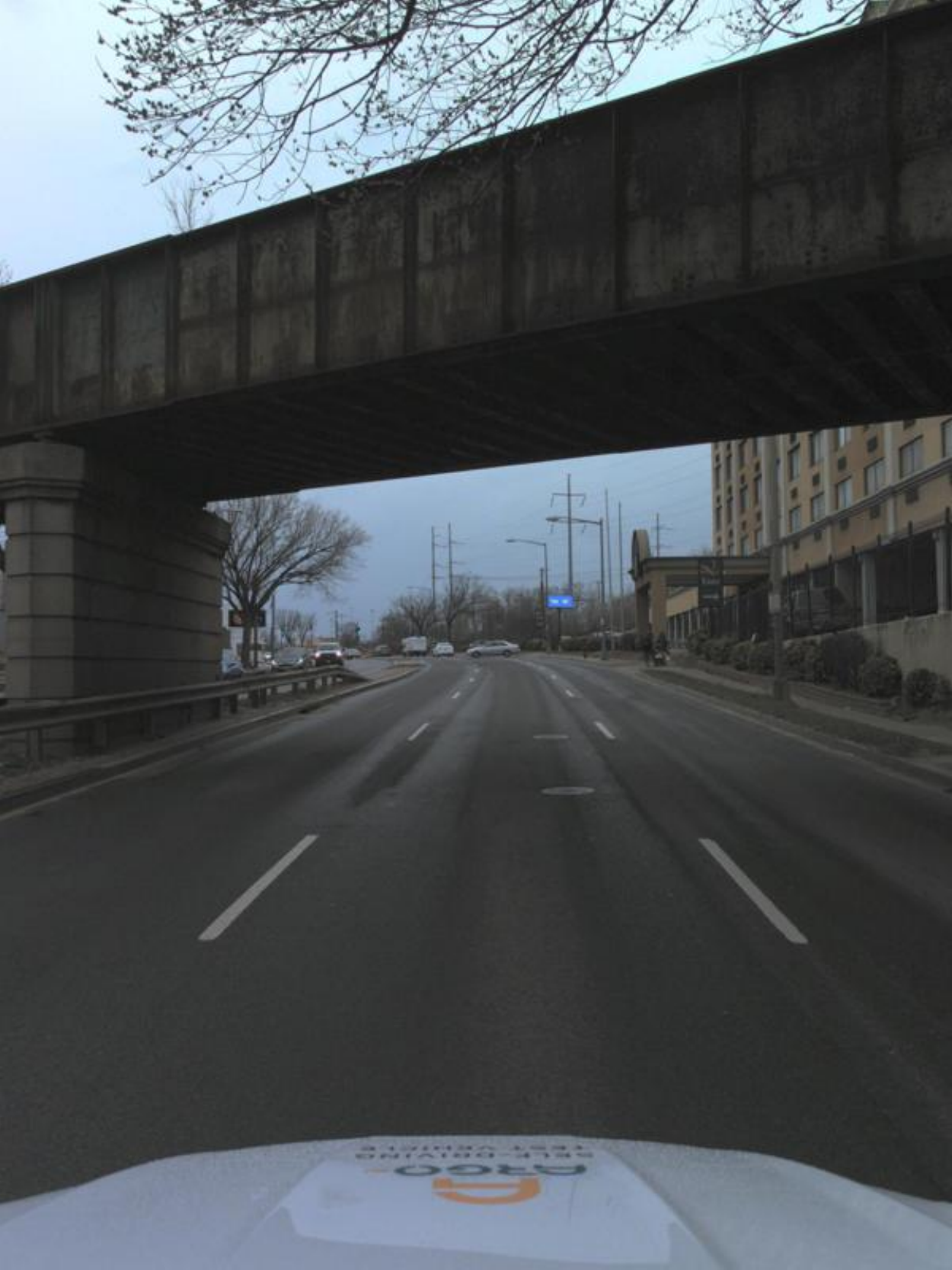}
\end{subfigure} &
\begin{subfigure}[b]{0.3\textwidth}
   \includegraphics[width=\textwidth]{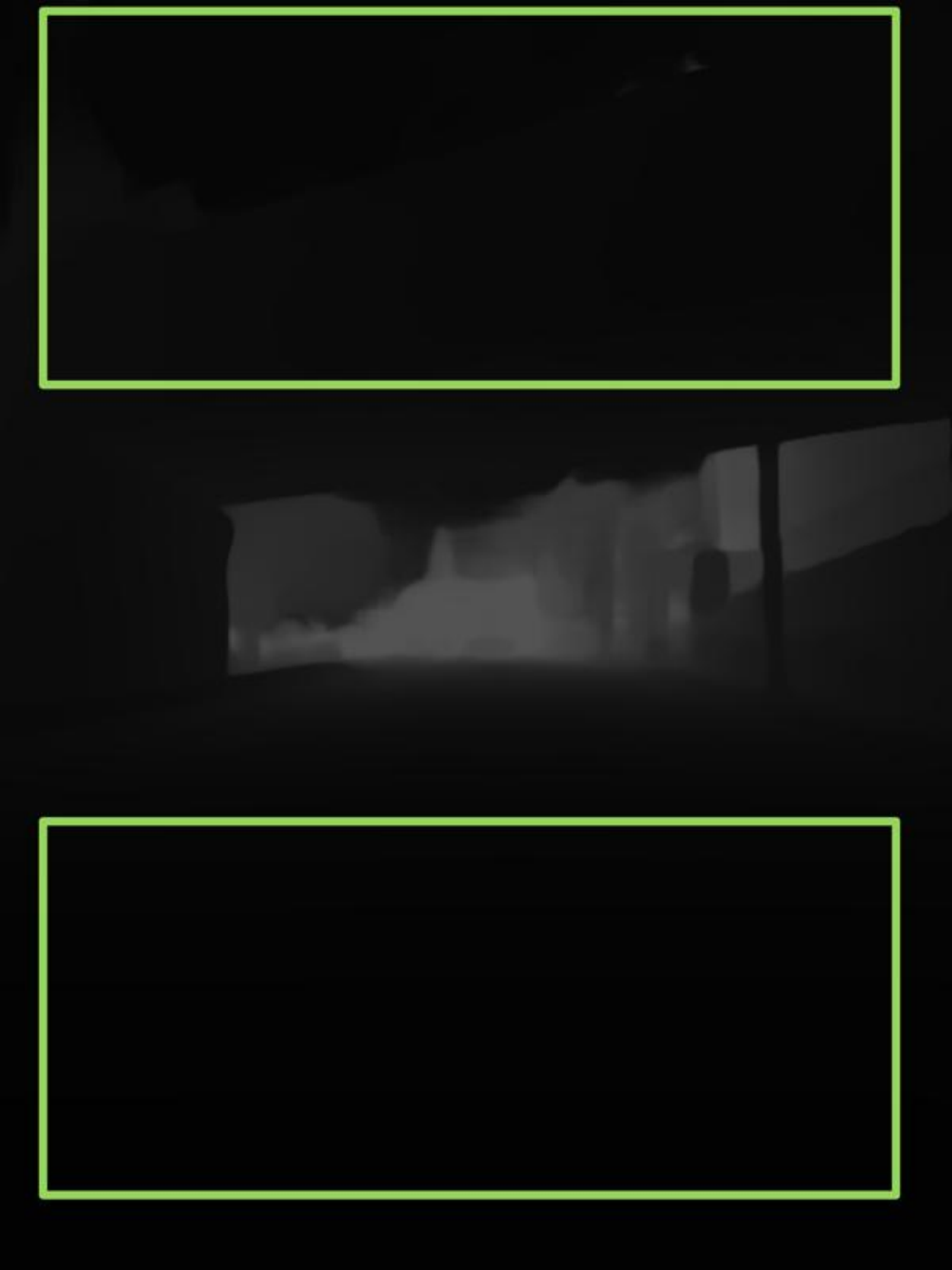}
\end{subfigure} &
\begin{subfigure}[b]{0.3\textwidth}
   \includegraphics[width=\textwidth]{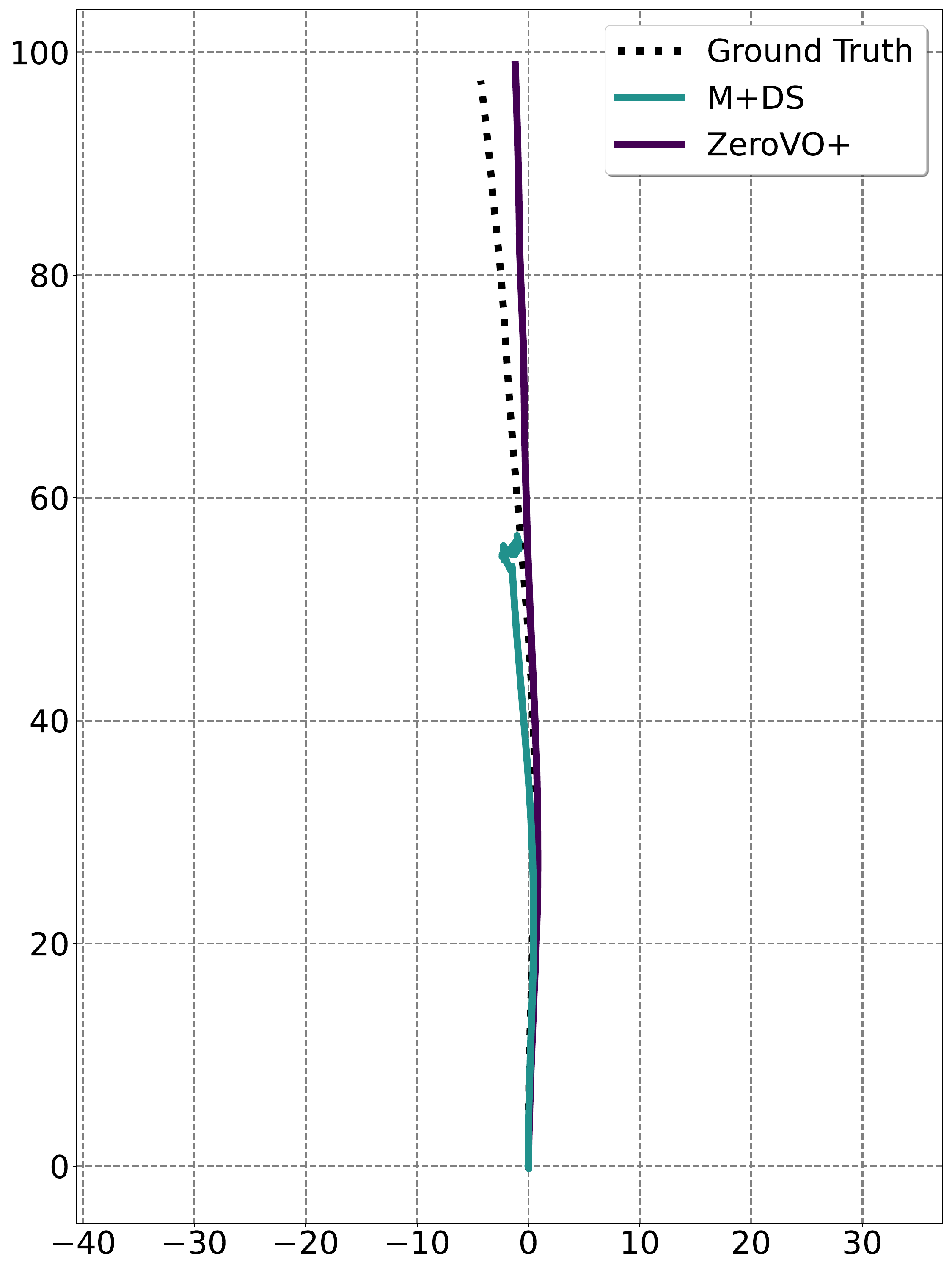}
\end{subfigure} \\

\hline

\hline
\begin{subfigure}[b]{0.3\textwidth}
   \includegraphics[width=\textwidth]{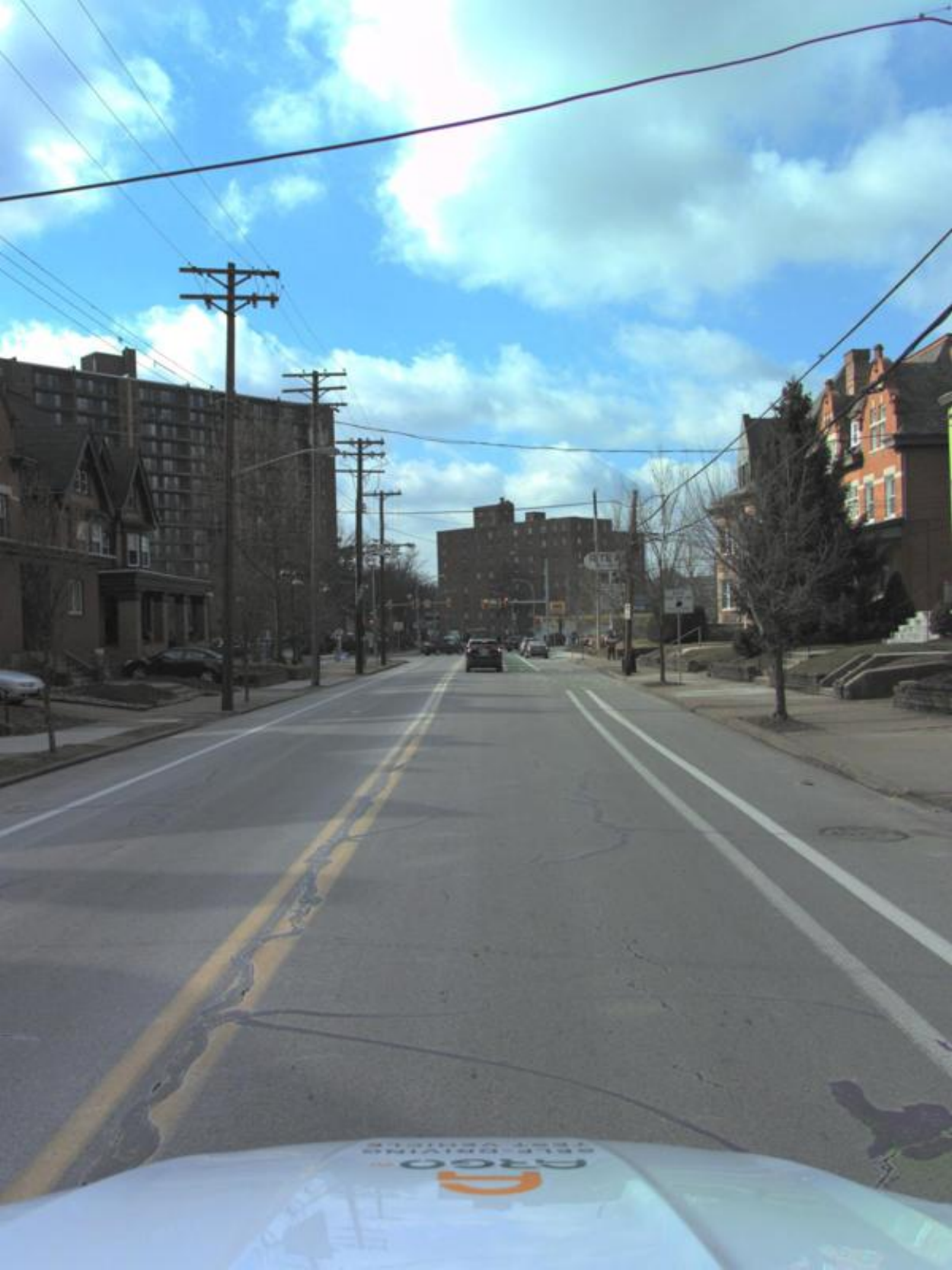}
\end{subfigure} &
\begin{subfigure}[b]{0.3\textwidth}
   \includegraphics[width=\textwidth]{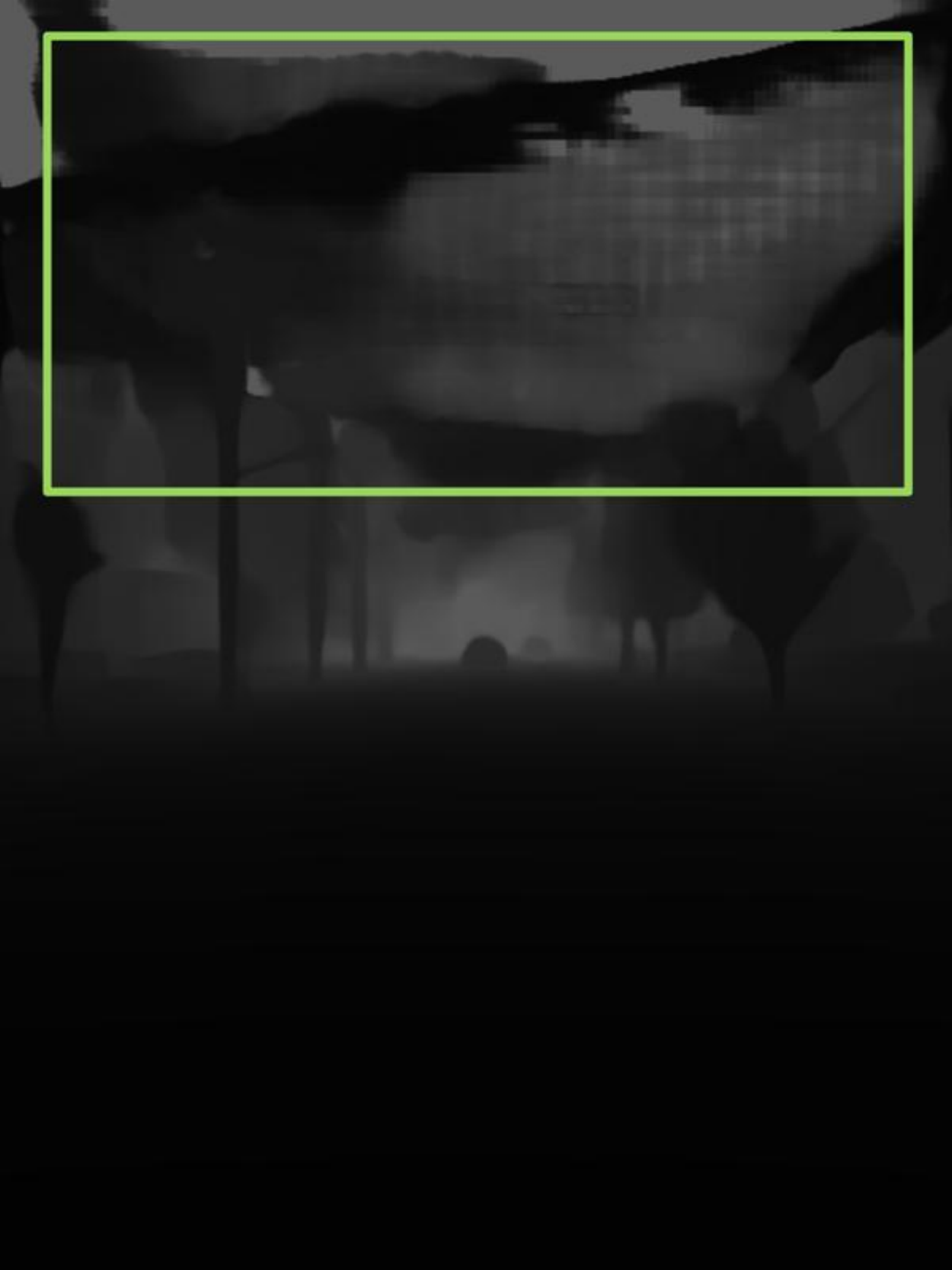}
\end{subfigure} &
\begin{subfigure}[b]{0.3\textwidth}
   \includegraphics[width=\textwidth]{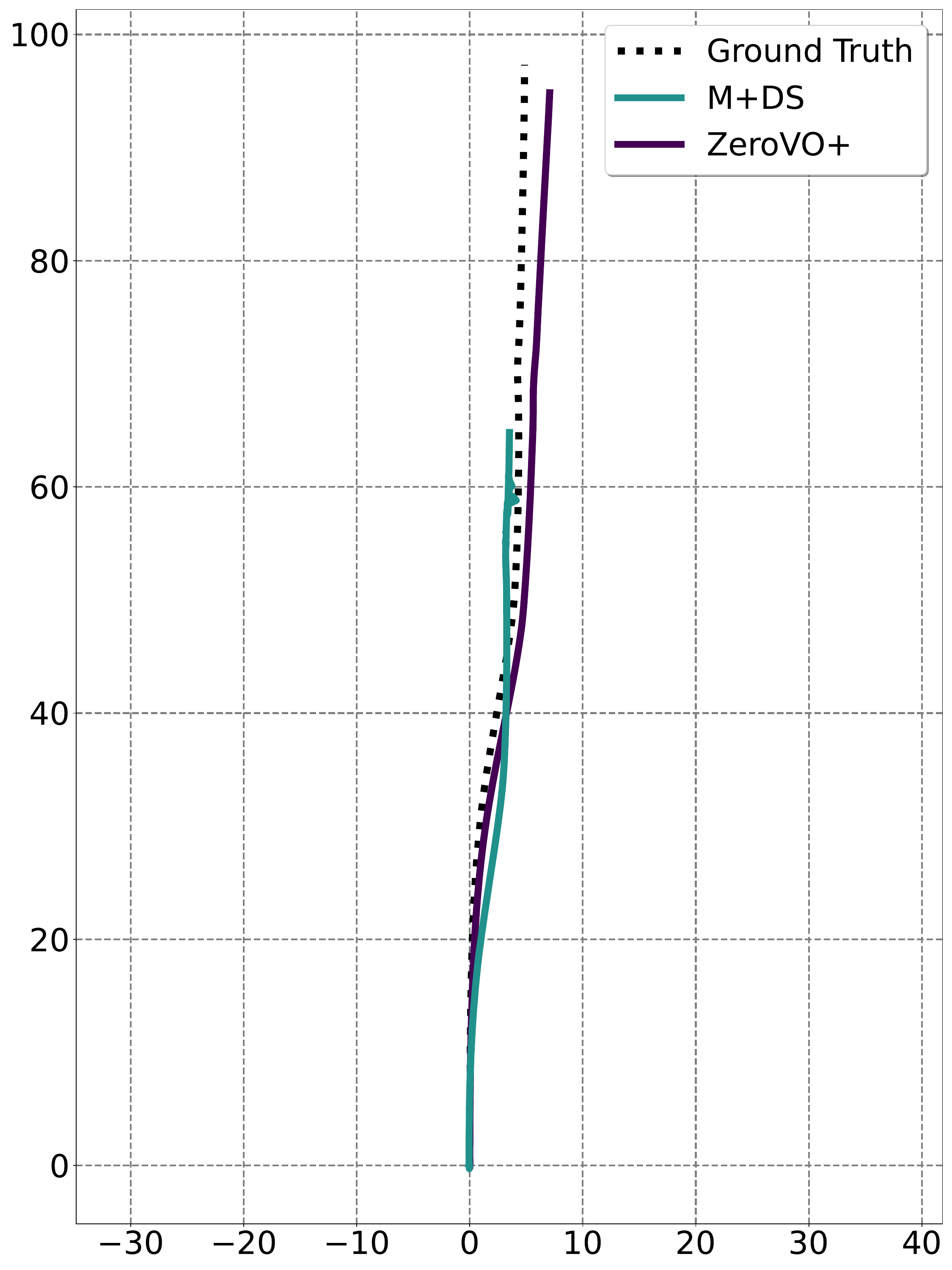}
\end{subfigure} \\
\hline
\end{tabular}

\end{table}

\clearpage

\end{document}